\documentclass{article} 

\PassOptionsToPackage{dvipsnames,table}{xcolor}
\usepackage{colortbl}
\definecolor{cvprblue}{rgb}{0.21,0.49,0.74}
\usepackage{iclr2025_conference,times}

\usepackage{amsmath,amsfonts,bm}

\def\eqref#1{equation~\ref{#1}}

\def\1{\bm{1}}

\DeclareMathAlphabet{\mathsfit}{\encodingdefault}{\sfdefault}{m}{sl}
\SetMathAlphabet{\mathsfit}{bold}{\encodingdefault}{\sfdefault}{bx}{n}

 \usepackage{diagbox}

\usepackage{hyperref}

\usepackage{float}
\usepackage[utf8]{inputenc} 
\usepackage[T1]{fontenc}    
\usepackage{url}   
\usepackage{lipsum}
\usepackage{booktabs}       
 \usepackage{nicefrac}       
\usepackage{microtype}      
\usepackage{xcolor}         
\usepackage{graphicx}
\usepackage{caption}
\usepackage{subcaption}
\usepackage{tabularx}
 \usepackage[utf8]{inputenc}
 \usepackage{wrapfig}

\usepackage{tablefootnote}

  \usepackage{subcaption,siunitx,booktabs}
\usepackage{array}
\usepackage{siunitx}
\usepackage{multirow}
\usepackage{nicematrix}
\usepackage{adjustbox}
 
\usepackage{soul}
 \usepackage{arydshln}
\usepackage{comment}

 \usepackage{algpseudocode}
\usepackage{algorithm}
 \usepackage{amsmath}

\setcitestyle{numbers}

\title{Adversarially Robust Out-of-Distribution Detection Using Lyapunov-Stabilized Embeddings}

\author{Hossein Mirzaei \& Mackenzie W. Mathis\\
École Polytechnique Fédérale de Lausanne (EPFL)\\
Geneva, CH \\
\texttt{hossein.mirzaeisadeghlou@epfl.ch, mackenzie.mathis@epfl.ch
} \\
}

\definecolor{lightgray}{gray}{0.95}
\newcommand{\gr}[1]{\textcolor{gray}{ #1}}

\definecolor{lightblue}{rgb}{0.88, 0.95, 1.0}

\iclrfinalcopy 

\begin{document}

\maketitle
 \begin{abstract}
Despite significant advancements in out-of-distribution (OOD) detection, existing methods still struggle to maintain robustness against adversarial attacks, compromising their reliability in critical real-world applications. Previous studies have attempted to address this challenge by exposing detectors to auxiliary OOD datasets alongside adversarial training. However, the increased data complexity inherent in adversarial training, and the myriad of ways that OOD samples can arise during testing, often prevent these approaches from establishing robust decision boundaries. To address these limitations, we propose AROS, a novel approach leveraging neural ordinary differential equations (NODEs) with Lyapunov stability theorem in order to obtain robust embeddings for OOD detection. By incorporating a tailored loss function, we apply Lyapunov stability theory to ensure that both in-distribution (ID) and OOD data converge to stable equilibrium points within the dynamical system. This approach encourages any perturbed input to return to its stable equilibrium, thereby enhancing the model’s robustness against adversarial perturbations. To not use additional data, we generate fake OOD embeddings by sampling from low-likelihood regions of the ID data feature space, approximating the boundaries where OOD data are likely to reside. To then further enhance robustness, we propose the use of an orthogonal binary layer following the stable feature space, which maximizes the separation between the equilibrium points of ID and OOD samples. We validate our method through extensive experiments across several benchmarks, demonstrating superior performance, particularly under adversarial attacks. Notably, our approach improves robust detection performance from 37.8\% to \textbf{80.1\%} on CIFAR-10 vs. CIFAR-100 and from 29.0\% to \textbf{67.0\%} on CIFAR-100 vs. CIFAR-10.Code and pre-trained models are available at \url{https://github.com/AdaptiveMotorControlLab/AROS}.
\end{abstract}

\section{Introduction}

Deep neural networks have demonstrated remarkable success in computer vision, achieving significant results across a wide range of tasks. However, these models are vulnerable to adversarial examples—subtly altered inputs that can lead to incorrect predictions \cite{papernot2016limitations,carlini2017towards,szegedy2013intriguing}. As a result, designing a defense mechanism has emerged as a critical task. Various strategies have been proposed, and adversarial training has become one of the most widely adopted approaches \cite{madry2018towards,zhang2019theoretically,tramer2018ensemble}. Recently, Neural Ordinary Differential Equations (NODEs) have attracted attention as a defense strategy by leveraging principles from control theory. By leveraging the dynamical system properties of NODEs, and imposing stability constraints, these methods aim to enhance robustness with theoretical guarantees. However, they have been predominantly studied in the context of classification tasks \cite{ledyaev1999lyapunov,carrara2019robustness,svoboda2019peernets,rahnama2020robust,li2020implicit,rodriguez2022lyanet,yang2022closer,dashkovskiy2023robust,zeqiri2023efficient}, and not in out-of-distribution (OOD) detection.

\begin{figure}[t] 
    \centering 
    \includegraphics[width=\linewidth]{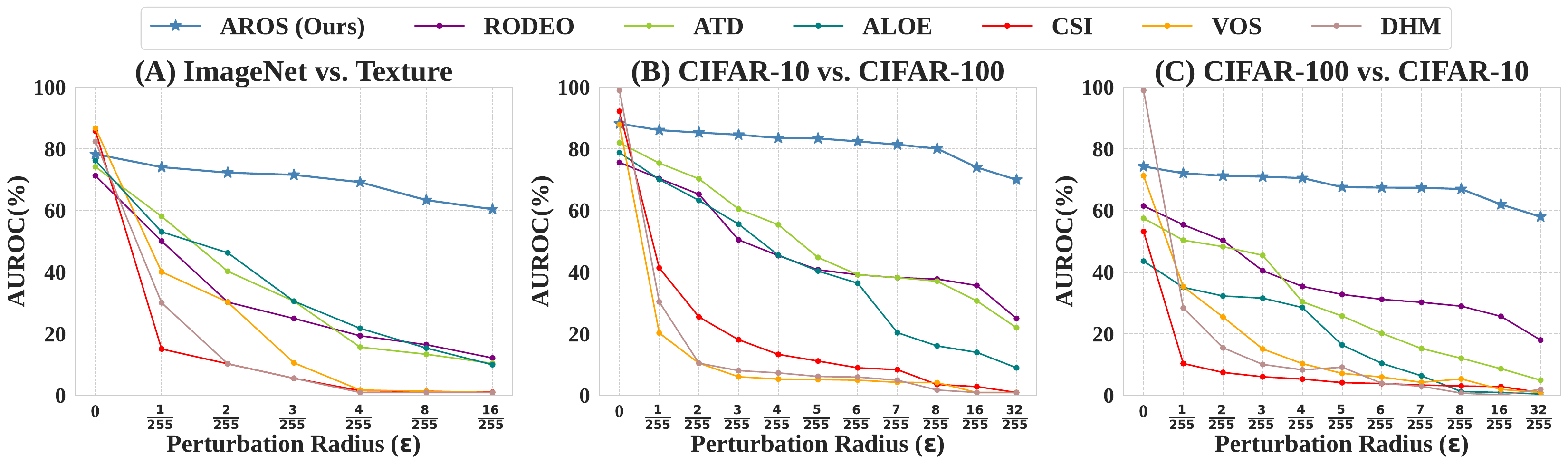} 
    \caption{\textbf{OOD detection performance for various models under different perturbation magnitudes.} The perturbations are generated using  $\text{PGD}^{1000}$
 ($\ell_\infty$) attack targeting both test ID and OOD samples. (A) ImageNet is used as the ID dataset, while the Texture dataset is used as the OOD during test time. (B) CIFAR-10 is utilized as the ID, with CIFAR-100 as the OOD. (C) CIFAR-100 is used as the ID, with CIFAR-10 as the OOD. A perfect detector achieves an AUROC of 100\%, a random detector scores 50\%, and a fully compromised detector under attack scores 0\%. Notably, no other model achieves detection performance \textbf{above random} (i.e., greater than 50\% AUROC) at $\epsilon=\frac{8}{255}$. } 
    \label{fig:Comparison_epsilon} 
\end{figure}

OOD detection is a safety-critical task that is crucial for deploying models in the real world. In this task, training is limited to in-distribution (ID) data, while the inference task involves identifying OOD samples, i.e., samples that deviate from the ID data \cite{yang2021generalized,bendale2015towards}. Recent advancements have demonstrated impressive performance gains across various detection benchmarks \cite{fort2021exploring,cao2022deep,xue2022boosting,du2024does}. However, a significant challenge arises concerning the robustness of OOD detectors against adversarial attacks. An adversarial attack on a detector involves introducing minor perturbations to test samples, causing the detector to predict OOD as ID samples or vice versa. Yet, a robust OOD detector is  imperative, especially in scenarios like medical diagnostics and autonomous driving \cite{petersen2010alzheimer,goodge2021robustness,kong2021opengan,roberts2021noveltyIndustry,lo2022adversarially}. Recently, several approaches have sought to address this challenge by first demonstrating that relying solely on ID data is insufficient for building adversarially robust detectors \cite{hendrycks2018deep,chen2020robust,shao2020open,goodge2021robustness,chen2021atom,lo2022adversarially,meinke2022provably,fort2022adversarial,shao2022open,azizmalayeri2022your,franco2023diffusion,bethune2023robust,mirzaeirodeo,lorenz2024deciphering}. Consequently, new methods propose incorporating copious amounts of auxiliary OOD data in conjunction with adversarial training to improve the detector's robustness. While effective, a significant gap remains between detector performance on clean data and their robustness against adversarial attacks (see Figure \ref{fig:Comparison_epsilon}, Tables \ref{Table1:Cifar_OOD}, \ref{Table2.a:All_OOD_Result}, and \ref{Table2.b:Open-Set Recognition}).

This performance gap primarily arises from the wide variety of potential OOD samples encountered during testing. Relying exclusively on an auxiliary dataset to generate perturbed OOD data can bias the model toward specific OOD instances, thereby compromising the detector’s ability to generalize to unseen OOD data during inference~\cite{yang2021generalized, du2022vos, morteza2022provable, ming2022poem, zou2024adversarial,mirzaei2022fake}. This limitation is particularly pronounced in adversarial settings, where adversarial training demands a higher level of data complexity compared to standard training \cite{madry2017towards,schmidt2018adversarially,nakkiran2019adversarial,zhang2019theoretically,addepalli2022efficient}. Additionally, the collection of auxiliary OOD data is a costly process, as it must be carefully curated to avoid overlap with ID semantics to ensure that the detector is not confused by data ambiguities \cite{du2022vos,ming2022poem}. Finally, as our empirical analysis reveals, existing OOD detection methods are vulnerable even to non-adversarial perturbations -- a concerning issue for open-world applications, where natural factors such as lighting conditions or sensor noise can introduce significant variability \cite{hendrycks2019benchmarking} (see Table \ref{Corruption_table}).

\textbf{Our Contribution:} To address these challenges, we propose AROS (\textbf{A}dversarially \textbf{R}obust \textbf{O}OD Detection through \textbf{S}tability), a novel approach that leverages NODEs with the Lyapunov stability theorem (Figure \ref{fig:example_figure}). This constraint asserts that small perturbations near stable equilibrium points decay over time, allowing the system state to converge back to equilibrium. By ensuring that both ID and OOD data are stable equilibrium points of the detector, the system’s dynamics mitigate the effects of perturbations by guiding the state back to its equilibrium. Instead of using extra OOD image data, we craft fake OOD samples in the embedding space by estimating the ID boundary. Additionally, we show that adding an orthogonal binary layer increases the separation between ID and OOD equilibrium points, enhancing robustness. We evaluate AROS under both adversarial and clean setups across various datasets, including large-scale datasets such as ImageNet \cite{deng2009imagenet} and real-world medical imaging data (i.e., ADNI \cite{petersen2010alzheimer}), and compare it to previous state-of-the-art methods. Under adversarial scenarios, we apply strong attacks, including $\text{PGD}^{1000}$ \cite{madry2017towards}, AutoAttack \cite{DBLP:journals/corr/abs-2003-01690}, and Adaptive AutoAttack \cite{liu2022practical}.

\section{Preliminaries} \label{Preliminaries}
\textbf{Out-of-Distribution Detection.} In an OOD detection setup, it is assumed that there are two sets: an ID dataset and an OOD dataset. We denote the ID dataset as $\mathcal{D}^{\text{in}}$, which consists of pairs $(\mathbf{x}^{\text{in}}, y^{\text{in}})$, where $\mathbf{x}^{\text{in}}$ represents the ID data, and $y^{\text{in}} \in \mathcal{Y}^{\text{in}} := \{1, \dots, K\}$ denotes the class label. Let $\mathcal{D}^{\text{out}}$ represent the OOD dataset, containing pairs $(\mathbf{x}^{\text{out}}, y^{\text{out}})$, where $y^{\text{out}} \in \mathcal{Y}^{\text{out}} := \{K+1, \dots, K+O\}$, and $\mathcal{Y}^{\text{out}} \cap \mathcal{Y}^{\text{in}} = \emptyset$ \cite{hendrycks2017a,fort2021exploring}. In practice, different datasets are often used for $\mathcal{D}^{\text{in}}$ and $\mathcal{D}^{\text{out}}$. Alternatively, another scenario is called open-set recognition, where a subset of classes within a dataset is considered as ID, while the remaining classes are considered as OOD \cite{yang2022openood,yang2021generalized,salehi2021unified,geng2020recent}. A trained model $\mathcal{F}$ assigns an OOD score $S_{\mathcal{F}}$ to each test input, with higher scores indicating a greater likelihood of being OOD. 
 
\noindent\textbf{Adversarial Attack on OOD Detectors.} \label{OOD_attack_target}
Adversarial attacks involve perturbing an input sample \(x\) to generate an adversarial example \(x^*\) that maximizes the loss function \(\ell(x^*; y)\). The perturbation magnitude is constrained by \(\epsilon\) to ensure that the alteration remains imperceptible. Formally, the adversarial example is defined as \(x^* = \arg \max_{x^{\prime}} \ell(x^{\prime}; y)\), subject to \(\|x - x^*\|_{p} \leq \epsilon\), where \(p\) denotes the norm (e.g., \(p = 2\), \(\infty\)) \cite{goodfellow2014explaining,szegedy2013intriguing,madry2017towards}. A widely used attack method is Projected Gradient Descent (PGD) \cite{madry2017towards}, which iteratively maximizes the loss by following the gradient sign of \(\ell(x^*; y)\) with a step size \(\alpha\). For adversarial evaluation \cite{chen2020robust,azizmalayeri2022your,mirzaeirodeo}, we adapt this approach by targeting the OOD score \(S_\mathcal{F}(x)\). Specifically, the adversarial attack aims to mislead the detector by increasing the OOD score for ID samples and decreasing it for OOD samples, causing misprediction:
$
x_0^* = x, \quad x_{t+1}^* = x_t^* + \alpha \cdot \operatorname{sign}\left( \mathbb{I}(y) \cdot \nabla_x S_\mathcal{F}(x_t^*) \right), \quad x^* = x_n^*,
$
where $n$ is the number of steps, and \(\mathbb{I}(y) = +1\) if \(y \in \mathcal{Y}^{\text{in}}\) and \(-1\) if \(y \in \mathcal{Y}^{\text{out}}\). 
This approach is consistently applied across all attacks considered in our study.

\textbf{Neural ODE and Stability.} In the NODE framework, the input and output are treated as two distinct states of a continuous dynamical system, whose evolution is described by trainable layers parameterized by weights \(\phi\) and denoted as \( h_{\phi} \). The state of the neural ODE, represented by \( Z \), evolves over time according to these dynamics, establishing a continuous mapping between the input and output \cite{chen2018neural,dupont2019augmented,grathwohl2018ffjord}. The relationship between the input and output states is governed by the following differential equations: 
$\quad \quad 
\frac{dz(t)}{dt} = h_{\phi}(z(t), t), \quad z(0) = z_{\text{input}}, \quad z(T) = z_{\text{output}}.$

 \vspace{-5pt}
\section{Related Work} \label{related_work} 

\textbf{OOD Detection Methods.} Existing OOD detection methods can be broadly categorized into post-hoc and training-based approaches. 
Post-hoc methods involve training a classifier on ID data and subsequently using statistics from the classifier's outputs or intermediate representations to identify OOD samples. 
For instance, Hendrycks et al.~\cite{hendrycks2017a} propose using the maximum softmax probability distributions (MSP) as a metric. The MD method \cite{lee2018simple} leverages the Mahalanobis Distance in the feature space, and OpenMax \cite{bendale2016towards} recalibrates classification probabilities to improve OOD detection. 
Training-based methods, modify the training process to enhance OOD detection capabilities. Such modifications can include defining additional loss functions, employing data augmentation techniques, or incorporating auxiliary networks. Examples of training-based methods designed for standard OOD detection include VOS \cite{du2022vos}, DHM \cite{cao2022deep}, CATEX \cite{liu2024category}, and CSI \cite{tack2020csi}. On the other hand, ATOM \cite{chen2021atom}, ALOE \cite{chen2020robust}, ATD \cite{azizmalayeri2022your}, and RODEO \cite{mirzaeirodeo} have been developed specifically for robust detection. For detailed descriptions of these methods, please refer to Appendix~\ref{appendix:model_details}.

\textbf{Stable NODE for Robustness.}
 TiSODE \cite{yan2019robustness} introduces a time-invariant steady NODE to constrain trajectory evolution by keeping the integrand close to zero. Recent works employ Lyapunov stability theory to develop provable safety certificates for neural network systems, particularly in classification tasks. PeerNets \cite{svoboda2019peernets} was among the first to use control theory and dynamical systems to improve robustness. Kolter et al.~\cite{kolter2019learning} designed a Lyapunov function using neural network architectures to stabilize a base dynamics model's equilibrium. ASODE~\cite{li2022defending} uses non-autonomous NODEs with Lyapunov stability constraints to mitigate adversarial perturbations in slowly time-varying systems. LyaDEQ \cite{chu2024lyapunov} introduces a new module based on ICNN \cite{amos2017input} into its pipeline, leveraging deep equilibrium models and learning a Lyapunov function to enhance stability. SODEF \cite{kang2021stable} enhances robustness against adversarial attacks by applying regularizers to stabilize the behavior of NODE under the time-invariant assumption. In Table \ref{Table4.a:lyapunov_CLS}, we analyze these stability-based classifiers as OOD detectors and highlight the potential of Lyapunov’s theorem as a framework for robust OOD detection, and show our method's ability to improve performance over these excellent baselines.

\begin{figure}[t] 
    \centering 
    \includegraphics[width=\linewidth]{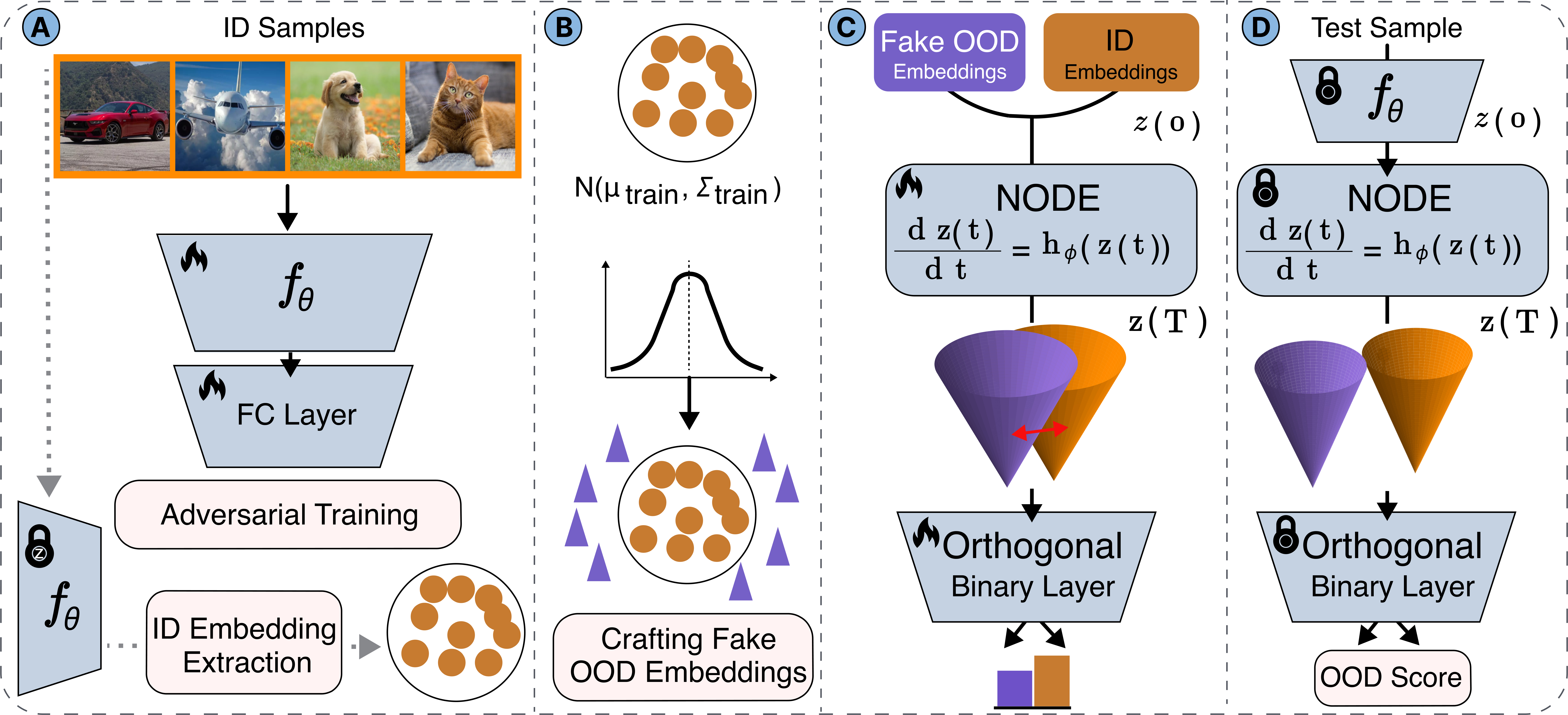}
    \caption{\textbf{An illustration of  AROS}. \textbf{(A)} To obtain robust initial features for OOD detection, we perform adversarial training on a classifier using only ID samples. \textbf{(B)} We estimate the ID distribution within the embedding space and generate fake OOD embeddings as a proxy for real OOD data. This enables the creation of two balanced classes of samples: ID and fake OOD. \textbf{(C)} The model incorporates a NODE layer $h_{\phi}$ and an Orthogonal Binary Layer $B_{\eta}$. Using these two classes, we train the pipeline with the loss function $\mathcal{L}_{\text{SL}}$ to stabilize the system dynamics. \textbf{(D)} During inference, an input passes through the feature extractor $f_{\theta}$, NODE $h_{\phi}$, and Orthogonal Binary Layer $B_{\eta}$, and the resulting likelihood from $B_{\eta}$ serves as the OOD score. The complete algorithmic workflow of AROS can be found in Appendix \ref{appendix:Psudocode}.} 
    \label{fig:example_figure} 
\end{figure}

\section{Proposed Method}

\textbf{Motivation.} A robust detector should be resistant to shifting ID test samples to OOD, and vice versa, under adversarial attack. A common approach for developing robust OOD detectors involves employing adversarial training on ID data, combined with an auxiliary real OOD dataset, to expose the detector to potential vulnerable perturbations. The core intuition is that adversarial training on ID data alone, without an accompanying OOD dataset, leaves the detector susceptible to perturbations that alter the boundary between ID and OOD data during testing \cite{chen2020robust, shao2020open, chen2021atom, goodge2021robustness, azizmalayeri2022your, lo2022adversarially, meinke2022provably, shao2022open, bethune2023robust, franco2023diffusion, mirzaeirodeo}. Beyond the unsatisfactory performance of the prior approach, there are further challenges with this strategy. A key issue is the cost of preparing an auxiliary dataset disjoint from the ID data, along with ensuring that the selected OOD images adequately cover the boundary between ID and OOD samples—a critical factor for such frameworks \cite{du2023dream,mirzaeirodeo,chen2021atom,kong2021opengan}. Moreover, adversarial training of neural networks is notably more data-intensive than standard setups, further increasing complexity \cite{schmidt2018adversarially,nakkiran2019adversarial,zhang2019theoretically,addepalli2022efficient}. There is also the concern that exclusively relying on perturbed OOD data may introduce biases toward specific OOD examples \cite{du2022vos,ming2022poem}. To address these challenges, we propose AROS, which utilizes provable stability theorems in the embedding space to develop a robust OOD detector without requiring exposure to perturbed OOD image data.

\textbf{Overview of AROS.}  AROS ensures that perturbed input samples remain close to their non-perturbed counterparts in the feature space by leveraging the Lyapunov stability theorem \cite{dawson2022safe,uddin2021altitude,sharma2020lyapunov,chang2019neural}. By using a NODE, we consider the model as a dynamical system and design it so that ID and OOD samples converge to distinct stable equilibrium points of that system. This approach prevents significant deviations in the output when adversarial perturbations are applied. However, since OOD data is unavailable, we craft fake OOD samples in the embedding space by estimating the boundaries of the ID distribution and sampling from the corresponding low-likelihood regimes. To further avoid any misprediction between OOD and ID data caused by perturbations, we maximize the distance between their equilibrium points by leveraging an orthogonal binary layer for classification. In the following, we will thoroughly explain each proposed component, highlighting the benefits of AROS.

\subsection{Fake Embedding Crafting Strategy}

There have been efforts to utilize synthetic features, primarily under clean scenarios \cite{du2022vos, ming2022poem, strater2024generalad, du2023dream}. However, for adversarial settings, prior work has often relied on a large pre-trained model and additional data. In contrast, our approach limits information to ID samples, proposing to craft OOD data from ID data in the embedding space. These generated OOD samples are subsequently utilized in the training step.

We employ a well-trained encoder to transform ID training data into robust embedding spaces. To achieve these embeddings, we first adversarially train a classifier on ID training samples using cross-entropy loss $\mathcal{L}_{\text{CE}}$ and the $\text{PGD}^{10}$($l_{\infty}$) attack. By removing the last fully connected layer from the classifier, we utilize the remaining encoder, denoted as $f_{\theta}$, to extract ID embeddings $r$, where $r = f_{\theta}(x)$ from an ID training sample $x$ (Figure~\ref{fig:example_figure}A). Specifically, by considering $\mathcal{D}_{\text{train}}^{\text{in}} $ with $K$ classes, we estimate their distribution as a $K$ class-conditional Gaussian distribution, a well-known approach in the detection literature \cite{du2022vos,cohen2021transformaly,venkataramanan2023gaussian,pmlr-v139-jin21a,MD,ren2021simple}.

We then select fake embeddings $r$ from the feature space corresponding to class $j$ such that $r \sim \mathcal{N}(\hat{\mu}_j, \hat{\Sigma}_j)$ satisfies:

$\quad \quad  \quad \quad \quad \quad \quad \quad     \frac{1}{(2\pi)^{d/2} |\hat{\Sigma}_j|^{1/2}} \exp\left( -\frac{1}{2} (r - \hat{\mu}_j)^{T} \hat{\Sigma}_j^{-1} (r - \hat{\mu}_j) \right) < \beta, \quad \quad \quad  \quad  \quad \quad   \quad  (1) \quad
$
where, $\beta$ serves as a likelihood threshold, and we set that to a small value (e.g., 0.001) (Figure~\ref{fig:example_figure}B). Additionally, we conduct an ablation study to evaluate the impact of different values of $\beta$ and discuss practical considerations (see Appendix~\ref{Hyper_ablation_beta}).Our comprehensive ablation experiments demonstrate the consistent performance of AROS across varying $\beta$ values.\\
Here, $d$ is the dimensionality of the feature vectors $r$, and $j = 1, \dots, K$. The terms $\hat{\mu}_j$ and $\hat{\Sigma}_j$ represent the mean vector and covariance matrix of the $j$-th class of ID training samples in feature space, respectively:

$\quad \quad \quad \quad \quad 
\hat{\mu}_j = \frac{1}{n_j} \sum_{i: y_i = j} f_{\theta}(x_i), \quad
\hat{\Sigma}_j = \frac{1}{n_j - 1} \sum_{i: y_i = j} (f_{\theta}(x_i) - \hat{\mu}_j)(f_{\theta}(x_i) - \hat{\mu}_j)^{T},\quad  \quad (2)
$
where $n_j$ is the number of samples in class $j$. By sampling equally across each class of \( \mathcal{D}^{\text{in}}_{\text{train}} \), we generate a set of synthetic, ``fake'' OOD embeddings (Figure~\ref{fig:example_figure}C), denoted as \( r_{\text{OOD}} \). We then construct a balanced training set by taking the union of the embeddings of ID samples and the OOD embeddings, defining it as:
$X_{\text{train}} = \left\{  f_\theta(\mathcal{D}^{\text{in}}_{\text{train}}) \cup r_{\text{OOD}}  \right\}. \label{D_train_define}$
We define the labels \( y \) for this set as 0 for ID and 1 for fake OOD embeddings.


\subsection{Lyapunov Stability for Robust OOD Detection}

As mentioned, several approaches have been proposed to apply Lyapunov's theorem to deep networks in practice, including methods such as LyaDEQ \cite{chu2024lyapunov}, ASODE \cite{li2022defending}, and SODEF \cite{kang2021stable}. Here, we utilize their framework to define the objective function and also benchmark our approach to these baselines. Amongst them, SODEF adopts a time-invariant \cite{kang2021stable,yan2019robustness} assumption, which makes stability analysis more practical, as the behavior of the neural ODE depends solely on the state \( z(t) \), independent of the specific time that the state is reached. This assumption implies that the equilibrium points of the NODE remain constant over time, facilitating a more tractable analysis of how perturbations evolve around these points \cite{yan2019robustness,massaroli2020dissecting,kang2021stable}. This is supported by our experiments in Table \ref{Table4.a:lyapunov_CLS}, which highlight SODEF's superior robustness. Consequently, we adopt the time-invariant framework and use their approach to define the loss function. In order to gain intuition for our approach, we provide the basic mathematical overview of how we leverage the Lyapunov theorems. In this study, as a practical consideration, we assume that the networks utilized have continuous first derivatives with respect to the input \( z(0)\), which has been shown to be a reasonable assumption \cite{hornik1991approximation}.

For a given dynamic system $\frac{dz(t)}{dt} = h_{\phi}(z(t))$, a state $z^\star$ is an equilibrium point of system if $z^\star$ satisfies $h(z^\star) = 0$.  An equilibrium point is stable if the trajectories starting near $z^\star$ remain around it all the time. More formally:

\textbf{Definition 1:} \label{Definition1}(Lyapunov stability \cite{chen1984linear}).\ An equilibrium $z^\star$ is said to be stable in the sense of Lyapunov if, for every $\varepsilon > 0$, there exists $\delta > 0$ such that, if $\|z(0) - z^\star\| < \delta$, then $\|z(t) - z^\star\| < \varepsilon$ for all $t \geq 0$. If $z^\star$ is stable, and 
$
\lim_{t \to \infty} \|z(t) - z^\star\| = 0,
$ 
$z^\star$ is said to be asymptotically stable.

\textbf{Theorem 1:}\label{Theorem 1} (Hartman–Grobman Theorem \cite{arrowsmith1995differential}). \textit{Consider a time-invariant system with continuous first derivatives, represented by  $\frac{d\mathbf{z}(t)}{dt} = h(\mathbf{z}(t))$.
For a fixed point \( \mathbf{z}^* \), if the Jacobian matrix \( \nabla h \) evaluated at \( \mathbf{z}^* \) has no eigenvalues with a real part equal to zero, the behavior of the original nonlinear dynamical system can be analyzed by studying the linearization of the system around this fixed point. The linearized system is given by $\frac{d\mathbf{z}'(t)}{dt} = \mathbf{A} \mathbf{z}'(t),$
where \( \mathbf{A} \) is the Jacobian matrix evaluated at \( \mathbf{z}^* \). This allows for a simplified analysis of the local dynamics in the vicinity of \( \mathbf{z}^* \). } 

\textbf{Theorem 2:}\label{Theorem 2} (Lyapunov Stability Theorem \cite{chen1984linear}) \textit{The equation 
$\frac{d\mathbf{z}'(t)}{dt} = \mathbf{A} \mathbf{z}'(t),$
is  asymptotically stable  if and only if all eigenvalues of \(\mathbf{A}\) have negative real parts.} 

\textbf{Theorem 3:}\label{Theorem 3} (Levy–Desplanques Theorem \cite{horn2012matrix}) \textit{Let $ A = [a_{ij}]$ be an $n$-dimensional square matrix and suppose it is strictly diagonally dominant, i.e., $ |a_{ii}| \geq \sum_{i \neq j} |a_{ij}| \ \text{and} \ a_{ii} \leq 0$ for all $i$. Then every eigenvalue of \( A \) has a negative real part.}

\textit{Definition 1} introduces the concept of asymptotic stability. Building on this, \textit{Theorem 1} demonstrates that the behavior of a nonlinear, time-invariant system near a fixed point can be effectively analyzed through its linearization. \textit{Theorem 2} then establishes a key condition for the asymptotic stability of linear systems: all eigenvalues of the system matrix must have negative real parts. To facilitate the verification of this stability condition, \textit{Theorem 3} provides a practical criterion based on the matrix's eigenvalues. In the subsequent section, we will introduce an objective function designed to adhere to these stability criteria.

\subsection{Orthogonal Binary Layer and Training Step}

We propose incorporating an orthogonal binary layer \cite{li2019orthogonal} denoted as $ B_{\eta}$ after the NODE $h_{\phi}$ in our pipeline to maximize the distance between the equilibrium points of ID and OOD data. Intuitively, this layer prevents the misalignment of convergence between perturbed OOD data and ID data by maximizing the distance between their equilibrium points. Given the output $z$ from the $h_{\phi}$, the orthogonal binary layer $B_{\eta}$ applies a transformation using weights $w$ such that $w^T w = I$,  ensuring orthogonality. Although Lyapunov stability encourages perturbed inputs to converge to neighborhoods of their unperturbed counterparts, the infinite-depth nature of NODE \cite{li2021future} makes them susceptible to degraded activations due to exploding or vanishing gradients \cite{pascanu2012understanding}. The introduction of an orthogonal layer mitigates this risk. Moreover, encouraging orthogonality within neural networks has demonstrated multiple benefits, such as preserving gradient norms and enforcing low Lipschitz constants—both of which contribute to enhanced robustness \cite{pauli2021training,huang2021training,zuhlke2024adversarial}.

To satisfy the aforementioned conditions, we optimize the following empirical Lagrangian $\mathcal{L}_{\text{SL}}$ with training data $(X_{\text{train}}, y)$:
\begin{align}
\mathcal{L}_{\text{SL}}=\min_{\phi,\eta} \frac{1}{\left|  X_{\text{train}} \right|}   \Bigg( & \ell_{\text{CE}}( B_{\eta}(h_{\phi}(X_{\text{train}})) , y)  + \gamma_1 \| h_{\phi}(X_{\text{train}})\|_2 \notag + \gamma_2 \text{exp}\left( -\sum_{i=1}^n [\nabla h_{\phi}(X_{\text{train}})]_{ii} \right) \notag \\
& + \gamma_3 \text{exp}\left( \sum_{i=1}^n \left( -|[\nabla h_{\phi}(X_{\text{train}})]_{ii}| + \sum_{j \neq i} |[\nabla h_{\phi}(X_{\text{train}})]_{ij}| \right) \right) \Bigg) \tag{3}
\end{align}
Note that here, $X_{\text{train}}$ serves as the initial hidden state, i.e., $z(0)$, for the NODE layer. The first term, $\ell_{\text{CE}}$, is a cross-entropy loss function. The second term forces $z(0)$ to be near the equilibrium points, while the remaining terms ensure strictly diagonally dominant derivatives, as described in Theorem 3. The $\text{exp}(.)$ function is selected as a monotonically increasing function with a minimum bound to limit the unbounded influence of the two regularizers, preventing them from dominating the loss.  We set $\gamma_1 = 1$ to balance the first regularization term with $\ell_{\text{CE}}$, and $\gamma_2 = \gamma_3 = 0.05$ to assign small, equal values that effectively enforce stability without overpowering the other terms. By setting $\gamma_2$ and $\gamma_3$ equal, we ensure that both stability conditions contribute equally. Details of the ablation study on these hyperparameters, along with other training step specifics, are provided in Appendices~\ref{gamma_ablation} and \ref{appendix:evaluation_and_experimental_setup}. By optimizing this objective function, the model learns Lyapunov-stable representations where ID and OOD equilibrium points are well-separated in the feature space after the NODE. The $B_{\eta}$ captures the probability distribution over the binary classes (ID vs. fake OOD), and for the OOD score of an input $x$, we use its probability assigned to the OOD class (Figure~\ref{fig:example_figure}D).



\begin{table*}[!htb]
    \centering
    \caption{ \textit{Table 1.} Performance of OOD detection methods under clean evaluation, random corruption (Gaussian noise), and PGD ($l_{\infty}$) adversarial attack with 1000 steps and $\frac{8}{255}$, as well as AutoAttack and Adaptive AutoAttack (AA), measured by AUROC (\%). A clean evaluation is one where no attack is made on the data. For corruption evaluation, Gaussian noise from the ImageNet-C~\cite{hendrycks2019benchmarking} benchmark was used. The best results are highlighted in \textbf{bold}, and the second-best results are \underline{underlined} in each row.}          
\label{Table1:Cifar_OOD}
\footnotesize{  \scriptsize $^\text{\textdagger}$ These methods leveraged auxiliary datasets and these $^*$ used large pretrained models as part of their pipeline.}

    \resizebox{\linewidth}{!}{
    
    \begin{tabular}{c}
    
        \begin{tabular}{lll*{12}{c}} 
            \specialrule{1.5pt}{\aboverulesep}{\belowrulesep}
            \noalign{\smallskip}
            \multicolumn{2}{c}{Dataset} & Attack& & & \multicolumn{4}{c}{Method} \\      
            \cmidrule(lr){1-2}\cmidrule(lr){3-3}   \cmidrule(lr){4-14}  
            \multirow{2}{*}{$\mathcal{D}_{in}$} & \multirow{2}{*}{$\mathcal{D}_{out}$} && \textbf{VOS} & \textbf{DHM} &  \textbf{ CATEX}$^*$     & \textbf{CSI} &  \textbf{ATOM}$^\text{\textdagger}$   &  \textbf{ALOE}$^\text{\textdagger}$   &  \textbf{ATD}$^{\text{\textdagger}*}$   &  \textbf{RODEO}$^{\text{\textdagger}*}$  & \textbf{AROS}  \\
 

            & & &  \small{   {\tiny (ResNet)}}&\small{  {\tiny(WideResNet)}} & \tiny (CLIP-ViT) &\small{   {\tiny(ResNet)}} & \small{ {\tiny(DenseNet)}}& \small{  {\tiny(WideResNet)}}& \small{   {\tiny(WideResNet)}}&\text{ \small {\tiny(CLIP-ViT)}}& \small{  {\tiny(WideResNet)}}\\
            \noalign{\smallskip}
            \specialrule{1.5pt}{\aboverulesep}{\belowrulesep}
            \noalign{\smallskip}
            
            \multirow{6}{*}{\rotatebox[origin=c]{90}{\textbf{CIFAR10}}} & \multirow{6}{*}{\rotatebox[origin=c]{90}{\textbf{CIFAR100}}} &\textbf{Clean}& 87.9  &\textbf{100.0}  & 88.3 &  92.2 & \underline{94.2}  & 78.8 & 82.0  & 75.6  & 88.2  \\
            \cmidrule(lr){3-14} 
            &   &\textbf{\small   Corruption }& 56.2 & 57.7 & 60.4& 54.7 & 57.3 & 54.5 & \underline{59.2} & 58.6 & \textbf{84.3} \\
             \cmidrule(lr){3-14} 
            && \textbf{$ {\text{PGD}}^{1000}$}  &  4.2& 1.8& 0.8&  3.6 & 1.6& 16.1 & 37.1&  \underline{37.8}& \textbf{80.1} \\

             \cmidrule(lr){3-14} 
            
            && \textbf{AutoAttack}  &  0.0& 1.2& 0.0&  0.4 & 0.5& 14.8 & \underline{36.2}&  35.9& \textbf{78.9} \\

                      \cmidrule(lr){3-14}

            && \textbf{AdaptiveAA}  &  0.0&0.0& 1.7&0.0&0.0&11.5&\underline{34.8} & 32.3& \textbf{76.4}\\

            \specialrule{1.5pt}{\aboverulesep}{\belowrulesep}

            \multirow{6}{*}{\rotatebox[origin=c]{90}{\textbf{CIFAR100}}} & \multirow{6}{*}{\rotatebox[origin=c]{90}{\textbf{CIFAR10}}} &\textbf{Clean}& 71.3 &\textbf{100.0} & 85.1 &53.2 &\underline{87.5 } &43.6 &57.5  & 61.5 & 74.3  \\

            \cmidrule(lr){3-14} 
            
            &   &\textbf{\small  Corruption}& 53.8 &\underline{ 58.2 }& 57.4 & 50.1& 55.3 & 56.1 & 56.0 &54.9 & \textbf{71.8 }\\
            
             \cmidrule(lr){3-14} 
            
            && \textbf{$ {\text{PGD}}^{1000}$}  &  5.4&0.0& 4.0&2.8&2.0&1.3&12.1 & \underline{29.0}& \textbf{67.0 }\\

  \cmidrule(lr){3-14}

            && \textbf{AutoAttack}  &  2.6&0.0& 0.3&0.9&0.0&0.0&10.5 &\underline{ 28.3}& \textbf{66.5} \\

                      \cmidrule(lr){3-14}

            && \textbf{AdaptiveAA}  & 0.0&1.4& 0.0&0.0&0.0&0.2&9.4 & \underline{26.7}& \textbf{65.2}\\

                      \specialrule{1.5pt}{\aboverulesep}{\belowrulesep}

        \end{tabular}
    \end{tabular}
    }

\end{table*}

\begin{table*}[!htb]

\label{combined_table}
\resizebox{\linewidth}{!}{
\begin{tabular}{c}
\begin{subtable}{\linewidth}
\centering
\caption{ \textit{Table 2a.} Performance of OOD detection methods under clean evaluation and $\text{PGD}^{1000}(l_{\infty})$ measured by AUROC (\%). The perturbation budget $\epsilon$ is set to $\frac{8}{255}$ for low-resolution datasets and $\frac{4}{255}$ for high-resolution datasets. The table cells denote results in the `\gr{Clean}/$\text{PGD}^{1000}$ ' format.
 }\label{Table2.a:All_OOD_Result}
\resizebox{\linewidth}{!}{
\begin{tabular}{lll*{12}{c}} 
\specialrule{1.5pt}{\aboverulesep}{\belowrulesep}
\noalign{\smallskip}

\multicolumn{2}{c}{\textbf{Dataset}}&    & &\multicolumn{4}{c}{\textbf{Method}}  
     \\      
\cmidrule(lr){1-2} \cmidrule(lr){3-14}  

\multirow{2}{*}{ $\mathcal{D}_{in}$  } & \multirow{2}{*}{  $\mathcal{D}_{out}$  }&\multirow{2}{*}{ \textbf{VOS}  }&\multirow{2}{*}{\textbf{DHM} } &\multirow{2}{*}{\textbf{CATEX} }&\multirow{2}{*}{\textbf{CSI}}&\multirow{2}{*}{\textbf{ATOM}} & \multirow{2}{*}{\textbf{ALOE}} &\multirow{2}{*}{\textbf{ATD}} &\multirow{2}{*}{\textbf{RODEO}}& \textbf{AROS}    \\
  & & & &  & & & &  & & \small{(Ours)}  \\

\noalign{\smallskip}
\specialrule{1.5pt}{\aboverulesep}{\belowrulesep}

\noalign{\smallskip}

\multirow{7}{*}{\rotatebox[origin=c]{90}{\textbf{CIFAR-10}}}  &\multirow{1}{*}{\textbf{CIFAR-100}} & \gr{87.9/}4.2& \gr{\textbf{100.0}/}1.8& \gr{88.3/}0.8& \gr{92.2/}3.6 &\gr{\underline{94.2}/}1.6& \gr{78.8/}16.1 &\gr{82.0/}37.1& \gr{75.6/}\underline{37.8}& \gr{88.2/}\textbf{80.1 }
\\  

      \cmidrule(lr){2-2} \cmidrule(lr){3-3} 
    \cmidrule(lr){3-14}   

\noalign{\smallskip}

&\multirow{1}{*}{\textbf{SVHN}} & \gr{\underline{93.3}/}2.8&\gr{\textbf{100.0}/} 4.5&\gr{91.6/}2.3& \gr{97.4/}1.7&\gr{89.2/}4.7&\gr{83.5/}26.6 &\gr{87.9/}\underline{39.0} & \gr{83.0/}38.2&\gr{93.0/}\textbf{86.4 }
\\  

      \cmidrule(lr){2-2} \cmidrule(lr){3-3} 
    \cmidrule(lr){3-14}   

\noalign{\smallskip}

 &\multirow{1}{*}{\textbf{Places}} &\gr{89.7/}5.2& \gr{\textbf{99.6}/}0.0&\gr{90.4/} 4.7&\gr{93.6/}0.1&\gr{98.7/}5.6&\gr{85.1/}21.9&\gr{92.5/}59.8& \gr{\underline{96.2}/}\underline{70.2}&\gr{90.8/}\textbf{83.5} 
\\  

      \cmidrule(lr){2-2} \cmidrule(lr){3-3} 
    \cmidrule(lr){3-14}   

\noalign{\smallskip}

 &\multirow{1}{*}{\textbf{LSUN}}  &\gr{98.0/}7.3& \gr{\textbf{100.0}/}2.6&\gr{\underline{95.1}/}0.8&\gr{97.7/}0.0
 
 &\gr{99.1/}1.0&\gr{98.7/}50.7&\gr{96.0/}68.1 &\gr{99.0/}\textbf{ 85.1}&\gr{90.6/}\underline{ 82.4 }
\\  

      \cmidrule(lr){2-2} \cmidrule(lr){3-3} 
    \cmidrule(lr){3-14}   

\noalign{\smallskip}

 &\multirow{1}{*}{\textbf{iSUN}}  &\gr{94.6/}0.5& \gr{\underline{99.1}/}2.8&\gr{93.2/} 4.4&\gr{95.4/}3.6&\gr{\textbf{99.5}/}2.5&\gr{98.3/}49.5&\gr{94.8/}65.9&\gr{97.7/}\underline{78.7}&\gr{88.9/}\textbf{81.2 }
\\  

\specialrule{1.5pt}{\aboverulesep}{\belowrulesep}

 \multirow{7}{*}{\vspace{-5mm}\rotatebox[origin=c]{90}{\textbf{CIFAR-100}}}

&\multirow{1}{*}{\textbf{CIFAR-10}} &\gr{\underline{71.3}/}5.4&\gr{\textbf{100.0}/}2.6& \gr{85.1/} 4.0&\gr{53.2/}0.7&\gr{87.5/}2.0&\gr{43.6/}1.3&\gr{57.5/}12.1 & \gr{61.5/}\underline{29.0}& \gr{74.3/}\textbf{67.0}
\\  

      \cmidrule(lr){2-2} \cmidrule(lr){3-3} 
    \cmidrule(lr){3-14}   

\noalign{\smallskip}

&\multirow{1}{*}{\textbf{SVHN}} &\gr{92.6/}3.2& \gr{\textbf{100.0}/}0.8&\gr{\underline{94.6}/}5.7&\gr{90.5/}4.2&\gr{92.8/}5.3& \gr{74.0/}18.1& \gr{72.5/}27.6&\gr{76.9/}\underline{31.4}& \gr{81.5/}\textbf{70.6 }
\\  

      \cmidrule(lr){2-2} \cmidrule(lr){3-3} 
    \cmidrule(lr){3-14}   

\noalign{\smallskip}

 &\multirow{1}{*}{\textbf{Places}} &\gr{75.5/}0.0& \gr{\textbf{100.0}/}3.9&\gr{87.3/}1.4&\gr{73.6/}0.0&\gr{\underline{94.8}/}3.0&\gr{75.0/}12.4&\gr{83.3/}40.0&\gr{93.0/}\underline{66.6}&\gr{77.0/}\textbf{69.2}
\\  

      \cmidrule(lr){2-2} \cmidrule(lr){3-3} 
    \cmidrule(lr){3-14}   

\noalign{\smallskip}

 &\multirow{1}{*}{\textbf{LSUN}}  &\gr{92.9/}5.7&\gr{\textbf{100.0}/}1.6&\gr{94.0/}8.9&\gr{63.4/}1.8&\gr{96.6/}1.5&\gr{98.7/}50.7&\gr{96.0/}68.1&\gr{\underline{98.1}/}\underline{63.1}&\gr{74.3/}\textbf{68.1} 
\\  

      \cmidrule(lr){2-2} \cmidrule(lr){3-3} 
    \cmidrule(lr){3-14}   

\noalign{\smallskip}

 &\multirow{1}{*}{\textbf{iSUN}}  &\gr{70.2/}4.5&\gr{99.6/}3.6&\gr{81.2/}0.0&\gr{81.4/}3.0& \gr{96.4/}1.4&\gr{\textbf{98.3}/}49.5&\gr{\underline{94.8}/}\underline{65.9}&\gr{95.1/}65.6&\gr{72.8/}\textbf{67.9} 
\\

\specialrule{1.5pt}{\aboverulesep}{\belowrulesep}

\multirow{7}{*}{\vspace{-5mm}\rotatebox[origin=c]{90}{\textbf{ImageNet-1k}}} 

&\multirow{1}{*}{\textbf{Texture}} &\gr{\underline{86.7}/}0.8&\gr{82.4/}0.0&\gr{92.7/}0.0&\gr{85.8/}0.6&  \gr{\textbf{88.9}/}7.3 & \gr{76.2/}\underline{21.8}& \gr{74.2/}15.7& \gr{71.3/}19.4&\gr{78.3/}\textbf{69.2 } 
\\  

      \cmidrule(lr){2-2} \cmidrule(lr){3-3} 
    \cmidrule(lr){3-14}   

\noalign{\smallskip}

&\multirow{1}{*}{\textbf{iNaturalist}} &\gr{\underline{94.5}/}0.0&\gr{80.7/}0.0&\gr{\underline{97.9}/}2.0&\gr{85.2/}1.7& \gr{83.6/}10.5& \gr{78.9/}\underline{19.4} &\gr{72.5/}12.6&\gr{72.7/}15.0 &\gr{84.6/}\textbf{75.3}
\\  

      \cmidrule(lr){2-2} \cmidrule(lr){3-3} 
    \cmidrule(lr){3-14}   

\noalign{\smallskip}

 &\multirow{1}{*}{\textbf{Places}} &\gr{\underline{90.2}/}0.0&\gr{76.2/}0.4&\gr{\textbf{90.5}/}0.0&\gr{83.9/}0.2& \gr{84.5/}12.8 &\gr{78.6/}15.3&\gr{75.4/}17.5&\gr{69.2/}\underline{18.5}&\gr{76.2/}\textbf{68.1}
\\  

      \cmidrule(lr){2-2} \cmidrule(lr){3-3} 
    \cmidrule(lr){3-14}   

\noalign{\smallskip}

 &\multirow{1}{*}{\textbf{LSUN}}  &\gr{\underline{91.9}/}0.0&\gr{82.5/}0.0&\gr{\textbf{92.9}/}0.4&\gr{78.4/}1.9&\gr{85.3/}11.2&\gr{77.4/} \underline{16.9}&\gr{68.3/}15.1&\gr{70.4/}16.2&\gr{79.4/}\textbf{69.0}
\\  

      \cmidrule(lr){2-2} \cmidrule(lr){3-3} 
    \cmidrule(lr){3-14}   

\noalign{\smallskip}

 &\multirow{1}{*}{\textbf{iSUN}}  &\gr{\underline{92.8}/}2.7&\gr{81.6/}0.0&\gr{\textbf{93.7}/}0.0&\gr{77.5/}0.0&\gr{80.3/}14.1&\gr{75.3/}11.8&\gr{76.6/}15.8&\gr{72.8/}\underline{17.3}&\gr{80.3/}\textbf{71.6 }
\\

\specialrule{1.5pt}{\aboverulesep}{\belowrulesep}

 &\multirow{1}{*}{ \textit{Mean}}  &\gr{88.1/}2.8&\textbf{\gr{93.4}}\gr{/}1.6&\gr{91.2/}2.3&\gr{83.3/}1.5&\gr{\underline{91.4}/}5.6&\gr{81.4/}25.5&\gr{81.6/}37.4& \gr{82.1/}\underline{44.4}& \gr{82.0/}\textbf{74.0}
\\  
\specialrule{1.5pt}{\aboverulesep}{\belowrulesep}

\end{tabular}}
\end{subtable}

\\[10pt]

\begin{subtable}{\linewidth}
\centering
 \caption{ \textit{Table 2b.} Performance (\gr{Clean}/$\text{PGD}^{1000}$) of OOD detection methods under clean and $\text{PGD}^{1000}(l_{\infty})$, measured by AUROC (\%), on the OSR setup, which splits one dataset's classes randomly to create $\mathcal{D}_{in}$ and $\mathcal{D}_{out}$. \label{Table2.b:Open-Set Recognition}}

\resizebox{\linewidth}{!}{
\begin{tabular}{ll*{11}{c}} 
\specialrule{1.5pt}{\aboverulesep}{\belowrulesep}
\noalign{\smallskip}

\multicolumn{1}{c}{\textbf{Dataset}}&    & &\multicolumn{4}{c}{\textbf{Method}}  
     \\      
\cmidrule(lr){1-1} \cmidrule(lr){2-13}  

\multirow{2}{*}{    }&\multirow{2}{*}{\textbf{VOS}}&\multirow{2}{*}{\textbf{DHM}} &\multirow{2}{*}{\textbf{CATEX}}&\multirow{2}{*}{\textbf{CSI}}&\multirow{2}{*}{\textbf{ATOM}} & \multirow{2}{*}{\textbf{ALOE}} &\multirow{2}{*}{\textbf{ATD}} &\multirow{2}{*}{\textbf{RODEO}}& \textbf{AROS}    \\
  & & &  & & & & &  & \small{(Ours)}  \\

\noalign{\smallskip}
\specialrule{1.5pt}{\aboverulesep}{\belowrulesep}

\noalign{\smallskip}

\multirow{1}{*}{\textbf{MNIST}} &\gr{86.3/}4.8& \gr{92.6/}0.4&\gr{92.3/}1.9&\gr{93.6/}6.1&\gr{74.8/}4.1&\gr{79.5/}37.3&\gr{68.7/}56.5&\gr{\textbf{97.2}/}\underline{85.0}&\gr{\underline{94.4}/}\textbf{86.3 }
\\  

\cmidrule(lr){1-1} \cmidrule(lr){2-13}   

\noalign{\smallskip}

\multirow{1}{*}{\textbf{FMNIST}} &\gr{78.1/}2.0&\gr{85.9/}0.0&\gr{87.0/}0.4&\gr{84.6/}1.2&\gr{64.3/}4.2&\gr{72.6/}28.5 &\gr{59.6/}42.1&\gr{\textbf{87.7}/}\underline{65.3}& \gr{\underline{84.1}/}\textbf{72.6}
\\  

\cmidrule(lr){1-1} \cmidrule(lr){2-13}   

\noalign{\smallskip}

\multirow{1}{*}{\textbf{CIFAR-10}} &\gr{74.7/}0.0&\gr{\underline{90.8}/}0.0&\gr{\textbf{95.1}/}0.0&\gr{91.4/}0.6&\gr{68.3/}5.0&\gr{52.4/}25.6&\gr{49.0/}32.4 &\gr{79.6/}\underline{62.7}& \gr{78.8/}\textbf{69.5}
\\  

\cmidrule(lr){1-1} \cmidrule(lr){2-13}   

\noalign{\smallskip}

\multirow{1}{*}{\textbf{CIFAR-100}} &\gr{63.5/}0.0&\gr{78.6/}0.0
&\gr{\textbf{91.9}/}0.0&\gr{\underline{86.7}/}1.9&\gr{51.4/}2.6&\gr{49.8/}18.2&\gr{50.5/}\underline{36.1}&\gr{64.1/}35.3& \gr{67.0/}\textbf{58.2} 
\\  

\cmidrule(lr){1-1} \cmidrule(lr){2-13}   

\noalign{\smallskip}

\multirow{1}{*}{\textbf{Imagenette}}  &\gr{76.7/}0.0&\gr{84.2/}0.0&\gr{\textbf{96.4}/}1.6&\gr{\underline{92.8}/}0.0&\gr{63.5/}8.2&\gr{61.7/}14.2&\gr{63.8/}28.4&\gr{70.6/}\underline{39.4}&\gr{78.2/}\textbf{67.5}
\\ 
 \cmidrule(lr){1-1} \cmidrule(lr){2-13}

\multirow{1}{*}{\textbf{ADNI}} &\gr{73.5/}4.1&\gr{69.4/}5.2&\gr{\textbf{86.9}/}0.1&\gr{\underline{82.1}/}0.0&\gr{66.9/}2.3&\gr{64.0/}11.0&\gr{68.3/}\underline{33.9}&\gr{75.5/}24.6&\gr{80.9/}\textbf{61.7}
\\ 
\specialrule{1.5pt}{\aboverulesep}{\belowrulesep}

 \multicolumn{1}{c}{ \textit{Mean} } &\gr{75.5/}1.8& \gr{\underline{83.6}/}0.9 &\gr{\textbf{91.6}/}0.7&\gr{88.5/}1.6&\gr{64.9/}4.4&\gr{63.3/}22.5&\gr{60.0/}38.2&\gr{79.1/}\underline{52.1}&\gr{80.6/}\textbf{69.3}
\\

\specialrule{1.5pt}{\aboverulesep}{\belowrulesep}

\end{tabular}}
\end{subtable}
\end{tabular}}
\end{table*}

\begin{table*}[!htb]
    \centering
\caption{\textit{Table 3.} Performance of OOD detection methods under various types of non-adversarial perturbations, referred to as image corruptions, as introduced in the CIFAR-10-C and CIFAR-100-C datasets \cite{hendrycks2019benchmarking}, measured by AUROC (\%). Specifically, test inputs, including both ID and OOD, are perturbed with a particular corruption in each experiment.}
    \label{Corruption_table} 
    \resizebox{\linewidth}{!}{
                \renewcommand{\arraystretch}{0.8}
    \begin{tabular}{c}
    
        \begin{tabular}{lll*{19}{c}} 
            \specialrule{1.5pt}{\aboverulesep}{\belowrulesep}
            \noalign{\smallskip}
            \multicolumn{2}{c}{Dataset} & Methods& \multicolumn{14}{c}{Corruption} &&\multirow{3}{*}{ \textit{Mean}  }  \\      
            \cmidrule(lr){1-2}\cmidrule(lr){3-3}   \cmidrule(lr){4-18}  
            \multirow{2}{*}{$\mathcal{D}_{in}$} & \multirow{2}{*}{$\mathcal{D}_{out}$} && \multirow{2}{*}{\small \textbf{Gauss.} } & \multirow{2}{*}{\small \textbf{Shot}} & \multirow{2}{*}{\small \textbf{Impulse}} & \multirow{2}{*}{\small \textbf{Defocus}} & \multirow{2}{*}{\small \textbf{Glass}} & \multirow{2}{*}{\small \textbf{Motion}} & \multirow{2}{*}{\small \textbf{Zoom}} & \multirow{2}{*}{\small \textbf{Snow}} & \multirow{2}{*}{\small \textbf{Frost}}  & \multirow{2}{*}{\small \textbf{Fog}} & \multirow{2}{*}{\small \textbf{Bright}} & \multirow{2}{*}{\small \textbf{Contrast}} & \multirow{2}{*}{\small \textbf{Elastic}} & \multirow{2}{*}{\small \textbf{Pixel}} & \multirow{2}{*}{\small \textbf{JPEG}}
  \\
            & & & & & & & & & &&   &&& &&&&  \\
            \noalign{\smallskip}
            \specialrule{1.5pt}{\aboverulesep}{\belowrulesep}
            \noalign{\smallskip}
            
            \multirow{12}{*}{\rotatebox[origin=c]{90}{\textbf{\small CIFAR-10-C}}} & \multirow{12}{*}{\rotatebox[origin=c]{90}{\textbf{\small CIFAR-100-C}}}

&\textbf{\small VOS}\cellcolor{lightgray} &56.2\cellcolor{lightgray}&67.5\cellcolor{lightgray}&\underline{76.5}\cellcolor{lightgray}&77.7\cellcolor{lightgray}&73.9\cellcolor{lightgray}&\underline{78.7}\cellcolor{lightgray}&76.3\cellcolor{lightgray}&72.0\cellcolor{lightgray}&54.1\cellcolor{lightgray}&\underline{77.0}\cellcolor{lightgray}&58.5\cellcolor{lightgray}&\underline{79.1}\cellcolor{lightgray}&\textbf{81.2}\cellcolor{lightgray}&\textbf{83.6}\cellcolor{lightgray}&74.4\cellcolor{lightgray}&72.5\cellcolor{lightgray}\\

\cmidrule(lr){3-18} 
\vspace{-0.5mm}
&   &\textbf{\small DHM} &  57.7& 78.7& 72.4& 75.4& 75.6& 73.9& 77.5 & 75.8& 70.8& 56.8& 74.5& 58.0& 77.4& 78.4& 80.6  & 72.2  \\

\cmidrule(lr){3-18} 

&& \textbf{\small CATEX}\cellcolor{lightgray}   &\underline{62.4}\cellcolor{lightgray}& \textbf{80.9}\cellcolor{lightgray}&73.0\cellcolor{lightgray}& \underline{78.4}\cellcolor{lightgray}&76.3\cellcolor{lightgray}& 78.6\cellcolor{lightgray}&\textbf{81.3}\cellcolor{lightgray}& \underline{79.9}\cellcolor{lightgray}& \underline{78.9}\cellcolor{lightgray}& 58.3\cellcolor{lightgray}& 80.0\cellcolor{lightgray}& 54.0\cellcolor{lightgray}& 79.0\cellcolor{lightgray}& 80.5\cellcolor{lightgray}&82.4\cellcolor{lightgray}&\underline{74.9}\cellcolor{lightgray} \\

\cmidrule(lr){3-18} 

&& \textbf{\small CSI}   &54.7&58.0&58.7&62.9&61.7&69.0&65.9&77.2&69.2&74.8&\textbf{91.9}&65.8&74.2&62.6&74.9&68.1\\

\cmidrule(lr){3-18} 

&& \textbf{\small ATOM}\cellcolor{lightgray}   &57.3\cellcolor{lightgray}&75.5\cellcolor{lightgray}&63.6\cellcolor{lightgray}&70.7\cellcolor{lightgray}&72.2\cellcolor{lightgray}&69.9\cellcolor{lightgray}&74.6\cellcolor{lightgray}&77.2\cellcolor{lightgray}&76.5\cellcolor{lightgray}&55.3\cellcolor{lightgray}&80.5\cellcolor{lightgray}&54.1\cellcolor{lightgray}&74.7\cellcolor{lightgray}&77.4\cellcolor{lightgray}&80.8\cellcolor{lightgray}&70.7\cellcolor{lightgray}\\

\cmidrule(lr){3-18} 

&& \textbf{\small ALOE}   &54.5&76.4&64.0&71.5&73.0&70.9&75.5&78.2&77.9&56.3&81.5&54.0&76.9&79.3&82.1&71.5\\

\cmidrule(lr){3-18} 

&& \textbf{\small ATD}\cellcolor{lightgray}  &59.2\cellcolor{lightgray}&\underline{79.2}\cellcolor{lightgray}&71.0\cellcolor{lightgray}&76.7\cellcolor{lightgray}&\underline{76.9}\cellcolor{lightgray}&75.6\cellcolor{lightgray}&\underline{79.5}\cellcolor{lightgray}&78.2\cellcolor{lightgray}&74.9\cellcolor{lightgray}&59.5\cellcolor{lightgray}&77.8\cellcolor{lightgray}&59.5\cellcolor{lightgray}&79.0\cellcolor{lightgray}&\underline{80.8}\cellcolor{lightgray}&\textbf{82.9}\cellcolor{lightgray}&73.7\cellcolor{lightgray}\\

\cmidrule(lr){3-18} 

&& \textbf{\small RODEO}   &58.6&76.0&68.5&73.5&73.8&72.1&75.5&74.5&70.9&57.8&74.5&57.7&75.3&76.8&79.5&71.0\\

\cmidrule(lr){3-18} 
&& \textbf{\small AROS}\cellcolor{lightgray}  & \textbf{84.3}\cellcolor{lightgray}& 76.5\cellcolor{lightgray}& \textbf{79.2}\cellcolor{lightgray}& \textbf{83.8}\cellcolor{lightgray}& \textbf{77.3}\cellcolor{lightgray}& \textbf{82.0}\cellcolor{lightgray}& \textbf{81.3}\cellcolor{lightgray}&\textbf{83.4}\cellcolor{lightgray}& \textbf{84.0}\cellcolor{lightgray}& \textbf{84.0}\cellcolor{lightgray}& \underline{84.7}\cellcolor{lightgray}& \textbf{83.3}\cellcolor{lightgray}& \underline{80.7}\cellcolor{lightgray}&79.6\cellcolor{lightgray}   &\underline{82.5}\cellcolor{lightgray}&\textbf{81.8}\cellcolor{lightgray}\\

                \specialrule{1.5pt}{\aboverulesep}{\belowrulesep}

                \multirow{12}{*}{\rotatebox[origin=c]{90}{\textbf{\small CIFAR-100-C}}} & \multirow{12}{*}{\rotatebox[origin=c]{90}{\textbf{\small CIFAR-10-C}}} &\textbf{\small VOS}\cellcolor{lightgray} &  53.8\cellcolor{lightgray}& 55.7\cellcolor{lightgray}&  65.6\cellcolor{lightgray}&  \underline{58.2}\cellcolor{lightgray}&  \cellcolor{lightgray}47.1& \cellcolor{lightgray} 51.4& \cellcolor{lightgray} 57.6& \cellcolor{lightgray} 53.9& \cellcolor{lightgray} 59.0 & \cellcolor{lightgray} 57.2&\cellcolor{lightgray}  56.5& \cellcolor{lightgray} 54.8& \cellcolor{lightgray} 48.2& \cellcolor{lightgray} \underline{59.4}& \cellcolor{lightgray} 51.1&\cellcolor{lightgray}55.3\\
                    
                \cmidrule(lr){3-18} 
                \vspace{-0.5mm}
                &   &\textbf{\small DHM} & \underline{58.2} &59.9& 64.0 &57.7& 48.9&58.0 & 57.4& 57.6& 58.5& 57.9& 58.1& 58.3&49.8& 55.6& \underline{56.7} & 57.1\\
                    
                \cmidrule(lr){3-18} 
                    
                && \textbf{\small CATEX}\cellcolor{lightgray}   &  57.4\cellcolor{lightgray}& \underline{60.2}\cellcolor{lightgray}& \underline{65.7}\cellcolor{lightgray}& \textbf{59.6}\cellcolor{lightgray}& \underline{64.9}\cellcolor{lightgray}& \underline{62.9}\cellcolor{lightgray}&\underline{59.3}\cellcolor{lightgray}&\underline{67.5}\cellcolor{lightgray}& \underline{61.4}\cellcolor{lightgray}& \textbf{59.8}\cellcolor{lightgray}& \underline{60.0}\cellcolor{lightgray}& \textbf{64.2}\cellcolor{lightgray}& \underline{56.8}\cellcolor{lightgray}& 57.5\cellcolor{lightgray}& \textbf{58.6}\cellcolor{lightgray}  &\underline{61.0}\cellcolor{lightgray}\\
                    
                \cmidrule(lr){3-18} 
                    
                && \textbf{\small CSI}   &  50.1& 48.8& 50.6& 47.8&47.5& 46.9& 46.8& 50.6& 50.3& 51.8& 49.9&   52.2& 42.9& 48.0& 47.7 &48.8\\

                \cmidrule(lr){3-18} 
                    
                && \textbf{\small ATOM}\cellcolor{lightgray}   &  55.3\cellcolor{lightgray}& 51.2\cellcolor{lightgray}& 53.1\cellcolor{lightgray}&50.2\cellcolor{lightgray}&49.9\cellcolor{lightgray}& 49.2\cellcolor{lightgray}&49.6\cellcolor{lightgray}& 53.1\cellcolor{lightgray}& 52.8\cellcolor{lightgray}& 54.4\cellcolor{lightgray}& 52.4\cellcolor{lightgray}& 54.8\cellcolor{lightgray}&45.0\cellcolor{lightgray}& 50.4\cellcolor{lightgray}& 50.8\cellcolor{lightgray}&51.5\cellcolor{lightgray}\\

                \cmidrule(lr){3-18} 
                    
                && \textbf{\small ALOE}   & 56.1& 53.4&62.8&54.5&51.8&54.9& 54.1& 54.4& 55.6&54.8& 52.7& 56.4& 47.8& 51.7&53.2&54.3\\

                \cmidrule(lr){3-18} 
                    
                && \textbf{\small ATD}\cellcolor{lightgray}  &  56.0\cellcolor{lightgray}& 57.4\cellcolor{lightgray}& 61.5\cellcolor{lightgray}& 57.5\cellcolor{lightgray}&44.8\cellcolor{lightgray}& 57.1\cellcolor{lightgray}&54.2\cellcolor{lightgray}& 56.9\cellcolor{lightgray}& 58.3\cellcolor{lightgray}& 55.2\cellcolor{lightgray}&53.7\cellcolor{lightgray}& 57.5\cellcolor{lightgray}& 49.3\cellcolor{lightgray}&50.8\cellcolor{lightgray}& 56.0\cellcolor{lightgray}&55.1\cellcolor{lightgray}\\

                \cmidrule(lr){3-18} 
                    
                && \textbf{\small RODEO}   &  54.9& 58.1&60.6& 56.4& 51.0&60.5&58.9&58.4&57.9&54.6& 57.4&52.3& 52.7&53.5& 51.2&55.9\\
                \cmidrule(lr){3-18} 
                    
                && \textbf{\small AROS}\cellcolor{lightgray}  &  \textbf{71.8}\cellcolor{lightgray}&  \textbf{74.8}\cellcolor{lightgray}& \textbf{67.7}\cellcolor{lightgray}&  \textbf{59.6}\cellcolor{lightgray}&  \textbf{72.6}\cellcolor{lightgray}& \textbf{73.9}\cellcolor{lightgray}&  \textbf{65.7}\cellcolor{lightgray}& \textbf{68.5}\cellcolor{lightgray}& \textbf{64.4}\cellcolor{lightgray}&\textbf{59.8}\cellcolor{lightgray}& \textbf{75.0}\cellcolor{lightgray}&  \textbf{64.2}\cellcolor{lightgray}&  \textbf{72.8}\cellcolor{lightgray}& \textbf{69.5}\cellcolor{lightgray}& \textbf{58.6}\cellcolor{lightgray} &\textbf{67.9}\cellcolor{lightgray}\\

                \specialrule{1.5pt}{\aboverulesep}{\belowrulesep}

            \end{tabular}
        \end{tabular}
    }
\end{table*}
\begin{table*}[!ht]
    \centering
    \begin{subtable}{.49\linewidth}
        \caption{\textit{Table 4a.} Comparison of post-hoc OOD detection methods using different classifiers trained with various strategies and evaluated with multiple scoring functions. The comparison (\gr{Clean}/$\text{PGD}^{1000}$) is conducted under clean and $\text{PGD}^{1000}$ conditions, measured by AUROC (\%).}\label{Table4.a:lyapunov_CLS}
        \centering
        \resizebox{\linewidth}{!}{%
         \begin{tabular}{l*{6}{c}}
            \toprule
            \multirow{3}{*}{Classifier} & \multirow{2}{*}{Posthoc}   &\multicolumn{2}{c}{\textbf{CIFAR-10}}& \multicolumn{2}{c}{\textbf{CIFAR-100}}\\
            \cmidrule(lr){3-4} \cmidrule(lr){5-6}
            &Method&  \textbf{CIFAR-100} & \textbf{SVHN} &  \textbf{CIFAR-10} & \textbf{SVHN}\\
            \midrule   

             &  MSP&   \gr{87.9/}0.0 &   \gr{91.8/}1.4 &  \gr{75.4/}0.2 &  \gr{71.4/}3.6 \\
       Standard & MD &\gr{\textbf{88.5}/}4.3 & \gr{\textbf{99.1}/}0.6  & \gr{75.0/}1.9 & \gr{\textbf{98.4}/}0.6 \\
            &  OpenMax& \gr{86.4/}0.0 &  \gr{\underline{94.7}/}2.8&  \gr{\underline{77.6}/}0.0 & \gr{\underline{93.9}/}4.2 \\
       \midrule  
             &  MSP& \gr{79.3/}16.0& \gr{85.1/}19.7 &  \gr{67.2/}10.7 &  \gr{74.6/}11.3 \\
       AT & MD&\gr{81.4/}25.6 & \gr{88.2/}27.5 & \gr{71.8/}15.0& \gr{81.5/}19.7 \\
            &  OpenMax&  \gr{82.4/}27.8 &  \gr{86.5/}26.9 &  \gr{\textbf{80.0}/}16.4 &  \gr{75.4/}22.9 \\
       \midrule   
             &  MSP& \gr{84.2/}10.6 &  \gr{89.3/}15.4 & \gr{69.7/}12.5 & \gr{76.1/}23.8\\
       ODENet & MD&\gr{80.7/}9.1& \gr{84.6/}13.0 & \gr{66.4/}14.8 & \gr{72.9/}16.4 \\
            &  OpenMax& \gr{83.8/}14.2 &  \gr{87.4/}20.9 &  \gr{70.3/}15.6 & \gr{75.6/}18.2  \\

                   \midrule

& MSP & \gr{77.5/}56.5 & \gr{83.7/}58.5 & \gr{69.1/}48.0 & \gr{69.4/}53.3 \\
LyaDEQ & MD & \gr{79.1/}56.9 & \gr{82.0/}56.5 & \gr{60.3/} 53.4  & \gr{69.3/}54.2 \\
& OpenMax & \gr{76.0/}47.4 & \gr{77.5/}56.5 & \gr{67.8/}57.1 & \gr{73.3/} 58.0  \\
                   \midrule  

& MSP & \gr{76.3/}56.3 & \gr{80.5/}62.5 & \gr{64.6/}44.9 & \gr{64.6/}58.9 \\
ASODE & MD & \gr{74.9/}49.5 & \gr{76.1/}54.4 & \gr{59.3/}\ 52.0  & \gr{72.1/}55.1 \\
& OpenMax & \gr{72.6/}44.2 & \gr{75.9/}57.9 & \gr{66.1/}52.1 & \gr{80.5/} 50.4  \\

                   \midrule  
                         &  MSP & \gr{83.5/}61.9 &  \gr{86.4/}65.3 &  \gr{67.2/}53.1 & \gr{73.7/}60.4 \\
       SODEF & MD& \gr{75.4/}57.7 & \gr{81.9/}64.2& \gr{65.8/}\underline{58.4} & \gr{71.8/}62.5 \\
            &  OpenMax& \gr{82.8/}\underline{65.3} & \gr{86.4/}\underline{69.1} & \gr{66.3/}56.6 & \gr{75.2/}\underline{64.9 }\\

\cdashline{1-6}
                   \noalign{\vskip 5.5pt}

         AROS   &N/A& \gr{\underline{88.2}/}\textbf{80.1} &  \gr{93.0/}\textbf{86.4}&  \gr{74.3/}\textbf{67.0 }& \gr{82.5/}\textbf{70.6}

\\

            \bottomrule
        \end{tabular}
        }
     \end{subtable}%
    \hfill
    \begin{subtable}{.49\linewidth}
        \caption{\textit{Table 4b.} Performance of OOD detection methods under clean and $\text{PGD}^{1000}$, measured by AUPR$\uparrow$ (\%) and FPR95 $\downarrow$(\%) metrics. The perturbation budget $\epsilon$ is set to $\frac{8}{255}$. The table cells present results in the `\gr{Clean}/$\text{PGD}^{1000}$' format.\label{Table4.b:various_metrics}}
        \centering
        \resizebox{\linewidth}{!}{%
            \renewcommand{\arraystretch}{0.8}
         \begin{tabular}{l*{6}{c}}
            \toprule
            \multirow{3}{*}{Method} & \multirow{3}{*}{Metric}   &\multicolumn{2}{c}{\textbf{CIFAR-10}}& \multicolumn{2}{c}{\textbf{CIFAR-100}}\\
            \cmidrule(lr){3-4} \cmidrule(lr){5-6}
            &&  \textbf{CIFAR-100} & \textbf{SVHN} &  \textbf{CIFAR-10} & \textbf{SVHN}\\
            \midrule                    \noalign{\vskip 5.5pt}

           \multirow{2}{*}{VOS}  & {\small AUPR$\uparrow$} & \gr{85.8/}0.0 &  \gr{90.4/}6.2 &  \gr{75.8/}0.0 & \gr{93.9/}7.6\\
        & {\small FPR95$\downarrow$} &\gr{35.2/}100.0 & \gr{38.2/}99.8 & \gr{48.7/}100.0 & \gr{41.5/}98.2  \\ 
             \midrule  

           \multirow{2}{*}{DHM}  & {\small AUPR$\uparrow$} & \gr{\textbf{100.0}/}0.3 &  \gr{\textbf{100.0}/}4.8 &  \gr{\textbf{100.0}/}0.0 &  \gr{\textbf{100.0}/}3.2 \\
        & {\small FPR95$\downarrow$} &\gr{\textbf{0.2}/}99.2 & \gr{\textbf{0.0}/}98.5 & \gr{\textbf{1.1}/}100.0 & \gr{\textbf{0.4}/}99.7  \\ 
             \midrule

           \multirow{2}{*}{CATEX}  & {\small AUPR$\uparrow$} & \gr{89.5/}0.4&  \gr{93.1/}7.6 &  \gr{84.2/}0.0 &  \gr{\underline{96.6}/}1.3 \\
        & {\small FPR95$\downarrow$} &\gr{36.6/}99.1 & \gr{27.3/}95.6 & \gr{42.8/}100.0 & \gr{37.1/}98.4  \\ 
             \midrule
           \multirow{2}{*}{CSI}  & {\small AUPR$\uparrow$} &  \gr{93.4/}0.0&  \gr{98.2/}4.6 &  \gr{65.8/}0.0 &  \gr{82.9/}0.4\\
        & {\small FPR95$\downarrow$} &\gr{40.6/}100.0 & \gr{37.4/}99.1 & \gr{65.2/}100.0 & \gr{42.6/}97.5  \\ 
             \midrule
           \multirow{2}{*}{ATOM}  & {\small AUPR$\uparrow$} & \gr{\underline{97.9}/}5.8 &  \gr{\underline{98.3}/}11.6 &  \gr{\underline{89.3}/}5.1 &  \gr{94.6/}7.2 \\
        & {\small FPR95$\downarrow$} &\gr{\underline{24.0}/}96.4 & \gr{\underline{12.7}/}93.1 & \gr{\underline{38.6}/}98.0 & \gr{\underline{29.2}/}97.9  \\ 
             \midrule
           \multirow{2}{*}{ALOE}  & {\small AUPR$\uparrow$} & \gr{80.4/}21.7&  \gr{86.5/}27.3 &  \gr{54.8/}9.2 &  \gr{85.1/}18.6 \\
        & {\small FPR95$\downarrow$} &\gr{38.6/}89.2 & \gr{45.1/}93.7 & \gr{72.8/}96.1 & \gr{57.4/}84.8  \\ 
             \midrule
           \multirow{2}{*}{ATD}  & {\small AUPR$\uparrow$} & \gr{81.9/}44.6 &  \gr{85.3/}\underline{53.7} &  \gr{61.4/}\underline{27.2} & \gr{68.3/}26.1\\
        & {\small FPR95$\downarrow$} &\gr{47.3/}86.2 & \gr{42.4/}83.0 & \gr{68.2/}94.8 & \gr{59.0/}91.9  \\ 
             \midrule 
           \multirow{2}{*}{RODEO}  & {\small AUPR$\uparrow$} & \gr{83.5/}\underline{47.0}&  \gr{88.2/}51.6 &  \gr{72.8/}26.5 & \gr{81.7/}\underline{42.9}\\
        & {\small FPR95$\downarrow$} &\gr{42.9/}\underline{81.3} & \gr{49.6/}\underline{75.4} & \gr{ 65.3/}\underline{89.0 }& \gr{61.8/}\underline{83.5}  \\ 
        
                   \noalign{\vskip 5.5pt}

\cdashline{1-6}
                   \noalign{\vskip 5.5pt}
           \multirow{2}{*}{AROS}  & {\small AUPR$\uparrow$} & \gr{87.2/}\textbf{80.5} &  \gr{97.2/}\textbf{91.4} &  \gr{71.0/}\textbf{65.3} &  \gr{72.4/}\textbf{66.8} \\
        & {\small FPR95$\downarrow$} &\gr{39.3/}\textbf{45.2} & \gr{15.5/}\textbf{27.0 }& \gr{54.2/}\textbf{67.8} & \gr{46.3/}\textbf{62.7}  \\

            \bottomrule
        \end{tabular}
        }
    \end{subtable}
 \end{table*}

\begin{table*}[t]
    \centering
    \caption{\textit{Table 5.}  
An ablation study (\gr{Clean}/$\text{PGD}^{1000}$), measured by AUROC (\%), on our method with the exclusion of different components while keeping all others intact. The left side is the configurations.}     \label{table5:ablation}

    \vspace{-7pt}
    \resizebox{ \linewidth}{!}
    {\begin{tabular}{@{}ccccccc cc ccccc} 

    \specialrule{1.5pt}{\aboverulesep}{\belowrulesep}
    \multirow{3}{*}{\textbf{Config}} & \multicolumn{5}{c}{\textbf{Components}} & &  & \multicolumn{2}{c}{\textbf{CIFAR10}} & \multicolumn{2}{c}{\textbf{CIFAR100}}&\multicolumn{2}{c}{\textbf{ImageNet-1k}}\\
      \cmidrule(lr){2-7} \cmidrule(lr){9-10}\cmidrule(lr){11-12}\cmidrule(lr){13-14}

    & Adv. Trained
 &  Fake &  Orthogonal & Extra&\multirow{2}{*}{\textbf{$\mathcal{L}_{\text{CE}}$}} & \multirow{2}{*}{\textbf{$\mathcal{L}_{\text{SL}}$}}    && \multirow{2}{*}{\textbf{CIFAR100}} & \multirow{2}{*}{\textbf{SVHN}} & \multirow{2}{*}{\textbf{CIFAR10}} & \multirow{2}{*}{\textbf{SVHN}} & \multirow{2}{*}{\textbf{Texture}}&\multirow{2}{*}{\textbf{iNaturalist}}\\

    & Backbone  &  Sampling & Binary Layer & Data&& & & & & & && \\

    \specialrule{1.5pt}{\aboverulesep}{\belowrulesep} 
  \textbf{A} & \checkmark & \checkmark & \checkmark & - & \checkmark & - & \vline& \gr{81.4/}17.6  & \gr{86.9/}23.5 & \gr{68.4/}12.7 & \gr{79.0/}16.2 & \gr{76.4/}18.8&\gr{82.7/}20.3\\

  \textbf{B} & -&\checkmark & \checkmark &-& - & \checkmark & \vline&   \gr{\underline{90.1}/}56.7& \gr{\underline{93.8}/}51.5 & \gr{75.2/}41.8 & \gr{82.0/}47.5& \gr{\textbf{81.9}/}36.0  &\gr{\underline{84.9}/}48.6\\
  
      \textbf{C} & \checkmark  &\checkmark & - & - & - &\checkmark & \vline &  \gr{85.6/}67.3 & \gr{88.2/}74.6 & \gr{66.9/}57.1 & \gr{78.4/}63.3 & \gr{75.4/}60.7 &\gr{79.8/}70.2\\

  \textbf{D} & \checkmark  &- & \checkmark & -& - &\checkmark & \vline & \gr{85.3/}76.5 & \gr{89.4/}78.1 & \gr{70.5/}61.3 & \gr{74.4/}62.5 & \gr{76.1/}67.4 &\gr{81.3/}72.7\\

  \textbf{E}\textit{\tiny (Ours )}  & \checkmark &\checkmark & \checkmark & - & -& \checkmark & \vline &   \gr{88.2/}\underline{80.1 }&  \gr{93.0/}\underline{86.4}&  \gr{\underline{74.3}/}\underline{67.0 }&   \gr{\underline{81.5}/}\underline{70.6}&   \gr{78.3/}\underline{69.2}&  \gr{84.6/}\underline{75.3}\\

 \textbf{F} \textit{\tiny (Ours+Data)} & \checkmark&\checkmark & \checkmark & \checkmark &-& \checkmark & \vline &  \gr{\textbf{90.4}/}\textbf{81.6} & \gr{\textbf{94.2}/}\textbf{87.9} &  \gr{\textbf{75.7}/}\textbf{68.1} &   \gr{\textbf{82.2}/}\textbf{71.8} &   \gr{\underline{79.2}/}\textbf{70.4}&  \gr{\textbf{85.1}/}\textbf{76.8}\\
  
    \noalign{\vskip 2pt}

    \specialrule{1.5pt}{\aboverulesep}{\belowrulesep}

     \end{tabular}}

\end{table*}

\section{Experiments}

Here we present empirical evidence to validate the effectiveness of our method under various setups, including adversarial attacks, corrupted inputs (non-adversarial perturbations), and clean inputs (non-perturbed scenarios). We note that the backbone architecture for the methods considered is the same as described in Table \ref{Table1:Cifar_OOD}.\\ First, we adversarially train a classifier on ID data and then use it to map the data into a robust embedding space. A Gaussian distribution is fitted around these embeddings, and low-likelihood regions of the distribution are sampled to create fake OOD data as a proxy for OOD test samples. We then demonstrate that time invariance, which establishes that the NODE’s behavior does not explicitly depend on time, leads to more stable behavior of the detector under adversarial attacks (see Section \ref{Time_invariance}). Consequently, we leverage Lyapunov stability regularization under a time-invariant assumption for training. However, a potential challenge arises when ID and OOD equilibrium points are located near each other. As a remedy, we introduce an orthogonal binary layer \cite{li2019orthogonal, xu2022orthogonal, behpour2024gradorth} that enhances the separation between ID and OOD data by increasing the distance between their neighborhoods of Lyapunov-stable equilibrium. This enhances the model’s robustness against shifting adversarial samples from OOD to ID and vice versa. Finally, we use the orthogonal binary layer’s confidence output as the OOD score during inference.

\textbf{Experimental Setup \& Datasets.} We evaluated OOD detection methods under both adversarial and clean scenarios (see Tables \ref{Table1:Cifar_OOD} and \ref{Table2.a:All_OOD_Result}). Each experiment utilized two disjoint datasets: one as the ID dataset and the other as the OOD test set. For Table \ref{Table1:Cifar_OOD}, CIFAR-10 or CIFAR-100 \cite{krizhevsky2009learning} served as the ID datasets. Table \ref{Table2.a:All_OOD_Result} extends the evaluation to ImageNet-1k as the ID dataset, with OOD datasets being comprised of Texture \cite{cimpoi2014describing}, SVHN \cite{netzer2011reading}, iNaturalist \cite{van2018inaturalist}, Places365 \cite{zhou2017places}, LSUN \cite{yu2015lsun}, and iSUN \cite{xu2015isun} -- all sourced from disjoint categories~\cite{hendrycks2019benchmarking}. 

An OSR \cite{vaze2021open} setup was also tested, in which each experiment involved a single dataset that was randomly split into ID (60\%) and OOD (40\%) subclasses, with results averaged over 10 trials. Datasets used for OSR included CIFAR-10, CIFAR-100, ImageNet-1k, MNIST \cite{lecun1998gradient}, FMNIST \cite{xiao2017fashion}, and Imagenette \cite{imagenette} (Table \ref{Table2.b:Open-Set Recognition}). Additionally, models were evaluated on corrupted data using the CIFAR-10-C and CIFAR-100-C benchmarks \cite{hendrycks2019benchmarking} (see Table \ref{Corruption_table}). Specifically, both the ID and OOD data were perturbed with corruptions that did not alter semantics but introduced slight distributional shifts during testing. Further details on the datasets are provided in Appendix~\ref{appendix:dataset_details}.

 \textbf{Evaluation Details.}  For adversarial evaluation, all ID and OOD test data were perturbed by using a fully end-to-end PGD ($l_{\infty}$) attack targeting their OOD scores (as described in Section~\ref{OOD_attack_target}). We used $\epsilon = \frac{8}{255}$ for low-resolution images and $\epsilon = \frac{4}{255}$ for high-resolution images. The PGD attack steps denoted as $M$ were set to 1000, with 10 random initializations sampled from the interval $(-\epsilon, \epsilon)$. The step size for the attack was set to $\alpha = 2.5 \times \frac{\epsilon}{M}$ \cite{madry2018towards}. Additionally, we considered AutoAttack and Adaptive AutoAttack (Table \ref{Table1:Cifar_OOD}). Details on how these attacks are tailored for the detection task can be found in Appendix \ref{appendix:evaluation_and_experimental_setup}. As the primary evaluation metric, we used AUROC, representing the area under the receiver operating characteristic curve. Additionally, we used AUPR and FPR95 as supplementary metrics, with results presented in Table \ref{Table4.b:various_metrics}. AUPR represents the area under the precision-recall curve, while FPR95 measures the false positive rate when the model correctly identifies 95\% of the true positives.
 
 \textbf{Reported and Re-Evaluated Results.} Some methods may show different results here compared to those reported in their original papers \cite{chen2021atom,chen2020robust} due to our use of stronger attacks we incorporated for evaluation, or the more challenging benchmarks used. For example, ALOE \cite{chen2020robust} considered a lower perturbation budget for evaluation (i.e.,$\frac{1}{255}$), and the ATD \cite{azizmalayeri2022your} and RODEO \cite{mirzaeirodeo} benchmarks used CIFAR-10 vs. a union of several datasets, rather than CIFAR-10 vs. CIFAR-100. The union set included datasets such as MNIST, which is significantly different from CIFAR-10, leading to a higher reported robust performance.

\textbf{Results Analysis.} 
Without relying on additional datasets or pretrained models, AROS significantly outperforms existing methods in adversarial settings, achieving up to a 40\% improvement in AUROC and demonstrating competitive results under clean setups (see Table~\ref{Table2.b:Open-Set Recognition}). Specifically, AROS also exhibits greater robustness under various corruptions, further underscoring its effectiveness in OOD detection. We further verify our approach through an extensive ablation study of various components in AROS (see Section~\ref{Ablation_Section}).

We note the superiority of AROS compared to representative methods in terms of robust OOD detection. Notably, AROS, \textit{without} relying on pre-trained models or extra datasets, improves adversarial robust OOD detection performance from 45.9\% to \textbf{74.0\%}. In the OSR setup, the results increased from  52.1\% to \textbf{69.3\%}. Similar gains are observed in robustness against corruptions, as shown in Table~\ref{Corruption_table}. 

For instance, performance improved from  72.5\% to \textbf{81.8\%} on the CIFAR-10-C vs. CIFAR-100-C setup, and from  61.0\% to \textbf{67.9\%} on the CIFAR-100-C vs. CIFAR-10-C benchmark. Meanwhile, AROS achieves competitive results in clean scenarios (82.0\%) compared to state-of-the-art methods like DHM (93.4\%), though it should be noted that DHM performs near zero under adversarial attacks. The trade-off between robustness and clean performance is well-known in the field~\cite{zhang2019theoretically,tsipras2018robustness,madry2017towards}, and AROS offers the best overall balance among existing methods. Furthermore, we demonstrate that by using pre-trained models or auxiliary data, AROS's clean performance can be further improved (see Appendix~\ref{appendix:AROS_extra}). Moreover, we provide additional experiments in Appendix~\ref{appendix:AROS_extra} to support our claims.

\textbf{Classifier Training Strategies for Robust OOD Detection.}\label{Time_invariance}
We assessed the impact of different training strategies on the robust OOD detection performance of various classifiers, including those trained with standard training, adversarial training (AT), and NODE-based methods such as ODENet, LyaDEQ, SODEF, and ASODE. To utilize these classifiers as OOD detectors, various post-hoc score functions were applied, as described in Section \ref{related_work}. The results are presented in Table \ref{Table4.a:lyapunov_CLS}. In brief, adversarially trained classifiers exhibit enhanced robustness compared to standard training but still fall short of optimal performance. Furthermore, the time-invariance assumption in SODEF leads to improved robust performance relative to ODENet, LyaDEQ, and ASODE by effectively constraining the divergence between output states, which motivated us to explore similar frameworks. Notably, AROS demonstrates superior performance compared to all these approaches. 

\textbf{Implementation details.} 
 We use a WideResNet-70-16 model as $f_\theta$ \cite{zagoruyko2016wide} and train it for 200 epochs on classification using ${\text{PGD}}^{10}$. For the integration of $h_{\phi}$, an integration time of $T=5$ is applied. To implement the orthogonal layer $B_{\eta}$, we utilize the \texttt{geotorch.orthogonal} library. Training with the loss $\mathcal{L}_\text{SL}$ is performed over 100 epochs. We used SGD as the optimizer, employing a cosine learning rate decay schedule with an initial learning rate of 0.05 and a batch size of 128. See Appendix ~\ref{appendix:AROS_extra} for more details and additional ablation studies on different components of AROS.

\section{Ablation Study} \label{Ablation_Section}

\textbf{AROS Components.} To verify the effectiveness of AROS, we conducted ablation studies across various datasets. The corresponding results are presented in Table \ref{table5:ablation}. In each experiment, individual components were replaced with alternative ones, while the remaining elements were held constant. In \textit{Config A}, we ignored the designed loss function $\mathcal{L}_{\text{SL}}$ and instead utilized the cross-entropy loss function $\mathcal{L}_{\text{CE}}$ for binary classification. \textit{Config B} represents the scenario in which we train the classifier in the first step without adversarial training on the ID data, instead using standard training. This reduces robustness as $f_{\theta}$ becomes more susceptible to perturbations within ID classes, ultimately making the final detector more vulnerable to attacks. In \textit{Config C}, the orthogonal binary layer was replaced with a regular binary layer. In \textit{Config D}, rather than estimating the ID distribution and sampling OOD data in the embedding space, we substituted this process by creating random Gaussian noise in the embedding space as fake OOD data. This removes the conditioning of the fake OOD distribution on the ID data and, as a result, makes them unrelated. This is in line with previous works that have shown that related and nearby auxiliary OOD samples are more useful \cite{mirzaei2022fake,kong2021opengan}. \textit{Config E} represents our default pipeline. Finally, in \textit{Config F}, we extended AROS by augmenting the fake OOD embedding data with additional OOD images (i.e., Food-101 \cite{bossard2014food}) alongside the proposed fake OOD strategy. Specifically, we transformed these additional OOD images into the embedding space using $f_{\theta}$ and combined them with the crafted fake embeddings, which led to enhanced performance.

\vspace{-3pt}
\section{Conclusions}
In this paper we introduce AROS, a framework for improving OOD detection under adversarial attacks. By leveraging Lyapunov stability theory, AROS drives ID and OOD samples toward stable equilibrium points to mitigate adversarial perturbations. Fake OOD samples are generated in the embedding space, and a tailored loss function is used to enforce stability. Additionally, an orthogonal binary layer is employed to enhance the separation between ID and OOD equilibrium points. Limitations and future directions can be found in the Appendix~\ref{appendix:limitations}.

\bibliography{iclr2025_conference}

\begin{thebibliography}{100}

\bibitem{papernot2016limitations}
Nicolas Papernot, Patrick McDaniel, Somesh Jha, Matt Fredrikson, Z~Berkay Celik, and Ananthram Swami.
\newblock The limitations of deep learning in adversarial settings.
\newblock In {\em 2016 IEEE European symposium on security and privacy (EuroS\&P)}, pages 372--387. IEEE, 2016.

\bibitem{carlini2017towards}
Nicholas Carlini and David Wagner.
\newblock Towards evaluating the robustness of neural networks.
\newblock In {\em 2017 ieee symposium on security and privacy (sp)}, pages 39--57. Ieee, 2017.

\bibitem{szegedy2013intriguing}
Christian Szegedy, Wojciech Zaremba, Ilya Sutskever, Joan Bruna, Dumitru Erhan, Ian Goodfellow, and Rob Fergus.
\newblock Intriguing properties of neural networks.
\newblock {\em arXiv preprint arXiv:1312.6199}, 2013.

\bibitem{madry2018towards}
Aleksander Madry, Aleksandar Makelov, Ludwig Schmidt, Dimitris Tsipras, and Adrian Vladu.
\newblock Towards deep learning models resistant to adversarial attacks.
\newblock In {\em International Conference on Learning Representations}, 2018.

\bibitem{zhang2019theoretically}
Hongyang Zhang, Yaodong Yu, Jiantao Jiao, Eric Xing, Laurent El~Ghaoui, and Michael Jordan.
\newblock Theoretically principled trade-off between robustness and accuracy.
\newblock In {\em International conference on machine learning}, pages 7472--7482. PMLR, 2019.

\bibitem{tramer2018ensemble}
Florian Tramer, Alexey Kurakin, Nicolas Papernot, Ian Goodfellow, Dan Boneh, and Patrick McDaniel.
\newblock Ensemble adversarial training: Attacks and defenses.
\newblock In {\em International Conference on Learning Representations (ICLR)}, 2018.

\bibitem{ledyaev1999lyapunov}
Yuri~S Ledyaev and Eduardo~D Sontag.
\newblock A lyapunov characterization of robust stabilization.
\newblock {\em Nonlinear Analysis-Series A Theory and Methods and Series B Real World Applications}, 37(7):813--840, 1999.

\bibitem{carrara2019robustness}
Fabio Carrara, Roberto Caldelli, Fabrizio Falchi, and Giuseppe Amato.
\newblock On the robustness to adversarial examples of neural ode image classifiers.
\newblock In {\em 2019 IEEE International Workshop on Information Forensics and Security (WIFS)}, pages 1--6. IEEE, 2019.

\bibitem{svoboda2019peernets}
Jan Svoboda, Jonathan Masci, Federico Monti, Michael Bronstein, and Leonidas Guibas.
\newblock Peernets: Exploiting peer wisdom against adversarial attacks.
\newblock In {\em International Conference on Learning Representations}, 2019.

\bibitem{rahnama2020robust}
Arash Rahnama, Andre~T Nguyen, and Edward Raff.
\newblock Robust design of deep neural networks against adversarial attacks based on lyapunov theory.
\newblock In {\em Proceedings of the IEEE/CVF Conference on Computer Vision and Pattern Recognition}, pages 8178--8187, 2020.

\bibitem{li2020implicit}
Mingjie Li, Lingshen He, and Zhouchen Lin.
\newblock Implicit euler skip connections: Enhancing adversarial robustness via numerical stability.
\newblock In {\em International Conference on Machine Learning}, pages 5874--5883. PMLR, 2020.

\bibitem{rodriguez2022lyanet}
Ivan Dario~Jimenez Rodriguez, Aaron Ames, and Yisong Yue.
\newblock Lyanet: A lyapunov framework for training neural odes.
\newblock In {\em International conference on machine learning}, pages 18687--18703. PMLR, 2022.

\bibitem{yang2022closer}
Zonghan Yang, Tianyu Pang, and Yang Liu.
\newblock A closer look at the adversarial robustness of deep equilibrium models.
\newblock {\em Advances in Neural Information Processing Systems}, 35:10448--10461, 2022.

\bibitem{dashkovskiy2023robust}
Sergey Dashkovskiy, Oleksiy Kapustyan, and Vitalii Slynko.
\newblock Robust stability of a nonlinear ode-pde system.
\newblock {\em SIAM Journal on Control and Optimization}, 61(3):1760--1777, 2023.

\bibitem{zeqiri2023efficient}
Mustafa Zeqiri, Mark~Niklas M{\"u}ller, Marc Fischer, and Martin Vechev.
\newblock Efficient certified training and robustness verification of neural odes.
\newblock {\em arXiv preprint arXiv:2303.05246}, 2023.

\bibitem{yang2021generalized}
Jingkang Yang, Kaiyang Zhou, Yixuan Li, and Ziwei Liu.
\newblock Generalized out-of-distribution detection: A survey.
\newblock {\em arXiv preprint arXiv:2110.11334}, 2021.

\bibitem{bendale2015towards}
Abhijit Bendale and Terrance Boult.
\newblock Towards open world recognition.
\newblock In {\em Proceedings of the IEEE Conference on Computer Vision and Pattern Recognition (CVPR)}, pages 1893--1902, 2015.

\bibitem{fort2021exploring}
Stanislav Fort, Jie Ren, and Balaji Lakshminarayanan.
\newblock Exploring the limits of out-of-distribution detection.
\newblock {\em Advances in Neural Information Processing Systems}, 34, 2021.

\bibitem{cao2022deep}
Senqi Cao and Zhongfei Zhang.
\newblock Deep hybrid models for out-of-distribution detection.
\newblock In {\em Proceedings of the IEEE/CVF Conference on Computer Vision and Pattern Recognition}, pages 4733--4743, 2022.

\bibitem{xue2022boosting}
Feng Xue, Zi~He, Chuanlong Xie, Falong Tan, and Zhenguo Li.
\newblock Boosting out-of-distribution detection with multiple pre-trained models.
\newblock {\em arXiv preprint arXiv:2212.12720}, 2022.

\bibitem{du2024does}
Xuefeng Du, Zhen Fang, Ilias Diakonikolas, and Yixuan Li.
\newblock How does unlabeled data provably help out-of-distribution detection?
\newblock {\em arXiv preprint arXiv:2402.03502}, 2024.

\bibitem{petersen2010alzheimer}
Ronald~Carl Petersen, Paul~S Aisen, Laurel~A Beckett, Michael~C Donohue, Anthony~Collins Gamst, Danielle~J Harvey, CR~Jack~Jr, William~J Jagust, Leslie~M Shaw, Arthur~W Toga, et~al.
\newblock Alzheimer's disease neuroimaging initiative (adni) clinical characterization.
\newblock {\em Neurology}, 74(3):201--209, 2010.

\bibitem{goodge2021robustness}
Adam Goodge, Bryan Hooi, See~Kiong Ng, and Wee~Siong Ng.
\newblock Robustness of autoencoders for anomaly detection under adversarial impact.
\newblock In {\em Proceedings of the Twenty-Ninth International Conference on International Joint Conferences on Artificial Intelligence}, pages 1244--1250, 2021.

\bibitem{kong2021opengan}
Shu Kong and Deva Ramanan.
\newblock Opengan: Open-set recognition via open data generation.
\newblock In {\em Proceedings of the IEEE/CVF International Conference on Computer Vision}, pages 813--822, 2021.

\bibitem{roberts2021noveltyIndustry}
Emily Roberts and Rajesh Gupta.
\newblock Applying novelty detection techniques for quality assurance in manufacturing.
\newblock {\em Journal of Industrial and Production Engineering}, 38(4):251--262, 2021.

\bibitem{lo2022adversarially}
Shao-Yuan Lo, Poojan Oza, and Vishal~M Patel.
\newblock Adversarially robust one-class novelty detection.
\newblock {\em IEEE Transactions on Pattern Analysis and Machine Intelligence}, 2022.

\bibitem{hendrycks2018deep}
Dan Hendrycks, Mantas Mazeika, and Thomas Dietterich.
\newblock Deep anomaly detection with outlier exposure.
\newblock {\em arXiv preprint arXiv:1812.04606}, 2018.

\bibitem{chen2020robust}
Jiefeng Chen, Yixuan Li, Xi~Wu, Yingyu Liang, and Somesh Jha.
\newblock Robust out-of-distribution detection for neural networks.
\newblock {\em arXiv preprint arXiv:2003.09711}, 2020.

\bibitem{shao2020open}
Rui Shao, Pramuditha Perera, Pong~C Yuen, and Vishal~M Patel.
\newblock Open-set adversarial defense.
\newblock In {\em Computer Vision--ECCV 2020: 16th European Conference, Glasgow, UK, August 23--28, 2020, Proceedings, Part XVII 16}, pages 682--698. Springer, 2020.

\bibitem{chen2021atom}
Jiefeng Chen, Yixuan Li, Xi~Wu, Yingyu Liang, and Somesh Jha.
\newblock Atom: Robustifying out-of-distribution detection using outlier mining.
\newblock In {\em Machine Learning and Knowledge Discovery in Databases. Research Track: European Conference, ECML PKDD 2021, Bilbao, Spain, September 13--17, 2021, Proceedings, Part III 21}, pages 430--445. Springer, 2021.

\bibitem{meinke2022provably}
Alexander Meinke, Julian Bitterwolf, and Matthias Hein.
\newblock Provably adversarially robust detection of out-of-distribution data (almost) for free.
\newblock {\em Advances in Neural Information Processing Systems}, 35:30167--30180, 2022.

\bibitem{fort2022adversarial}
Stanislav Fort.
\newblock Adversarial vulnerability of powerful near out-of-distribution detection.
\newblock {\em arXiv preprint arXiv:2201.07012}, 2022.

\bibitem{shao2022open}
Rui Shao, Pramuditha Perera, Pong~C Yuen, and Vishal~M Patel.
\newblock Open-set adversarial defense with clean-adversarial mutual learning.
\newblock {\em International Journal of Computer Vision}, 130(4):1070--1087, 2022.

\bibitem{azizmalayeri2022your}
Mohammad Azizmalayeri, Arshia Soltani~Moakhar, Arman Zarei, Reihaneh Zohrabi, Mohammad Manzuri, and Mohammad~Hossein Rohban.
\newblock Your out-of-distribution detection method is not robust!
\newblock {\em Advances in Neural Information Processing Systems}, 35:4887--4901, 2022.

\bibitem{franco2023diffusion}
Nicola Franco, Daniel Korth, Jeanette~Miriam Lorenz, Karsten Roscher, and Stephan Guennemann.
\newblock Diffusion denoised smoothing for certified and adversarial robust out-of-distribution detection.
\newblock {\em arXiv preprint arXiv:2303.14961}, 2023.

\bibitem{bethune2023robust}
Louis B{\'e}thune, Paul Novello, Thibaut Boissin, Guillaume Coiffier, Mathieu Serrurier, Quentin Vincenot, and Andres Troya-Galvis.
\newblock Robust one-class classification with signed distance function using 1-lipschitz neural networks.
\newblock {\em arXiv preprint arXiv:2303.01978}, 2023.

\bibitem{mirzaeirodeo}
Hossein Mirzaei, Mohammad Jafari, Hamid~Reza Dehbashi, Ali Ansari, Sepehr Ghobadi, Masoud Hadi, Arshia~Soltani Moakhar, Mohammad Azizmalayeri, Mahdieh~Soleymani Baghshah, and Mohammad~Hossein Rohban.
\newblock Rodeo: Robust outlier detection via exposing adaptive out-of-distribution samples.
\newblock In {\em Forty-first International Conference on Machine Learning}, 2024.

\bibitem{lorenz2024deciphering}
Peter Lorenz, Mario Fernandez, Jens M{\"u}ller, and Ullrich K{\"o}the.
\newblock Deciphering the definition of adversarial robustness for post-hoc ood detectors.
\newblock {\em arXiv preprint arXiv:2406.15104}, 2024.

\bibitem{du2022vos}
Xuefeng Du, Zhaoning Wang, Mu~Cai, and Yixuan Li.
\newblock Vos: Learning what you don't know by virtual outlier synthesis.
\newblock {\em arXiv preprint arXiv:2202.01197}, 2022.

\bibitem{morteza2022provable}
Peyman Morteza and Yixuan Li.
\newblock Provable guarantees for understanding out-of-distribution detection.
\newblock {\em Proceedings of the AAAI Conference on Artificial Intelligence}, 36(7):7831--7840, 2022.

\bibitem{ming2022poem}
Yifei Ming, Ying Fan, and Yixuan Li.
\newblock Poem: Out-of-distribution detection with posterior sampling.
\newblock In {\em International Conference on Machine Learning}, pages 15650--15665. PMLR, 2022.

\bibitem{zou2024adversarial}
Xin Zou and Weiwei Liu.
\newblock On the adversarial robustness of out-of-distribution generalization models.
\newblock {\em Advances in Neural Information Processing Systems}, 36, 2024.

\bibitem{mirzaei2022fake}
Hossein Mirzaei, Mohammadreza Salehi, Sajjad Shahabi, Efstratios Gavves, Cees~GM Snoek, Mohammad Sabokrou, and Mohammad~Hossein Rohban.
\newblock Fake it till you make it: Near-distribution novelty detection by score-based generative models.
\newblock {\em arXiv preprint arXiv:2205.14297}, 2022.

\bibitem{madry2017towards}
Aleksander Madry, Aleksandar Makelov, Ludwig Schmidt, Dimitris Tsipras, and Adrian Vladu.
\newblock Towards deep learning models resistant to adversarial attacks.
\newblock {\em arXiv preprint arXiv:1706.06083}, 2017.

\bibitem{schmidt2018adversarially}
Ludwig Schmidt, Shibani Santurkar, Dimitris Tsipras, Kunal Talwar, and Aleksander Madry.
\newblock Adversarially robust generalization requires more data.
\newblock {\em Advances in neural information processing systems}, 31, 2018.

\bibitem{nakkiran2019adversarial}
Preetum Nakkiran.
\newblock Adversarial robustness may be at odds with simplicity.
\newblock {\em arXiv preprint arXiv:1901.00532}, 2019.

\bibitem{addepalli2022efficient}
Sravanti Addepalli, Samyak Jain, et~al.
\newblock Efficient and effective augmentation strategy for adversarial training.
\newblock {\em Advances in Neural Information Processing Systems}, 35:1488--1501, 2022.

\bibitem{hendrycks2019benchmarking}
Dan Hendrycks and Thomas Dietterich.
\newblock Benchmarking neural network robustness to common corruptions and perturbations.
\newblock {\em arXiv preprint arXiv:1903.12261}, 2019.

\bibitem{deng2009imagenet}
Jia Deng, Wei Dong, Richard Socher, Li-Jia Li, Kai Li, and Li~Fei-Fei.
\newblock Imagenet: A large-scale hierarchical image database.
\newblock In {\em 2009 IEEE conference on computer vision and pattern recognition}, pages 248--255. Ieee, 2009.

\bibitem{DBLP:journals/corr/abs-2003-01690}
Francesco Croce and Matthias Hein.
\newblock Reliable evaluation of adversarial robustness with an ensemble of diverse parameter-free attacks.
\newblock {\em CoRR}, abs/2003.01690, 2020.

\bibitem{liu2022practical}
Ye~Liu, Yaya Cheng, Lianli Gao, Xianglong Liu, Qilong Zhang, and Jingkuan Song.
\newblock Practical evaluation of adversarial robustness via adaptive auto attack, 2022.

\bibitem{hendrycks2017a}
Dan Hendrycks and Kevin Gimpel.
\newblock A baseline for detecting misclassified and out-of-distribution examples in neural networks.
\newblock In {\em International Conference on Learning Representations}, 2017.

\bibitem{yang2022openood}
Jingkang Yang, Pengyun Wang, Dejian Zou, Zitang Zhou, Kunyuan Ding, Wenxuan Peng, Haoqi Wang, Guangyao Chen, Bo~Li, Yiyou Sun, et~al.
\newblock Openood: Benchmarking generalized out-of-distribution detection.
\newblock {\em Advances in Neural Information Processing Systems}, 35:32598--32611, 2022.

\bibitem{salehi2021unified}
Mohammadreza Salehi, Hossein Mirzaei, Dan Hendrycks, Yixuan Li, Mohammad~Hossein Rohban, and Mohammad Sabokrou.
\newblock A unified survey on anomaly, novelty, open-set, and out-of-distribution detection: Solutions and future challenges.
\newblock {\em arXiv preprint arXiv:2110.14051}, 2021.

\bibitem{geng2020recent}
Chuanxing Geng, Sheng-jun Huang, and Songcan Chen.
\newblock Recent advances in open set recognition: A survey.
\newblock {\em IEEE transactions on pattern analysis and machine intelligence}, 43(10):3614--3631, 2020.

\bibitem{goodfellow2014explaining}
Ian~J Goodfellow, Jonathon Shlens, and Christian Szegedy.
\newblock Explaining and harnessing adversarial examples.
\newblock {\em arXiv preprint arXiv:1412.6572}, 2014.

\bibitem{chen2018neural}
Tian~Qi Chen, Yulia Rubanova, Jesse Bettencourt, and David Duvenaud.
\newblock Neural ordinary differential equations.
\newblock In {\em Advances in Neural Information Processing Systems}, volume~31, pages 6571--6583, 2018.

\bibitem{dupont2019augmented}
Emilien Dupont, Arnaud Doucet, and Yee~Whye Teh.
\newblock Augmented neural odes.
\newblock In {\em Advances in Neural Information Processing Systems (NeurIPS)}, 2019.

\bibitem{grathwohl2018ffjord}
Will Grathwohl, Ricky T.~Q. Chen, Jesse Bettencourt, Ilya Sutskever, and David Duvenaud.
\newblock Ffjord: Free-form continuous dynamics for scalable reversible generative models.
\newblock In {\em International Conference on Learning Representations (ICLR)}, 2019.

\bibitem{lee2018simple}
Kimin Lee, Kibok Lee, Honglak Lee, and Jinwoo Shin.
\newblock A simple unified framework for detecting out-of-distribution samples and adversarial attacks.
\newblock {\em Advances in neural information processing systems}, 31, 2018.

\bibitem{bendale2016towards}
Abhijit Bendale and Terrance~E. Boult.
\newblock Towards open set deep networks.
\newblock In {\em Proceedings of the IEEE Conference on Computer Vision and Pattern Recognition (CVPR)}, pages 1563--1572, 2016.

\bibitem{liu2024category}
Kai Liu, Zhihang Fu, Chao Chen, Sheng Jin, Ze~Chen, Mingyuan Tao, Rongxin Jiang, and Jieping Ye.
\newblock Category-extensible out-of-distribution detection via hierarchical context descriptions.
\newblock {\em Advances in Neural Information Processing Systems}, 36, 2024.

\bibitem{tack2020csi}
Jihoon Tack, Sangwoo Mo, Jongheon Jeong, and Jinwoo Shin.
\newblock Csi: Novelty detection via contrastive learning on distributionally shifted instances.
\newblock {\em Advances in neural information processing systems}, 33:11839--11852, 2020.

\bibitem{yan2019robustness}
Hanshu Yan, Jiawei Du, Vincent~YF Tan, and Jiashi Feng.
\newblock On robustness of neural ordinary differential equations.
\newblock {\em arXiv preprint arXiv:1910.05513}, 2019.

\bibitem{kolter2019learning}
J~Zico Kolter and Gaurav Manek.
\newblock Learning stable deep dynamics models.
\newblock {\em Advances in neural information processing systems}, 32, 2019.

\bibitem{li2022defending}
Xiyuan Li, Zou Xin, and Weiwei Liu.
\newblock Defending against adversarial attacks via neural dynamic system.
\newblock {\em Advances in Neural Information Processing Systems}, 35:6372--6383, 2022.

\bibitem{chu2024lyapunov}
Haoyu Chu, Shikui Wei, Ting Liu, Yao Zhao, and Yuto Miyatake.
\newblock Lyapunov-stable deep equilibrium models.
\newblock {\em Proceedings of the AAAI Conference on Artificial Intelligence}, 38(10):11615--11623, 2024.

\bibitem{amos2017input}
Brandon Amos, Lei Xu, and J~Zico Kolter.
\newblock Input convex neural networks.
\newblock In {\em International conference on machine learning}, pages 146--155. PMLR, 2017.

\bibitem{kang2021stable}
Qiyu Kang, Yang Song, Qinxu Ding, and Wee~Peng Tay.
\newblock Stable neural ode with lyapunov-stable equilibrium points for defending against adversarial attacks.
\newblock {\em Advances in Neural Information Processing Systems}, 34:14925--14937, 2021.

\bibitem{du2023dream}
Xuefeng Du, Yiyou Sun, Xiaojin Zhu, and Yixuan Li.
\newblock Dream the impossible: Outlier imagination with diffusion models.
\newblock {\em arXiv preprint arXiv:2309.13415}, 2023.

\bibitem{dawson2022safe}
Charles Dawson, Zengyi Qin, Sicun Gao, and Chuchu Fan.
\newblock Safe nonlinear control using robust neural lyapunov-barrier functions.
\newblock In {\em Conference on Robot Learning}, pages 1724--1735. PMLR, 2022.

\bibitem{uddin2021altitude}
Nur Uddin, Hendra~G Harno, and Rianto~Adhy Sasongko.
\newblock Altitude control system design of bicopter using lyapunov stability approach.
\newblock In {\em 2021 International Symposium on Electronics and Smart Devices (ISESD)}, pages 1--6. IEEE, 2021.

\bibitem{sharma2020lyapunov}
Aman Sharma and Narendra Kumar.
\newblock Lyapunov stability theory based non linear controller design for a standalone pv system.
\newblock In {\em 2020 IEEE International Conference for Innovation in Technology (INOCON)}, pages 1--7. IEEE, 2020.

\bibitem{chang2019neural}
Ya-Chien Chang, Nima Roohi, and Sicun Gao.
\newblock Neural lyapunov control.
\newblock In {\em Advances in Neural Information Processing Systems}, pages 3240--3249, 2019.

\bibitem{strater2024generalad}
Luc~PJ Str{\"a}ter, Mohammadreza Salehi, Efstratios Gavves, Cees~GM Snoek, and Yuki~M Asano.
\newblock Generalad: Anomaly detection across domains by attending to distorted features.
\newblock {\em arXiv preprint arXiv:2407.12427}, 2024.

\bibitem{cohen2021transformaly}
Matan~Jacob Cohen and Shai Avidan.
\newblock Transformaly--two (feature spaces) are better than one.
\newblock {\em arXiv preprint arXiv:2112.04185}, 2021.

\bibitem{venkataramanan2023gaussian}
Swaminathan Venkataramanan et~al.
\newblock Gaussian latent representations for uncertainty estimation using mahalanobis distance.
\newblock In {\em Proceedings of the IEEE/CVF International Conference on Computer Vision Workshops (ICCVW)}, 2023.

\bibitem{pmlr-v139-jin21a}
Xun Jin et~al.
\newblock Shape your space: A gaussian mixture regularization approach to latent space geometry optimization.
\newblock In {\em Advances in Neural Information Processing Systems}, pages 12688--12701, 2021.

\bibitem{MD}
Kimin Lee, Kibok Lee, Honglak Lee, and Jinwoo Shin.
\newblock A simple unified framework for detecting out-of-distribution samples and adversarial attacks.
\newblock In S.~Bengio, H.~Wallach, H.~Larochelle, K.~Grauman, N.~Cesa-Bianchi, and R.~Garnett, editors, {\em Advances in Neural Information Processing Systems}, volume~31. Curran Associates, Inc., 2018.

\bibitem{ren2021simple}
Jie Ren, Stanislav Fort, Jeremiah Liu, Abhijit~Guha Roy, Shreyas Padhy, and Balaji Lakshminarayanan.
\newblock A simple fix to mahalanobis distance for improving near-ood detection.
\newblock {\em arXiv preprint arXiv:2106.09022}, 2021.

\bibitem{massaroli2020dissecting}
Stefano Massaroli, Atsushi Yamashita, and Hajime Asama.
\newblock Dissecting neural odes.
\newblock In {\em Advances in Neural Information Processing Systems}, 2020.

\bibitem{hornik1991approximation}
Kurt Hornik.
\newblock Approximation capabilities of multilayer feedforward networks.
\newblock {\em Neural networks}, 4(2):251--257, 1991.

\bibitem{chen1984linear}
Chi-Tsong Chen.
\newblock {\em Linear system theory and design}.
\newblock Saunders college publishing, 1984.

\bibitem{arrowsmith1995differential}
DK~Arrowsmith and CM~Place.
\newblock Differential equations, maps and chaotic behavior.
\newblock {\em Champman and Hall, London}, 1995.

\bibitem{horn2012matrix}
Roger~A Horn and Charles~R Johnson.
\newblock {\em Matrix analysis}.
\newblock Cambridge university press, 2012.

\bibitem{li2019orthogonal}
Shuai Li, Kui Jia, Yuxin Wen, Tongliang Liu, and Dacheng Tao.
\newblock Orthogonal deep neural networks.
\newblock {\em IEEE transactions on pattern analysis and machine intelligence}, 43(4):1352--1368, 2019.

\bibitem{li2021future}
Mufan Li, Mihai Nica, and Dan Roy.
\newblock The future is log-gaussian: Resnets and their infinite-depth-and-width limit at initialization.
\newblock {\em Advances in Neural Information Processing Systems}, 34:7852--7864, 2021.

\bibitem{pascanu2012understanding}
R~Pascanu.
\newblock Understanding the exploding gradient problem.
\newblock {\em arXiv preprint arXiv:1211.5063}, 2012.

\bibitem{pauli2021training}
Patricia Pauli, Anne Koch, Julian Berberich, Paul Kohler, and Frank Allg{\"o}wer.
\newblock Training robust neural networks using lipschitz bounds.
\newblock {\em IEEE Control Systems Letters}, 6:121--126, 2021.

\bibitem{huang2021training}
Yujia Huang, Huan Zhang, Yuanyuan Shi, J~Zico Kolter, and Anima Anandkumar.
\newblock Training certifiably robust neural networks with efficient local lipschitz bounds.
\newblock {\em Advances in Neural Information Processing Systems}, 34:22745--22757, 2021.

\bibitem{zuhlke2024adversarial}
Monty-Maximilian Z{\"u}hlke and Daniel Kudenko.
\newblock Adversarial robustness of neural networks from the perspective of lipschitz calculus: A survey.
\newblock {\em ACM Computing Surveys}, 2024.

\bibitem{xu2022orthogonal}
Cong Xu, Xiang Li, and Min Yang.
\newblock An orthogonal classifier for improving the adversarial robustness of neural networks.
\newblock {\em Information Sciences}, 591:251--262, 2022.

\bibitem{behpour2024gradorth}
Sima Behpour, Thang~Long Doan, Xin Li, Wenbin He, Liang Gou, and Liu Ren.
\newblock Gradorth: a simple yet efficient out-of-distribution detection with orthogonal projection of gradients.
\newblock {\em Advances in Neural Information Processing Systems}, 36, 2024.

\bibitem{krizhevsky2009learning}
Alex Krizhevsky and Geoffrey Hinton.
\newblock Learning multiple layers of features from tiny images.
\newblock Technical report, Citeseer, 2009.

\bibitem{cimpoi2014describing}
Mircea Cimpoi, Subhransu Maji, Iasonas Kokkinos, Sammy Mohamed, and Andrea Vedaldi.
\newblock Describing textures in the wild.
\newblock In {\em Proceedings of the IEEE conference on computer vision and pattern recognition}, pages 3606--3613, 2014.

\bibitem{netzer2011reading}
Yuval Netzer, Tao Wang, Adam Coates, Alessandro Bissacco, Bo~Wu, and Andrew~Y Ng.
\newblock Reading digits in natural images with unsupervised feature learning.
\newblock In {\em NIPS Workshop on Deep Learning and Unsupervised Feature Learning}, volume 2011, page~5, 2011.

\bibitem{van2018inaturalist}
Grant Van~Horn, Oisin Mac~Aodha, Yang Song, Yin Cui, Chen Sun, Alex Shepard, Hartwig Adam, Pietro Perona, and Serge Belongie.
\newblock The inaturalist species classification and detection dataset.
\newblock In {\em Proceedings of the IEEE conference on computer vision and pattern recognition}, pages 8769--8778, 2018.

\bibitem{zhou2017places}
Bolei Zhou, Agata Lapedriza, Aditya Khosla, Aude Oliva, and Antonio Torralba.
\newblock Places: A 10 million image database for scene recognition.
\newblock {\em IEEE Transactions on Pattern Analysis and Machine Intelligence}, 40(6):1452--1464, 2017.

\bibitem{yu2015lsun}
Fisher Yu, Yinda Zhang, Shuran Song, Ari Seff, and Jianxiong Xiao.
\newblock Lsun: Construction of a large-scale image dataset using deep learning with humans in the loop.
\newblock In {\em arXiv preprint arXiv:1506.03365}, 2015.

\bibitem{xu2015isun}
Jia-Bin Xu and Jianxiong Xiao.
\newblock Isun: Large-scale scene understanding.
\newblock In {\em arXiv preprint arXiv:1502.03509}, 2015.

\bibitem{vaze2021open}
Sagar Vaze, Kai Han, Andrea Vedaldi, and Andrew Zisserman.
\newblock Open-set recognition: A good closed-set classifier is all you need?
\newblock {\em ICLR}, 2022.

\bibitem{lecun1998gradient}
Yann LeCun, L{\'e}on Bottou, Yoshua Bengio, and Patrick Haffner.
\newblock Gradient-based learning applied to document recognition.
\newblock {\em Proceedings of the IEEE}, 86(11):2278--2324, 1998.

\bibitem{xiao2017fashion}
Han Xiao, Kashif Rasul, and Roland Vollgraf.
\newblock Fashion-mnist: a novel image dataset for benchmarking machine learning algorithms.
\newblock In {\em arXiv preprint arXiv:1708.07747}, 2017.

\bibitem{imagenette}
Jeremy Howard.
\newblock Imagenette.
\newblock \url{https://github.com/fastai/imagenette}.
\newblock Accessed: 2024-09-27.

\bibitem{tsipras2018robustness}
Dimitris Tsipras, Shibani Santurkar, Logan Engstrom, Alexander Turner, and Aleksander Madry.
\newblock Robustness may be at odds with accuracy.
\newblock {\em arXiv preprint arXiv:1805.12152}, 2018.

\bibitem{zagoruyko2016wide}
Sergey Zagoruyko.
\newblock Wide residual networks.
\newblock {\em arXiv preprint arXiv:1605.07146}, 2016.

\bibitem{bossard2014food}
Lukas Bossard, Matthieu Guillaumin, and Luc Van~Gool.
\newblock Food-101--mining discriminative components with random forests.
\newblock In {\em Computer vision--ECCV 2014: 13th European conference, zurich, Switzerland, September 6-12, 2014, proceedings, part VI 13}, pages 446--461. Springer, 2014.

\bibitem{radford2021learning}
Alec Radford, Jong~Wook Kim, Chris Hallacy, Aditya Ramesh, Gabriel Goh, Sandhini Agarwal, Girish Sastry, Amanda Askell, Pamela Mishkin, Jack Clark, et~al.
\newblock Learning transferable visual models from natural language supervision.
\newblock In {\em International conference on machine learning}, pages 8748--8763. PMLR, 2021.

\bibitem{mirzaei2024universal}
Hossein Mirzaei, Mojtaba Nafez, Mohammad Jafari, Mohammad~Bagher Soltani, Mohammad Azizmalayeri, Jafar Habibi, Mohammad Sabokrou, and Mohammad~Hossein Rohban.
\newblock Universal novelty detection through adaptive contrastive learning.
\newblock In {\em Proceedings of the IEEE/CVF Conference on Computer Vision and Pattern Recognition}, pages 22914--22923, 2024.

\bibitem{mirzaeiscanning}
Hossein Mirzaei, Ali Ansari, Bahar~Dibaei Nia, Mojtaba Nafez, Moein Madadi, Sepehr Rezaee, Zeinab~Sadat Taghavi, Arad Maleki, Kian Shamsaie, Mahdi Hajialilue, et~al.
\newblock Scanning trojaned models using out-of-distribution samples.
\newblock In {\em The Thirty-eighth Annual Conference on Neural Information Processing Systems}.

\bibitem{moakhar2023seeking}
Arshia~Soltani Moakhar, Mohammad Azizmalayeri, Hossein Mirzaei, Mohammad~Taghi Manzuri, and Mohammad~Hossein Rohban.
\newblock Seeking next layer neurons' attention for error-backpropagation-like training in a multi-agent network framework.
\newblock {\em arXiv preprint arXiv:2310.09952}, 2023.

\bibitem{mirzaei2024killing}
Hossein Mirzaei, Mohammad Jafari, Hamid~Reza Dehbashi, Zeinab~Sadat Taghavi, Mohammad Sabokrou, and Mohammad~Hossein Rohban.
\newblock Killing it with zero-shot: Adversarially robust novelty detection.
\newblock In {\em ICASSP 2024-2024 IEEE International Conference on Acoustics, Speech and Signal Processing (ICASSP)}, pages 7415--7419. IEEE, 2024.

\bibitem{nie2022diffusion}
Weili Nie, Brandon Guo, Yujia Huang, Chaowei Xiao, Arash Vahdat, and Anima Anandkumar.
\newblock Diffusion models for adversarial purification.
\newblock {\em arXiv preprint arXiv:2205.07460}, 2022.

\bibitem{chen2024diffilter}
Yong Chen, Xuedong Li, Xu~Wang, Peng Hu, and Dezhong Peng.
\newblock Diffilter: Defending against adversarial perturbations with diffusion filter.
\newblock {\em IEEE Transactions on Information Forensics and Security}, 2024.

\bibitem{song2024mimicdiffusion}
Kaiyu Song, Hanjiang Lai, Yan Pan, and Jian Yin.
\newblock Mimicdiffusion: Purifying adversarial perturbation via mimicking clean diffusion model.
\newblock In {\em Proceedings of the IEEE/CVF Conference on Computer Vision and Pattern Recognition}, pages 24665--24674, 2024.

\bibitem{lee2023robust}
Minjong Lee and Dongwoo Kim.
\newblock Robust evaluation of diffusion-based adversarial purification.
\newblock In {\em Proceedings of the IEEE/CVF International Conference on Computer Vision}, pages 134--144, 2023.

\bibitem{dhariwal2021diffusion}
Prafulla Dhariwal and Alexander Nichol.
\newblock Diffusion models beat gans on image synthesis.
\newblock {\em Advances in Neural Information Processing Systems}, 34:8780--8794, 2021.

\bibitem{yu2023distribution}
Runpeng Yu, Songhua Liu, Xingyi Yang, and Xinchao Wang.
\newblock Distribution shift inversion for out-of-distribution prediction.
\newblock In {\em Proceedings of the IEEE/CVF Conference on Computer Vision and Pattern Recognition}, pages 3592--3602, 2023.

\bibitem{cao2024adversarially}
Yuanpu Cao, Lu~Lin, and Jinghui Chen.
\newblock Adversarially robust industrial anomaly detection through diffusion model.
\newblock {\em arXiv preprint arXiv:2408.04839}, 2024.

\bibitem{yang2023improving}
Xiangyuan Yang, Jie Lin, Hanlin Zhang, Xinyu Yang, and Peng Zhao.
\newblock Improving the transferability of adversarial examples via direction tuning.
\newblock {\em arXiv preprint arXiv:2303.15109}, 2023.

\bibitem{lin2024boosting}
Qinliang Lin, Cheng Luo, Zenghao Niu, Xilin He, Weicheng Xie, Yuanbo Hou, Linlin Shen, and Siyang Song.
\newblock Boosting adversarial transferability across model genus by deformation-constrained warping.
\newblock In {\em Proceedings of the AAAI Conference on Artificial Intelligence}, volume~38, pages 3459--3467, 2024.

\bibitem{wu2024improving}
Han Wu, Guanyan Ou, Weibin Wu, and Zibin Zheng.
\newblock Improving transferable targeted adversarial attacks with model self-enhancement.
\newblock In {\em Proceedings of the IEEE/CVF Conference on Computer Vision and Pattern Recognition}, pages 24615--24624, 2024.

\bibitem{gao2019strip}
Yansong Gao, Change Xu, Derui Wang, Shiping Chen, Damith~C Ranasinghe, and Surya Nepal.
\newblock Strip: A defence against trojan attacks on deep neural networks.
\newblock In {\em Proceedings of the 35th annual computer security applications conference}, pages 113--125, 2019.

\bibitem{salman2020adversarially}
Hadi Salman, Andrew Ilyas, Logan Engstrom, Ashish Kapoor, and Aleksander Madry.
\newblock Do adversarially robust imagenet models transfer better?
\newblock {\em Advances in Neural Information Processing Systems}, 33:3533--3545, 2020.

\bibitem{croce2020minimally}
Francesco Croce and Matthias Hein.
\newblock Minimally distorted adversarial examples with a fast adaptive boundary attack.
\newblock In {\em International Conference on Machine Learning}, pages 2196--2205. PMLR, 2020.

\bibitem{croce2020reliable}
Matteo Croce and Matthias Hein.
\newblock Reliable evaluation of adversarial robustness with autoattack.
\newblock In {\em International Conference on Machine Learning}, pages 2701--2711. PMLR, 2020.

\bibitem{andriushchenko2019square}
Maria Andriushchenko, Yang Song, and Zachary~C Lipton.
\newblock Square attack: a query-efficient black-box adversarial attack via random search.
\newblock In {\em International Conference on Learning Representations}, 2020.

\bibitem{Zhou_2023_ICCV}
Mu~Zhou, Lucas Stoffl, Mackenzie~Weygandt Mathis, and Alexander Mathis.
\newblock Rethinking pose estimation in crowds: Overcoming the detection information bottleneck and ambiguity.
\newblock In {\em Proceedings of the IEEE/CVF International Conference on Computer Vision (ICCV)}, pages 14689--14699, October 2023.

\bibitem{Ye2024SuperAnimalPP}
Shaokai Ye, Anastasiia Filippova, Jessy Lauer, Maxime Vidal, Steffen Schneider, Tian Qiu, Alexander Mathis, and Mackenzie~W. Mathis.
\newblock Superanimal pretrained pose estimation models for behavioral analysis.
\newblock {\em Nature Communications}, 15, 2024.

\bibitem{chen2023seeing}
Zijiao Chen, Jiaxin Qing, Tiange Xiang, Wan~Lin Yue, and Juan~Helen Zhou.
\newblock Seeing beyond the brain: Conditional diffusion model with sparse masked modeling for vision decoding.
\newblock In {\em Proceedings of the IEEE/CVF Conference on Computer Vision and Pattern Recognition}, pages 22710--22720, 2023.

\bibitem{example2024}
Scott Zimmerman.
\newblock Guide to the hartman-grobman and poincar´e-bendixon theorems, math181hm.
\newblock \url{https://dl.icdst.org/pdfs/files/56b3b117d3b53b088188facffc85f5e4.pdf}, 2008.
\newblock MATH181HM: Dynamical Systems.

\bibitem{hartman2002ordinary}
Philip Hartman.
\newblock {\em Ordinary differential equations}.
\newblock SIAM, 2002.

\bibitem{strogatz2018nonlinear}
Steven~H Strogatz.
\newblock {\em Nonlinear dynamics and chaos: with applications to physics, biology, chemistry, and engineering}.
\newblock CRC press, 2018.

\bibitem{perko2013differential}
Lawrence Perko.
\newblock {\em Differential equations and dynamical systems}, volume~7.
\newblock Springer Science \& Business Media, 2013.

\bibitem{vidyasagar2002nonlinear}
Mathukumalli Vidyasagar.
\newblock {\em Nonlinear systems analysis}.
\newblock SIAM, 2002.

\bibitem{bhatia2002stability}
Nam~Parshad Bhatia and Giorgio~P Szeg{\"o}.
\newblock {\em Stability theory of dynamical systems}.
\newblock Springer Science \& Business Media, 2002.

\bibitem{mcshane1934extension}
Edward~James McShane.
\newblock Extension of range of functions.
\newblock 1934.

\end{thebibliography}

\bibliographystyle{unsrt}

\newpage
      \begin{center}
\maketitle{\huge \textbf{Appendix} }
    \end{center}

\appendix
\renewcommand{\thesection}{A\arabic{section}}
\setcounter{section}{0}

\section{Review of Related Work}
\label{appendix:model_details}

 In our experimental evaluation we compare our results with works that have proposed a defense mechanism for their OOD detection method, as well as those that have not. We will describe each separately below. Subsequently, we will detail the post-hoc score functions.

\textbf{Adversarially Robust OOD Detection Methods.} Previous adversarially robust methods for OOD detection have built on the outlier exposure \cite{hendrycks2018deep} technique and conducted adversarial training on a union of exposed outliers and ID samples. These methods primarily aim to enhance the outlier exposure technique by improving the diversity of the extra OOD image dataset. To tackle this, ALOE leverages an auxiliary OOD dataset and perturbs the samples to adversarially maximize the KL-divergence between the model’s output and a uniform distribution. It is important to note that the adversarial evaluation in their pipeline was less rigorous than that in our approach, due primarily to the limited budget allocated for adversarial attacks. ATOM, in contrast, selectively samples informative outliers rather than using random outliers. The adversarial evaluation of ATOM is also not entirely standard, as it exclusively targets OOD test samples. However, a more comprehensive approach would involve attacking both ID and OOD test samples to ensure enhanced robustness. Meanwhile, ATD employs a GAN to generate auxiliary OOD images for adversarial training, instead of relying on an external dataset. RODEO aims to demonstrate that enhancing the diversity of auxiliary OOD images, while maintaining their stylistic and semantic alignment with ID samples, will improve robustness. Consequently, they utilize a pretrained CLIP model \cite{radford2021learning} and a diffusion model for OOD image synthesi\cite{mirzaei2022fake,mirzaei2024universal,salehi2021unified,mirzaeiscanning,moakhar2023seeking,mirzaei2024killing}.


\textbf{Standard OOD Detection Methods.}
CSI enhances the outlier detection task by building on standard contrastive learning through the introduction of 'distributionally-shifted augmentations'—transformations that encourage the model to treat augmented versions of a sample as OOD. This approach enables the model to learn representations that more effectively distinguish between ID and OOD samples. The paper also proposes a detection score based on the contrastive features learned through this training scheme, which demonstrates effectiveness across various OOD detection scenarios, including one-class, multi-class, and labeled multi-class settings.  DHM involves modeling the joint density  of data and labels   in a single forward pass. By factorizing this joint density into three sources of uncertainty (aleatoric, distributional, and parametric), DHMs aim to distinguish in-distribution samples from OOD samples. To achieve computational efficiency and scalability, the method employs weight normalization during training and utilizes normalizing flows (NFs) to model the probability distribution of the features. The key idea is to use bi-Lipschitz continuous mappings, enabled by spectral normalization, which allows the use of state-of-the-art deep neural networks for learning expressive and geometry-preserving representations of data. VOS uses virtual outliers to regularize the model's decision boundary and improve its ability to distinguish between ID and OOD data. The framework includes an unknown-aware training objective that uses contrastive learning to shape the uncertainty surface between the known data and the synthesized outliers. This method is effective for both object detection and image classification. CATEX use of two hierarchical contexts—perceptual and spurious—to describe category boundaries more precisely through automatic prompt tuning in vision-language models like CLIP.  The perceptual context distinguishes between different categories (e.g., cats vs. apples), while the spurious context helps identify samples that are similar but not ID (e.g., distinguishing cats from panthers). This hierarchical structure helps create more precise category boundaries.

\textbf{Post-hoc Methods for OOD Detection.} Post-hoc methods for OOD detection are approaches applied after training a classifier on ID samples. These methods utilize information from the classifier to indicate OOD detection. A simple but effective approach is the Maximum Softmax Probability (MSP) method. Applied to a $K$-class classifier $f_c$, MSP returns $\max_{c \in \{1, 2, ..., K\}} f_c(x)$ as the likelihood that the sample $x$ belongs to the ID set. In contrast, OpenMax replaces the softmax layer with a calibrated layer that adjusts the logits by fitting a class-wise probability model, such as the Weibull distribution. Another perspective on OOD detection is to measure the distance of a sample to class-conditional distributions. The Mahalanobis distance (MD) is a prominent method for this. For an ID set with $K$ classes, MD-based approaches fit a class-conditional Gaussian distribution $\mathcal{N} (\mu_k, \Sigma)$ to the pre-logit features $z$. The mean vector and covariance matrix are calculated as follows:

\[
\mu_k = \frac{1}{N} \sum_{i: y_i = k} z_i, \quad \Sigma = \frac{1}{N} \sum_{k=1}^{K} \sum_{i: y_i = k} (z_i - \mu_k)(z_i - \mu_k)^T, \quad k = 1, 2, ..., K. 
\]

The MD for a sample $z'$ relative to class $k$ is defined as: $MD_k(z') = (z' - \mu_k)^T \Sigma^{-1} (z' - \mu_k).$

The final score MD used for OOD detection is given by: $\text{score}_{\text{MD}}(x') = - \min_k \{MD_k(z')\}$

\section{Algorithm}\label{appendix:Psudocode}

Here we provide pseudocode for our proposed AROS framework, designed for adversarially robust OOD detection. We begin by leveraging adversarial training on ID data to obtain robust feature representations, utilizing the well-known practice of adversarial training with 10-step PGD. These features are then used to fit class-conditional multivariate Gaussians, from which we sample low-likelihood regions to generate fake OOD embeddings, effectively creating a proxy for real OOD data in the embedding space. By constructing a balanced training set of ID and fake OOD embeddings, we then employ a stability-based objective using a NODE pipeline, coupled with an orthogonal binary layer. This layer maximizes the separation between the equilibrium points of ID and OOD samples, promoting a robust decision boundary under perturbations. During inference, we compute the OOD score based on the orthogonal binary layer’s output, enabling the model to reliably distinguish ID samples from OOD samples, even in the presence of adversarial attacks.

\begin{algorithm}
\caption{Adversarially Robust OOD Detection through Stability (AROS)}\label{alg:aros}
\begin{algorithmic}[1]
\Require ID training samples $\mathcal{D}^{\text{in}}_{\text{train}}$ consisting of $N$ samples spanning $k$ classes, a $k$-class classifier $f_\theta$, a time-invariant NODE $h_{\phi}$, an orthogonal binary layer $B_{\eta}$, an $\ell_{\infty}$ norm constraint, perturbation budget $\epsilon$ set to $\frac{4}{255}$ for low-resolution and $\frac{8}{255}$ for high-resolution, Lyapunov-based loss $\mathcal{L}_{\text{ST}}$.
\Ensure Adversarially Robust OOD Detector  

\\  

\State \textbf{Step A-1. Adversarial Training of Classifier $f_{\theta}$ on ID:}
\For{each sample $(x, y) \in \mathcal{D}^{\text{in}}_{\text{train}}$}
    \State Generate adversarial example $x^*$ using $\text{PGD}^{10}$ to maximize the cross-entropy loss:
    \State $x^* \gets x + \alpha \cdot \text{sign}(\nabla_x \mathcal{L}_{\text{CE}}(f_\theta(x), y))$, constrained by $\epsilon$
    \State Train the classifier $f_\theta$ on $(x^*, y)$ to improve robustness
\EndFor

\State \textbf{Step A-2. Feature Extraction:}
 \State Map ID samples to the robust embedding space: $\text{ID}_{\text{features}} = f_{\theta}(\mathcal{D}^{\text{in}}_{\text{train}})$ 

\\

\State \textbf{Step B. Fake OOD Embedding Generation:}

\State Fit a class-conditional multivariate Gaussian $\mathcal{N}(\mu_j, \Sigma_j)$ on $\text{ID}_{\text{features}}$ for each class $j$

\For{$0 \leq j < K$}
    
\For{$0 \leq i < \left[\frac{N}{K}\right]$}

  \State Sample synthetic OOD embeddings $r$ from the low-likelihood regions of $\mathcal{N}(\mu_j, \Sigma_j)$
    \State $r_{\text{OOD}} \leftarrow r_{\text{OOD}} \cup \{r\}$
\EndFor
 \EndFor

\State Construct a balanced training set $X_{\text{train}} =\text{ID}_{\text{features}}  \cup r_{\text{OOD}}$

\\

\State \textbf{Step C. Lyapunov Stability Objective Function:}

\State Create the pipeline $B_{\eta}(h_{\phi}(.))$ and train it on $X_{\text{train}}$ using $\mathcal{L}_{\text{ST}}$

\\

\State \textbf{Step D. Inference:}

\For{each test sample $x_{\text{test}}$}
    \State Compute the OOD score based on the probability output of the orthogonal binary layer for each input image $x$: $\text{OOD score}(x) = f_{\theta}(B_{\eta}(h_{\phi}(x)))[1]$
\EndFor

\end{algorithmic}
\end{algorithm}

\newpage

\section{Supplementary Experimental Results and Details for AROS}

\label{appendix:AROS_extra}

\subsection{Additional Experimental Results}

Each experiment discussed in the main text was repeated 10 times, with the reported results representing the mean of these trials. Here, we provide the standard deviation across these runs, as summarized in Table \ref{Table6:std}.

\begin{table*}[!ht]
    \centering
     
    \begin{subtable}{\linewidth}
        \caption{\textbf{Table 6a: Standard deviation of AROS performance under both clean and PGD, across 10 repeated experiments }`\gr{Clean}/$\text{PGD}^{1000}$`.}\label{Table6:std}
        \centering
        \resizebox{\linewidth}{!}{%
         \begin{tabular}{l*{10}{c}}
            \toprule
            AROS&\multicolumn{3}{c}{\textbf{CIFAR-10}} & \multicolumn{3}{c}{\textbf{CIFAR-100}}&\multicolumn{3}{c}{\textbf{ImageNet-1k}}\\

            \cmidrule(lr){2-4} \cmidrule(lr){5-7}\cmidrule(lr){8-10}
      &\textbf{CIFAR-100}&\textbf{SVHN}&\textbf{LSUN}&\textbf{CIFAR-10}&\textbf{SVHN}&\textbf{LSUN}&\textbf{Texture}&\textbf{iNaturalist}&\textbf{LSUN}  \\

            \midrule   

              Mean Performance &\gr{88.2} / 80.1  &\gr{93.0} / 86.4 &\gr{90.6} / 82.4 & \gr{74.3} / 67.0&\gr{81.5} /7 0.6&\gr{74.3} / 68.1&\gr{78.3} / 69.2&\gr{84.6} / 75.3&\gr{79.4} / 69.0 \\
            \midrule   
                 Standard Deviation&$\pm$ 0.9 / $\pm$ 1.6  &$\pm$ 0.7 / $\pm$ 1.3  &$\pm$ 0.6/$\pm$ 0.9  & $\pm$ 1.3 / $\pm$ 1.8 &$\pm$ 1.4 / $\pm$ 2.0 &$\pm$ 1.8 / $\pm$ 2.3&$\pm$ 2.5 / $\pm$ 3.1&$\pm$ 1.9 / $\pm$ 3.0&$\pm$ 2.8 / $\pm$ 3.7 \\
              \midrule   

        \end{tabular}
        }
     \end{subtable}%
\end{table*}

\subsubsection{Comparative Analysis of AROS and Diffusion-Based Purification Methods}

We present additional experiments to further demonstrate the effectiveness of AROS. 
We compare AROS's performance against diffusion-based (purification) methods, with results detailed in Table \ref{Table7:Diffusion_vs_AROS}.

Purification techniques in adversarial training aim to enhance model robustness by `purifying' or `denoising' adversarial examples prior to feeding them into the model. The primary goal is to mitigate adversarial perturbations through preprocessing, often by leveraging neural networks or transformation-based methods to restore perturbed inputs to a state resembling clean data. A common approach involves training a generative diffusion model on the original training samples, which is then utilized as a purification module \cite{nie2022diffusion,chen2024diffilter,song2024mimicdiffusion}. In contrast, certain approaches categorize stable NODE-based methods as non-diffusion-based strategies for improving robustness \cite{lee2023robust}.

However, in the context of OOD detection, purification using diffusion methods \cite{dhariwal2021diffusion} may not be effective. This is primarily because diffusion models trained on ID samples tend to shift the features of unseen OOD data towards ID features during the reverse process. Such a shift can mistakenly transform OOD samples into ID, compromising both OOD detection performance and clean robustness \cite{yu2023distribution}. To emphasize this distinction, we compare our approach against purification using diffusion-based models. Specifically, we adopt the AdvRAD \cite{cao2024adversarially} setup for OOD detection and present a comparison with our method in Table~\ref{Table7:Diffusion_vs_AROS}.

\begin{table*}[!ht]
    \centering
     
    \begin{subtable}{\linewidth}
        \caption{\textbf{Table 7a: Comparison of AROS and AdvRAD under clean and $\text{PGD}^{1000}(l_{\infty})$ evaluation}, measured by AUROC (\%). The table cells denote results in the `\gr{Clean}/$\text{PGD}^{1000}$` format. The perturbation budget $\epsilon$ is set to $\frac{8}{255}$ for low-resolution datasets and $\frac{4}{255}$ for high-resolution datasets.
}\label{Table7:Diffusion_vs_AROS}
        \centering
        \resizebox{\linewidth}{!}{%
         \begin{tabular}{l*{10}{c}}
            \toprule
            Method&\multicolumn{3}{c}{\textbf{CIFAR-10}} & \multicolumn{3}{c}{\textbf{CIFAR-100}}&\multicolumn{3}{c}{\textbf{ImageNet-1k}}\\

            \cmidrule(lr){2-4} \cmidrule(lr){5-7}\cmidrule(lr){8-10}
      &\textbf{CIFAR-100}&\textbf{SVHN}&\textbf{LSUN}&\textbf{CIFAR-10}&\textbf{SVHN}&\textbf{LSUN}&\textbf{Texture}&\textbf{iNaturalist}&\textbf{LSUN}  \\
            \midrule   
AdvRAD&\gr{61.0/}49.0  &\gr{68.5/}52.7 &\gr{66.8/}50.7 &\gr{54.8/}49.4&\gr{60.1/}52.0&\gr{54.8/}50.2&\gr{57.7/}51.0&\gr{62.4/}56.5&\gr{58.5/}50.9   \\

            \midrule   

              AROS&\gr{88.2/}80.1  &\gr{93.0/}86.4 &\gr{90.6/} 82.4 &\gr{74.3/}67.0&\gr{81.5/}70.6&\gr{74.3/}68.1&\gr{78.3/} 69.2&\gr{84.6/}75.3&\gr{79.4/}69.0 \\
            \midrule

        \end{tabular}
        }
     \end{subtable}%
\end{table*}


\subsubsection{AROS Under Additional Adversarial Attacks}

 \begin{figure}[h]
  \begin{center}
    \includegraphics[width=1\linewidth]{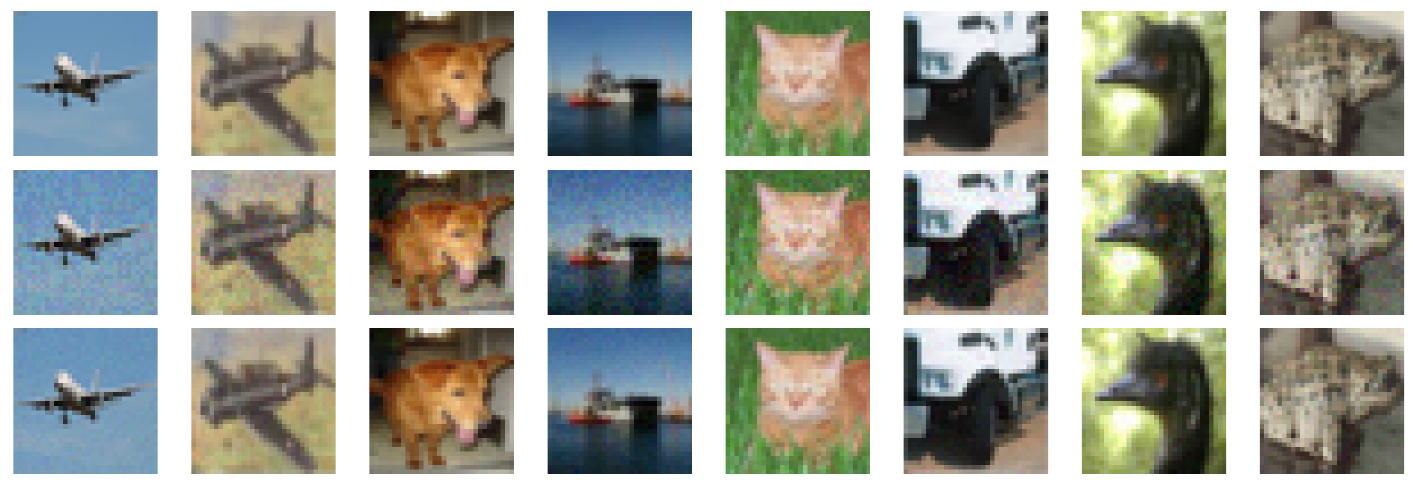}
\caption{Visualization of clean and perturbed images from the CIFAR-10 dataset to illustrate the impact of perturbations on semantic content. The first row depicts clean images, while the second and third rows show images perturbed with $L_\infty$ norm of $\frac{8}{255}$ and $L_2$ norm of $\frac{128}{255}$, respectively. Despite the added perturbations, the semantic content of the images remains unchanged, demonstrating that robustness expectations from models under these perturbations are fair.}

    \label{fig:cifar10_samples_perturbed}
  \end{center}
\end{figure}

In this section, we further emphasize the robustness of AROS by evaluating its performance under different attacks. First, we conducted additional experiments considering different perturbation norms, specifically substituting the $l_\infty$ norm with the $l_2$ norm in $\text{PGD}^{1000}$, setting $\epsilon = \frac{128}{255}$. The outcomes of this evaluation are presented in Table \ref{Table8:L2_norm}. See Figure \ref{fig:cifar10_samples_perturbed}  for an illustration of some clean and perturbed samples.

Next, we evaluated the robustness of AROS under $\text{PGD}^{1000}(l_\infty)$ with a higher perturbation budget (i.e., $\epsilon = \frac{16}{255}$), as shown in Table \ref{Table8:L_inf_norm_high_eps}. Furthermore, the performance of AROS under AutoAttack (AA) and adaptive AA is reported in Table \ref{Table9:Auto_attack_AA}.




Additionally, evaluating a model against a variety of transfer-based adversarial attacks is crucial for understanding its robustness in real-world scenarios, where adversarial examples crafted for one model can successfully deceive others. These attacks simulate diverse and challenging conditions by targeting features such as generalization, input transformations, and model invariance, providing a comprehensive assessment of the model's resilience. Such evaluations reveal potential vulnerabilities, measure performance under adversarial conditions, and offer insights into the model's ability to generalize across different attack strategies.

Motivated by this, we evaluate our method, as well as DHM and RODEO, against several adversarial attacks, including DTA \cite{yang2023improving}, DeCoWA \cite{lin2024boosting}, and SASD-WS \cite{wu2024improving}. We selected DHM and RODEO for comparison due to their strong performance in clean detection and robust detection, respectively. For these experiments, we utilized the implementation provided in the \texttt{https://github.com/Trustworthy-AI-Group/TransferAttack} repository. The results of this evaluation are presented in Table \ref{Table9b:Transfer_attack}.

The consistent and superior performance of AROS against the aforementioned attacks underscores its effectiveness in adversarial scenarios.

\begin{table*}[!ht]
    \centering
 
    \begin{subtable}{\linewidth}
        \caption{\textbf{Table 8a: Evaluation of AROS under $\text{PGD}^{1000}(l_{2})$,} where $\epsilon = \frac{128}{255}$ is used for low-resolution datasets and $\epsilon = \frac{64}{255}$ for high-resolution datasets. The results are measured by AUROC (\%).}\label{Table8:L2_norm}
        \centering
        \resizebox{\linewidth}{!}{%
            \begin{tabular}{l*{10}{c}}
                \toprule
                Attack & \multicolumn{3}{c}{\textbf{CIFAR-10}} & \multicolumn{3}{c}{\textbf{CIFAR-100}} & \multicolumn{3}{c}{\textbf{ImageNet-1k}} \\
                \cmidrule(lr){2-4} \cmidrule(lr){5-7} \cmidrule(lr){8-10}
                & \textbf{CIFAR-100} & \textbf{SVHN} & \textbf{LSUN} & \textbf{CIFAR-10} & \textbf{SVHN} & \textbf{LSUN} & \textbf{Texture} & \textbf{iNaturalist} & \textbf{LSUN} \\
                \midrule
                $\text{PGD}^{1000} (l_2)$ & 81.6 & 87.1 & 83.7 & 67.4 & 71.0 & 69.3 & 70.5 & 75.9 & 69.8 \\
                \midrule
            \end{tabular}
        }
    \end{subtable}
    
    \vspace{0.5cm} 

    \begin{subtable}{\linewidth}
        \caption{\textbf{Table 8b: : Evaluation of AROS under $\text{PGD}^{1000}(l_{\infty})$}, where $\epsilon = \frac{16}{255}$ is used for all datasets, including both high- and low-resolution ones. The results are reported in terms of AUROC (\%).}\label{Table8:L_inf_norm_high_eps}
        \centering
        \resizebox{\linewidth}{!}{%
            \begin{tabular}{l*{10}{c}}
                \toprule
                Attack & \multicolumn{3}{c}{\textbf{CIFAR-10}} & \multicolumn{3}{c}{\textbf{CIFAR-100}} & \multicolumn{3}{c}{\textbf{ImageNet-1k}} \\
                \cmidrule(lr){2-4} \cmidrule(lr){5-7} \cmidrule(lr){8-10}
                & \textbf{CIFAR-100} & \textbf{SVHN} & \textbf{LSUN} & \textbf{CIFAR-10} & \textbf{SVHN} & \textbf{LSUN} & \textbf{Texture} & \textbf{iNaturalist} & \textbf{LSUN} \\
                \midrule
                $\text{PGD}^{1000} (l_\infty)$ & 70.3 & 78.4 & 72.6 & 58.2 & 63.9 & 60.4 & 60.5 & 66.8 & 62.7 \\
                \midrule
            \end{tabular}
        }
    \end{subtable}
\end{table*}

\begin{table*}[!ht]
    \centering

    \begin{subtable}{\linewidth}
        \caption{\textbf{Table 9a: Evaluation of AROS under AutoAttack and Adaptive AA.} The perturbation budget $\epsilon$ is set to $\frac{8}{255}$ for low-resolution datasets and $\frac{4}{255}$ for high-resolution datasets. }\label{Table9:Auto_attack_AA}
        \centering
        \resizebox{\linewidth}{!}{%
            \begin{tabular}{l*{10}{c}}
                \toprule
                Attack & \multicolumn{3}{c}{\textbf{CIFAR-10}} & \multicolumn{3}{c}{\textbf{CIFAR-100}} & \multicolumn{3}{c}{\textbf{ImageNet-1k}} \\
                \cmidrule(lr){2-4} \cmidrule(lr){5-7} \cmidrule(lr){8-10}
                & \textbf{CIFAR-100} & \textbf{SVHN} & \textbf{LSUN} & \textbf{CIFAR-10} & \textbf{SVHN} & \textbf{LSUN} & \textbf{Texture} & \textbf{iNaturalist} & \textbf{LSUN} \\
                \midrule
                AutoAttack & 78.9 & 83.4 & 80.2 & 66.5 & 70.2 & 68.9 & 67.4 & 73.6 & 67.1 \\
                Adaptive AA & 76.4 & 82.9 & 78.6 & 65.2 & 67.4 & 68.3 & 66.1 & 70.5 & 66.9 \\
                \midrule
            \end{tabular}
        }
    \end{subtable}
    
    \vspace{0.5cm} 

    \begin{subtable}{\linewidth}
        \caption{\textbf{Table 9b: Comparison of AROS, DHM, and RODEO under Transfer-Based Attacks.} The perturbation budget $\epsilon$ is set to $\frac{16}{255}$. Results compare AROS, DHM, and RODEO across DTA, DeCoWA, and SASD-WS attacks.}\label{Table9b:Transfer_attack}
        \centering
        \resizebox{\linewidth}{!}{%
            \begin{tabular}{l*{11}{c}}
                \toprule
                Attack & \textbf{Methods} & \multicolumn{3}{c}{\textbf{CIFAR-10}} & \multicolumn{3}{c}{\textbf{CIFAR-100}} & \multicolumn{3}{c}{\textbf{ImageNet-1k}} \\
                \cmidrule(lr){3-5} \cmidrule(lr){6-8} \cmidrule(lr){9-11}
                & & \textbf{CIFAR-100} & \textbf{SVHN} & \textbf{LSUN} & \textbf{CIFAR-10} & \textbf{SVHN} & \textbf{LSUN} & \textbf{Texture} & \textbf{iNaturalist} & \textbf{LSUN} \\
                \midrule
                & AROS & \textbf{84.7} & \textbf{86.4} & 88.2 & \textbf{71.5} & \textbf{76.6} & \textbf{71.7} & \textbf{76.2} & \textbf{83.5} & \textbf{75.9} \\
                DTA & RODEO & 61.6 & 68.5 & \textbf{89.3} & 51.1 & 63.1 & 70.9 & 61.1 & 60.6 & 59.2 \\
                & DHM & 72.9 & 78.3 & 78.9 & 75.4 & 79.1 & 74.1 & 55.9 & 59.2 & 57.6 \\
                \midrule
                & AROS & \textbf{83.8} & \textbf{85.2} & \textbf{86.8} & \textbf{69.6} & \textbf{77.2} & \textbf{69.4} & \textbf{72.6} & \textbf{79.5} & \textbf{74.8} \\
                DeCoWA & RODEO & 57.8 & 64.6 & 80.4 & 43.0 & 57.7 & 67.5 & 54.7 & 54.8 & 52.5 \\
                & DHM & 60.1 & 64.3 & 64.9 & 68.9 & 62.1 & 60.8 & 54.9 & 58.4 & 56.0 \\
                \midrule
                & AROS & \textbf{82.3} & \textbf{84.9} & \textbf{81.7} & \textbf{67.6} & \textbf{75.0} & \textbf{67.7} & \textbf{72.4} & \textbf{77.5} & \textbf{71.2} \\
                SASD-WS & RODEO & 49.5 & 53.4 & 79.4 & 44.7 & 56.5 & 66.1 & 53.2 & 46.9 & 47.0 \\
                & DHM & 64.6 & 69.3 & 61.3 & 56.3 & 67.8 & 59.7 & 53.1 & 51.3 & 49.2 \\
                \midrule
            \end{tabular}
        }
    \end{subtable}
\end{table*}

 \subsection{Detecting Backdoored Samples}

We conducted an experiment to evaluate our method's ability to detect clean samples (without triggers) as ID and poisoned samples as OOD. The results are presented in this section, with STRIP \cite{gao2019strip} included as a baseline method for identifying Trojaned samples.

It is important to highlight a key distinction in the context of OOD detection, which underpins our study. In traditional OOD detection scenarios, ID and OOD datasets typically differ at the semantic level (e.g., CIFAR-10 versus SVHN). However, detecting backdoored samples presents a unique challenge: the presence of triggers modifies clean images at the pixel level rather than the semantic level. Consequently, this task often hinges on texture-level differences rather than semantic distinctions, necessitating models designed specifically to leverage this inductive bias effectively.

Despite these challenges, our results demonstrate that AROS achieves strong performance in detecting poisoned samples. This underscores the versatility and effectiveness of AROS, even when applied to the nuanced problem of backdoor detection. The results are presented in Table \ref{Table10:Backdoor_detect}.

\begin{table*}[!ht]
    \centering
     
    \begin{subtable}{\linewidth}
       \caption{\textbf{Table 10a: Effectiveness of the Proposed Method for Detecting Backdoor Attack Samples} across CIFAR-10, CIFAR-100, and GTSRB datasets. The results are presented for different backdoor attack methods, demonstrating the performance of AROS and STRIP.}
        \label{Table10:Backdoor_detect}
        \centering
        \resizebox{0.6\linewidth}{!}{%
         \begin{tabular}{lcccc}
            \toprule
            \textbf{Method} & \textbf{Backdoor Attack} & \textbf{CIFAR-10} & \textbf{CIFAR-100} & \textbf{GTSRB} \\
            \midrule   
              & Badnets & \textbf{80.3}& 67.5 & 72.8\\
                        AROS & Wanet &   \textbf{62.7} &\textbf{58.9}& \textbf{54.4}\\
                       & SSBA &  \textbf{57.2} & \textbf{72.6} &  \textbf{66.0}\\
            \midrule   

              & Badnets &  79.2& \textbf{86.0 }& \textbf{87.1} \\
                        STRIP & Wanet &39.5 & 48.5 & 35.6 \\
                       & SSBA & 36.4 & 68.5 & 64.1\\
            
            \bottomrule
        \end{tabular}
        }
     \end{subtable}%
\end{table*}

\subsubsection{Architecture comparison \& Enhancing Clean Performance with Transfer Learning}

Additionally, we investigate the influence of replacing our default backbone architecture, WideResNet, with alternative architectures (Table~\ref{Table12:Arcs}). We also explore transfer learning techniques to enhance clean performance by leveraging robust pre-trained classifiers. 

Lastly, while AROS demonstrates a significant improvement of up to 40\% in adversarial robustness, it shows a performance gap of approximately 10\% when compared to state-of-the-art clean detection methods. Although this trade-off between robustness and clean performance is well-documented in the literature \cite{zhang2019theoretically,tsipras2018robustness,madry2017towards}, our aim is to enhance clean performance. A promising approach would involve enhancing clean performance while preserving robustness by leveraging transfer learning through distillation from a large, robust pretrained model, rather than training the classifier from scratch, as is currently done in the pipeline. Specifically, by utilizing adversarially pretrained classifiers on ImageNet \cite{salman2020adversarially}, we aim to improve our clean performance. Results in lower part of Table \ref{Table12:Arcs} indicate that leveraging such pretrained models can improve clean performance by 6\%. Moreover, we also consider using different architectures trained from scratch to further demonstrate the robustness of AROS across various backbones.

\begin{table*}[!ht]
    \centering
     
    \begin{subtable}{\linewidth}
        \caption{\textbf{Table 11a: Ablation study on different backbone architectures \& Transfer learning.} Results are reported under clean and $\text{PGD}^{1000}(l_{\infty})$ evaluations, measured by AUROC (\%). Each table cell presents results in the `\gr{Clean}/$\text{PGD}^{1000}$` format.}\label{Table12:Arcs}
        \footnotesize{  \scriptsize $^\text{\textdagger}$ Denotes that the backbone is adversarially pretrained on ImageNet.}
        \centering
        \resizebox{\linewidth}{!}{%
         \begin{tabular}{l*{10}{c}}
            \toprule
            Backbone ($f_\theta$)&\multicolumn{3}{c}{\textbf{CIFAR-10}} & \multicolumn{3}{c}{\textbf{CIFAR-100}}&\multicolumn{3}{c}{\textbf{ImageNet-1k}}\\

            \cmidrule(lr){2-4} \cmidrule(lr){5-7}\cmidrule(lr){8-10}
      &\textbf{CIFAR-100}&\textbf{SVHN}&\textbf{LSUN}&\textbf{CIFAR-10}&\textbf{SVHN}&\textbf{LSUN}&\textbf{Texture}&\textbf{iNaturalist}&\textbf{LSUN}  \\
            \midrule   
 
              WideResNet (default)&\gr{88.2/}80.1  &\gr{93.0/}86.4 &\gr{90.6/} 82.4 &\gr{74.3/}67.0&\gr{81.5/}70.6&\gr{74.3/}68.1&\gr{78.3/} 69.2&\gr{84.6/}75.3&\gr{79.4/}69.0 \\
            \midrule

  PreActResNet &\gr{83.5/}74.6& \gr{92.1/}82.4& \gr{87.9/}79.6 &\gr{69.6/}62.2& \gr{76.7/}65.2 &\gr{69.9/}62.7 &\gr{73.4/}69.3& \gr{82.2/}73.6& \gr{76.8/}65.6 \\
            
            \midrule   

                 ResNet18&\gr{82.2/}74.4 &\gr{90.5/}80.7 &\gr{84.5/}81.0 &\gr{70.7/}63.7 &\gr{79.0/}66.4 &\gr{74.6/}66.4 &\gr{77.4/}64.4 &\gr{81.8/}72.3 &\gr{78.7/}67.8 \\
            
            \midrule

                ResNet50&\gr{86.5/}79.2 &\gr{92.5/}86.1 &\gr{90.8/}82.7 &\gr{74.5/}66.4 &\gr{81.7/}69.6 &\gr{72.8/}68.3 &\gr{77.9/}67.4 &\gr{82.7/}75.1 &\gr{77.9/}68.5 \\

            \specialrule{1.5pt}{\aboverulesep}{\belowrulesep}

              WideResNet $^\text{\textdagger}$&\gr{91.2/}75.2 &\gr{97.4/}80.9 &\gr{92.1/}77.9 &\gr{80.5/}60.1 &\gr{87.8/}70.0&\gr{76.7/}62.9 &-&- &- \\
            \midrule   
  PreActResNet$^\text{\textdagger}$ &\gr{93.1/}79.5 &\gr{96.3/}84.4& \gr{96.5/}76.6& \gr{75.4/}62.2& \gr{83.1/}66.7 &\gr{79.4/}64.9&-&- &-\\
            
            \midrule   

                 ResNet18$^\text{\textdagger}$& \gr{90.2/}78.0& \gr{97.9/}83.9 &\gr{97.5/}78.6 &\gr{76.2/}63.7 &\gr{83.0/}65.2 &\gr{78.0/}62.2&-&- &-\\
            
            \midrule

                ResNet50$^\text{\textdagger}$&\gr{90.6/}76.8 &\gr{95.5/}82.3 &\gr{94.6/}74.7 &\gr{79.5/}64.9 &\gr{85.7/}69.7 &\gr{78.7/}45.5 &- &- &- \\
            
            \midrule

 ViT-b-16$^\text{\textdagger}$&\gr{94.3/}71.6 &\gr{97.8/}79.7 &\gr{94.6/}76.8 &\gr{85.2/}61.0 &\gr{87.6/}66.9 &\gr{82.7/}59.2 &-&- &-\\
            
            \specialrule{1.5pt}{\aboverulesep}{\belowrulesep}

        \end{tabular}
        }
     \end{subtable}%
\end{table*}

\newpage

\subsection{Hyperparameter Ablation Study} \label{Hyper_ablation_beta}

\subsubsection{Ablation Study on Hyperparameter of Fake Generation  ($\beta$)}

To craft fake samples, we fit a GMM to the embedding space of ID samples. The objective is to sample from the GMM such that the likelihood of the samples is low, ensuring they do not belong to the ID distribution and are located near its boundaries. This approach generates near-OOD samples, which are valuable for understanding the distribution manifold and improving the detector's performance.

In practice, an encoder is used as a feature extractor to obtain embeddings of ID samples. A GMM is then fitted to these embeddings, and the likelihood of each training sample under the GMM is computed. The embeddings are subsequently sorted based on their likelihoods, and the $\beta$-th minimum likelihood of the training samples is used as a threshold. Random samples are drawn from the GMM, and their likelihoods are compared against this threshold. If a sample's likelihood is lower than the threshold, it is retained as fake OOD; otherwise, it is discarded.

For instance, setting $\beta = 0.1$ ensures that the crafted fake OOD samples have a likelihood of belonging to the ID distribution that is lower than 90\% of the ID samples. Conversely, setting $\beta = 0.5$ corresponds to randomly sampling from the ID distribution, rather than targeting low-likelihood regions, which leads to poor performance. An ablation study in our manuscript explores the effects of varying $\beta$ values on model performance.

The sensitivity of AROS to this hyperparameter is analyzed in detail. As shown in Table \ref{Table12:beta_hyper}, AROS demonstrates consistent performance for small values of $\beta$. Our experimental results indicate that selecting $\beta$ values in the range $[0.0, 0.1]$ achieves optimal performance, highlighting the robustness of AROS to changes in $\beta$.

\begin{table*}[!ht]
    \centering
     
    \begin{subtable}{\linewidth}
        \caption{\textbf{Table 12a: Ablation study on the $\beta$ hyperparameter.} Results are reported under clean and $\text{PGD}^{1000}(l_{\infty})$ evaluations, measured by AUROC (\%). Each table cell presents results in the `\gr{Clean}/$\text{PGD}^{1000}$` format.}\label{Table12:beta_hyper}
        \centering
        \resizebox{\linewidth}{!}{%
         \begin{tabular}{l*{10}{c}}
            \toprule
            Hyperparameter&\multicolumn{3}{c}{\textbf{CIFAR-10}} & \multicolumn{3}{c}{\textbf{CIFAR-100}}&\multicolumn{3}{c}{\textbf{ImageNet-1k}}\\

            \cmidrule(lr){2-4} \cmidrule(lr){5-7}\cmidrule(lr){8-10}
      &\textbf{CIFAR-100}&\textbf{SVHN}&\textbf{LSUN}&\textbf{CIFAR-10}&\textbf{SVHN}&\textbf{LSUN}&\textbf{Texture}&\textbf{iNaturalist}&\textbf{LSUN}  \\
            \midrule   
 
              $\beta=0.001 $(default)&\gr{88.2/}80.1  &\gr{93.0/}86.4 &\gr{90.6/} 82.4 &\gr{74.3/}67.0&\gr{81.5/}70.6&\gr{74.3/}68.1&\gr{78.3/} 69.2&\gr{84.6/}75.3&\gr{79.4/}69.0 \\
            \midrule   
  $\beta=0.01$&\gr{87.3/}80.2 &\gr{92.2/}85.3 &\gr{89.4/}80.7 &\gr{74.7/}65.8 &\gr{80.2/}68.9 &\gr{73.1/}66.6 &\gr{78.1/}69.5 &\gr{83.0/}74.2 &\gr{79.0/}68.5 \\
            
            \midrule   

                 $\beta=0.025$&\gr{86.3/}79.6 &\gr{93.8/}85.0 &\gr{89.3/}83.1 &\gr{72.7/}67.5 &\gr{81.7/}70.8 &\gr{74.3/}68.0 &\gr{76.7/}67.7 &\gr{84.7/}75.3 &\gr{79.8/}69.7
 \\
            
            \midrule

                $\beta=0.05$&\gr{86.5/}80.1 &\gr{92.5/}86.1 &\gr{90.8/}82.7 &\gr{74.5/}66.4 &\gr{81.7/}69.6 &\gr{72.8/}68.3 &\gr{77.9/}67.4 &\gr{82.7/}75.1 &\gr{77.9/}68.5 \\
            
            \midrule   

              $\beta=0.075$&\gr{88.4/}78.4 &\gr{92.7/}85.2 &\gr{89.1/}82.3 &\gr{72.4/}65.7 &\gr{81.5/}69.4 &\gr{72.5/}67.9 &\gr{78.6/}69.7 &\gr{84.8/}73.5 &\gr{78.8/}69.3 \\
            
            \midrule  

              $\beta=0.1$&\gr{86.6/}79.5 &\gr{92.0/}85.4 &\gr{90.9/}80.9 &\gr{72.3/}65.8 &\gr{80.3/}69.4 &\gr{73.3/}66.7 &\gr{77.8/}67.5 &\gr{84.1/}74.9 &\gr{78.2/}67.8 \\

            \midrule

 $\beta=0.25$&\gr{79.4/}67.5 &\gr{89.0/}79.5 &\gr{78.7/}69.4 &\gr{67.7/}54.3 &\gr{69.5/}60.3 &\gr{70.1/}62.7 &\gr{62.6/}65.0 &\gr{65.2/}57.9 &\gr{70.9/}62.0\\

      \midrule 

$\beta=0.5$&\gr{65.4/}53.0 &\gr{77.4/}62.3 &\gr{64.7/}54.6 &\gr{57.5/}40.9 &\gr{52.4/}49.2 &\gr{49.1/}41.9 &\gr{55.8/}48.9 &\gr{67.4/}59.7 &\gr{57.9/}53.8\\

            \specialrule{1.5pt}{\aboverulesep}{\belowrulesep}
        \end{tabular}
        }
     \end{subtable}%
\end{table*}

 \begin{figure}[h]
  \begin{center}
    \includegraphics[width=1\linewidth]{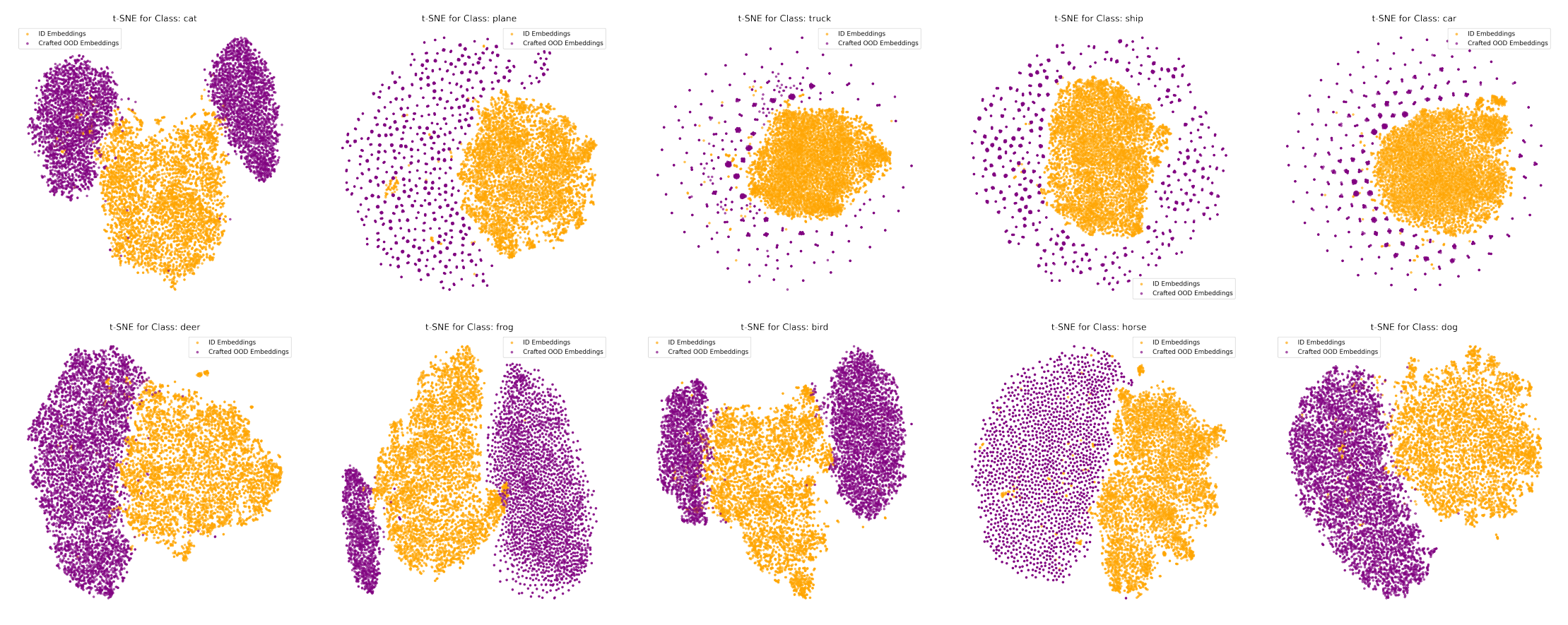}
\caption{t-SNE visualization of CIFAR-10 embeddings and the corresponding crafted OOD embeddings for each class. Orange points represent the ID embeddings for each class, while purple points represent the synthetic OOD embeddings crafted using a GMM. The visualization highlights the separability between ID and OOD embeddings. The crafted embeddings are positioned near the boundaries of the ID concepts, emphasizing they are near OOD samples and they coverage the OOD space. The $\beta$ hyperparameter used in this experiment is set to 0.001.}
    \label{fig:tsne_plot1}
  \end{center}
\end{figure}

 \label{appendix:tsne_plot2}
\begin{figure}[h]
  \begin{center}
    \includegraphics[width=0.6\linewidth]{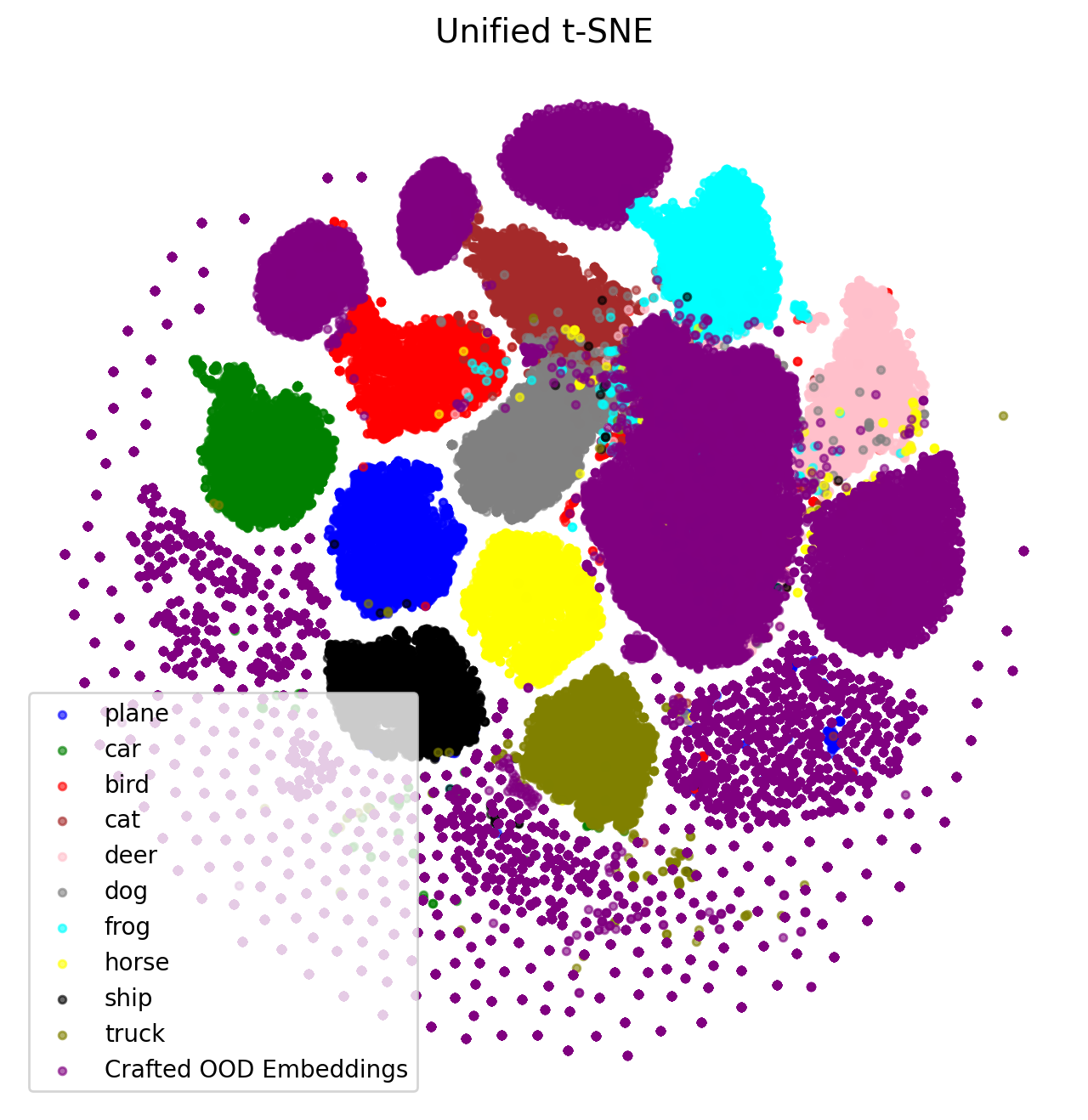}
\caption{Unified t-SNE visualization of embeddings for all CIFAR-10 classes and their corresponding crafted OOD embeddings. Each color represents a specific CIFAR-10 class, while the purple points represent the synthetic OOD embeddings crafted using a GMM. The figure demonstrates the clustering of ID embeddings for each class and the distinct distribution of crafted OOD embeddings. The $\beta$ hyperparameter used in this experiment is set to 0.01. Highlighting both 0.01 and 0.001 leads to crafting effective fake OOD samples, lying out of ID set.}

    \label{fig:tsne_plot2}
  \end{center}
\end{figure}

\subsubsection{Ablation Study on Hyperparameters of the Objective Function ($\gamma$)} \label{gamma_ablation}

In our proposed method, we introduce the empirical loss function $\mathcal{L}_{\text{SL}}$ as follows:
\begin{align}
\mathcal{L}_{\text{SL}} = \min_{\phi,\eta} \frac{1}{\left| X_{\text{train}} \right|} \Bigg( & \ell_{\text{CE}}( B_{\eta}(h_{\phi}(X_{\text{train}})), y ) + \gamma_1 \| h_{\phi}(X_{\text{train}}) \|_2 \notag \\
& + \gamma_2 \exp\left( -\sum_{i=1}^n [\nabla h_{\phi}(X_{\text{train}})]_{ii} \right) \notag \\
& + \gamma_3 \exp\left( \sum_{i=1}^n \left( -\left| [\nabla h_{\phi}(X_{\text{train}})]_{ii} \right| + \sum_{j \neq i} \left| [\nabla h_{\phi}(X_{\text{train}})]_{ij} \right| \right) \right) \Bigg)  
\end{align}

Here, $\gamma_1$ controls the regularization term $\| h_{\phi}(X_{\text{train}}) \|_2$, which encourages the system's state to remain near the equilibrium point. This term helps mitigate the effect of perturbations by ensuring that trajectories stay close to the equilibrium. The hyperparameters $\gamma_2$ and $\gamma_3$ weight the exponential terms designed to enforce Lyapunov stability conditions by ensuring that the Jacobian matrix satisfies strict diagonal dominance, as per Theorem 3 in the paper. Specifically, $\gamma_2$ weights the term $\exp\left( -\sum_{i=1}^n [\nabla h_{\phi}(X_{\text{train}})]_{ii} \right)$, encouraging the diagonal entries of the Jacobian to be negative with large magnitudes. $\gamma_3$ weights the term
\[
\exp\left( \sum_{i=1}^n \left( -\left|[\nabla h_{\phi}(X_{\text{train}})]_{ii}\right| + \sum_{j \neq i} \left|[\nabla h_{\phi}(X_{\text{train}})]_{ij}\right| \right) \right),
\]
promoting strict diagonal dominance by penalizing large off-diagonal entries relative to the diagonal entries.

We set $\gamma_1 = 1$ to balance the influence of the regularization term with the primary cross-entropy loss. This ensures sufficient regularization without overwhelming the classification objective. For $\gamma_2$ and $\gamma_3$, we choose $\gamma_2 = \gamma_3 = 0.05$ Inspired by related works that incorporate stability terms into the training process \cite{carrara2019robustness,svoboda2019peernets,rahnama2020robust,li2020implicit,rodriguez2022lyanet,yang2022closer,dashkovskiy2023robust,zeqiri2023efficient}.

We will now discuss why we select $\gamma_2$ and $\gamma_3$ equally:
\begin{itemize}
    \item \textbf{Balanced Contribution}: Both regularization terms serve complementary purposes in enforcing the stability conditions. Equal weighting ensures that neither term dominates, maintaining a balanced emphasis on both negative diagonal dominance and strict diagonal dominance.
    \item \textbf{Simplified Hyperparameter Tuning}: Setting $\gamma_2$ and $\gamma_3$ equal reduces the hyperparameter search space, simplifying the tuning process without sacrificing performance.
    \item \textbf{Empirical Validation}: Experiments showed that equal values yield robust performance, and deviating from this balance did not provide significant benefits.
\end{itemize}

Regarding the effect of $\gamma_1$:
\begin{itemize}
    \item \textbf{Low Values ($\gamma_1 < 1$)}: Reduces the emphasis on keeping the state near the equilibrium, making the system more susceptible to perturbations.
    \item \textbf{High Values ($\gamma_1 > 1$)}: Overly constrains the state to the equilibrium point, potentially limiting the model's capacity to learn discriminative features.
    \item \textbf{Chosen Value ($\gamma_1 = 1$)}: Offers a good balance, ensuring sufficient regularization without compromising learning.
\end{itemize}

Similarly, the effect of $\gamma_2$ and $\gamma_3$ is as follows:
\begin{itemize}
    \item \textbf{Low Values ($< 0.05$)}: Diminish the impact of the stability constraints, reducing robustness.
    \item \textbf{High Values ($> 0.05$)}: Overemphasize the stability terms, potentially hindering the optimization of the primary classification loss.
    \item \textbf{Equal Values ($\gamma_2 = \gamma_3$)}: Ensure balanced enforcement of both stability conditions, leading to optimal performance.
\end{itemize}

Our stability framework draws inspiration from prior works on deep equilibrium-based models \cite{carrara2019robustness,svoboda2019peernets,rahnama2020robust,li2020implicit,rodriguez2022lyanet,yang2022closer,dashkovskiy2023robust,zeqiri2023efficient}, which proposed similar regularization and hyperparameter tuning techniques. To evaluate the robustness of AROS with respect to these hyperparameters, we conducted extensive ablation studies, holding all components constant while varying the values of $\gamma_1$, $\gamma_2$, and $\gamma_3$. These experiments demonstrate that AROS consistently performs well across a broad range of hyperparameter values, including extreme cases (e.g., $\gamma_1 = 2$). The results of this analysis, presented in Table~\ref{Table10:GAMMA_Hyper}, confirm the robust performance of AROS under varying hyperparameter configurations.

\begin{table*}[!ht]
    \centering
     
    \begin{subtable}{\linewidth}
        \caption{\textbf{Table 13a: An ablation study on the hyperparameters} $\gamma_3$ and $\gamma_2$ and $\gamma_1$. Results are reported under clean and $\text{PGD}^{1000}(l_{\infty})$ evaluations, measured by AUROC (\%). Each table cell presents results in the `\gr{Clean}/$\text{PGD}^{1000}$` format.}\label{Table10:GAMMA_Hyper}
        \centering
        \resizebox{\linewidth}{!}{%
         \begin{tabular}{l*{10}{c}}
            \toprule
            Hyperparameter&\multicolumn{3}{c}{\textbf{CIFAR-10}} & \multicolumn{3}{c}{\textbf{CIFAR-100}}&\multicolumn{3}{c}{\textbf{ImageNet-1k}}\\

            \cmidrule(lr){2-4} \cmidrule(lr){5-7}\cmidrule(lr){8-10}
      &\textbf{CIFAR-100}&\textbf{SVHN}&\textbf{LSUN}&\textbf{CIFAR-10}&\textbf{SVHN}&\textbf{LSUN}&\textbf{Texture}&\textbf{iNaturalist}&\textbf{LSUN}  \\
            \midrule   
 
              $\gamma_1=1$,$\gamma_2=\gamma_3=0.05$ (default)&\gr{88.2/}80.1  &\gr{93.0/}86.4 &\gr{90.6/} 82.4 &\gr{74.3/}67.0&\gr{81.5/}70.6&\gr{74.3/}68.1&\gr{78.3/} 69.2&\gr{84.6/}75.3&\gr{79.4/}69.0 \\
            \midrule   
   $\gamma_1=1$,$\gamma_2=\gamma_3=0.025$&\gr{86.4/}78.2  &\gr{90.4/}86.3 &\gr{88.6/}80.4 &\gr{74.9/}64.1&\gr{78.8/}68.7&\gr{72.8/}68.3&\gr{79.2/}69.7&\gr{83.3/}76.0&\gr{78.8/}67.5 \\
            
             \midrule   
   $\gamma_1=1$,$\gamma_2=\gamma_3=0.075$&\gr{87.5/}80.3  &\gr{92.1/}87.2 &\gr{88.9/}81.6 &\gr{72.7/}67.7&\gr{80.8/}68.7&\gr{73.2/}68.8&\gr{76.4/}69.9&\gr{83.4/}73.4&\gr{80.2/}69.9\\
                \midrule   

   $\gamma_1=1$,$\gamma_2=\gamma_3=0.1$&\gr{86.9/}78.6 &\gr{93.6/}82.1 &\gr{85.8/}78.2 &\gr{70.0/}65.8&\gr{80.6/}66.2&\gr{72.6/}67.3&\gr{79.3/}69.6&\gr{82.8/}74.0&\gr{78.9/}67.2\\

                \midrule   

   $\gamma_1=1$,$\gamma_2=\gamma_3=0.25$&\gr{86.2/}78.7 & \gr{94.3/}82.3 & \gr{87.1/}77.9 & \gr{70.2/}65.2 & \gr{81.3/}65.8 & \gr{73.9/}66.7 & \gr{78.6/}70.1 & \gr{82.8/}74.1 & \gr{78.9/}68.2\\

                \midrule   

   $\gamma_1=1$,$\gamma_2=\gamma_3=0.5$&\gr{85.2/}76.9 & \gr{93.6/}81.5 & \gr{85.0/}76.0 & \gr{69.7/}64.2 & \gr{78.1/}64.4 & \gr{72.6/}65.2 & \gr{77.0/}66.7 & \gr{83.6/}74.4 & \gr{78.2/}67.3 \\

                   \midrule   

   $\gamma_1=\gamma_2=\gamma_3=1$&\gr{84.4/}75.3 & \gr{90.5/}80.5 & \gr{81.5/}73.4 & \gr{69.8/}62.2 & \gr{77.5/}65.3 & \gr{70.1/}64.4 & \gr{73.5/}67.3 & \gr{82.7/}70.9 & \gr{76.7/}65.4\\

                   \midrule   

   $\gamma_1=\gamma_2=\gamma_3=0.25$&\gr{85.4/}77.4 & \gr{94.6/}82.9 & \gr{86.3/}76.9 & \gr{69.9/}64.4 & \gr{79.9/}65.1 & \gr{74.4/}65.3 & \gr{77.2/}68.6 & \gr{85.4/}76.0 & \gr{79.2/}66.8\\

                   \midrule   

   $\gamma_1=0.5$,$\gamma_2=\gamma_3=0.05$&\gr{86.0/}78.6 & \gr{92.7/}82.5 & \gr{86.3/}77.7 & \gr{69.7/}66.1 & \gr{81.2/}65.5 & \gr{70.9/}67.1 & \gr{78.8/}68.6 & \gr{83.0/}73.7 & \gr{78.8/}67.1\\

                   \midrule   

   $\gamma_1=1$,$\gamma_2=0.1,\gamma_3=0.05$&\gr{84.0/}76.3 & \gr{91.5/}81.5 & \gr{81.7/}76.4 & \gr{70.1/}63.4 & \gr{76.2/}63.8 & \gr{69.8/}67.7 & \gr{77.4/}69.5 & \gr{83.2/}71.0 & \gr{78.5/}64.8\\

                   \midrule   

   $\gamma_1=1,\gamma_2=0.05,\gamma_3=0.1$ &\gr{83.5/}77.5 & \gr{90.2/}78.0 & \gr{81.4/}75.1 & \gr{67.2/}63.5 & \gr{76.7/}66.2 & \gr{67.5/}60.4 & \gr{74.3/}70.0 & \gr{83.6/}74.1 & \gr{74.2/}61.7\\

                   \midrule   

$\gamma_1=2$, $\gamma_2=0.5$, $\gamma_3=0.5$ & \gr{80.4/}74.0 & \gr{93.5/}80.7 & \gr{81.8/}70.9 & \gr{65.2/}66.0 & \gr{77.3/}60.0 & \gr{65.0/}60.1 & \gr{78.4/}63.3 & \gr{80.3/}70.6 & \gr{71.6/}59.5\\

            \specialrule{1.5pt}{\aboverulesep}{\belowrulesep}
        \end{tabular}
        }
     \end{subtable}%
\end{table*}

\subsection{Time Complexity}
The computational complexity of our model is reported in Table \ref{Table13:Time_complexity}.

\begin{table*}[!ht]
    \centering
     
    \begin{subtable}{\linewidth}
        \caption{\textbf{Table 14a: Time Complexity of Model Steps} for CIFAR-10, CIFAR-100, and ImageNet-1k, measured on a NVIDIA RTX A5000 GPU (on a workstations running Ubuntu 20.04, Intel Core i9-10900X: 10 cores, 3.70 GHz, 19.25 MB cache; within Docker).}
        \label{Table13:Time_complexity}
        \centering
        \resizebox{0.8\linewidth}{!}{%
         \begin{tabular}{lccc}
            \toprule
            \textbf{Step} & \textbf{CIFAR-10} & \textbf{CIFAR-100} & \textbf{ImageNet-1k} \\
            \midrule   
            Step 1: Adversarial Training of Classifier & 15 hours & 15 hours & 180 hours \\
                        \midrule   
            Step 2: Crafting Fake OOD Data & 7 hours & 7 hours & 150 hours \\
                        \midrule   
            Step 3: Training with Stability Loss ($\mathcal{L}_{\text{SL}}$) & 8 hours & 8 hours & 100 hours \\
            \bottomrule
        \end{tabular}
        }
     \end{subtable}%
\end{table*}

\subsection{Discussion on the Fake Generation Strategy}

\begin{figure}[h]
  \begin{center}
    \includegraphics[width=1\linewidth]{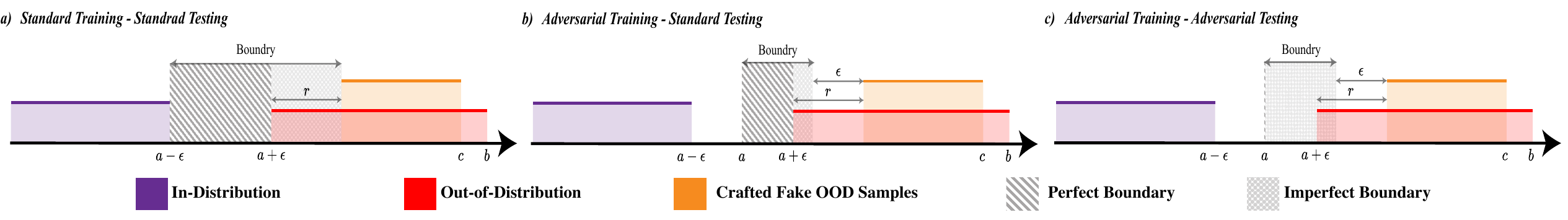}
    \caption{ A depiction in a one-dimensional feature space where the crafted fake OOD samples form a subset of actual OOD data. The gray area represents feasible thresholds separating ID (purple) and fake OOD data (orange). $r$ indicates the shift in exposed fake OOD samples from the real OOD samples. Bold gray lines represent perfect test AUROC thresholds. \textbf{Left}: In standard scenarios, perfect thresholds are abundant, even if the fake OOD samples are distant from the ID data. \textbf{Middle}: With adversarial training, the feasible thresholds decrease due to the maximum margin constraint, impacting the perfect thresholds. Large deviations in the crafted fake OOD data reduce the set of perfect thresholds. \textbf{Right}: For adversarial testing, the overlap between feasible and perfect thresholds narrows to point $a$, emphasizing the importance of near-OOD properties in adversarial contexts.}
    \label{fig:Theory_Plot}
  \end{center}
\end{figure}

It has been shown that utilizing auxiliary realistic OOD samples is generally effective for improving OOD detection performance. However, this strategy comes with several challenges, as discussed before.

First, in certain scenarios, access to an external realistic OOD dataset may not be feasible, and acquiring such data can be challenging. Even when a suitable dataset is available, it must be processed to remove ID concepts to prevent the detector from being misled. This preprocessing step is both time-consuming and computationally expensive. Additionally, studies highlight a potential risk of bias being introduced into the detector when trained on specific auxiliary datasets. Such reliance on a particular realistic dataset may undermine the detector's ability to generalize effectively to diverse OOD samples. These issues become even more pronounced in adversarial training setups, where the complexity of the required data is significantly higher. Motivated by these challenges, this study proposes an alternative strategy that does not depend on the availability of an auxiliary OOD dataset. Notably, our approach is flexible and can incorporate auxiliary OOD datasets as additional information if they are available.
To validate this, we conducted an experiment assuming access to a realistic OOD dataset (i.e., Food-101). In this scenario, we computed embeddings of the real OOD samples and used them alongside crafted fake OOD embeddings during training. The results, presented in Table 4 (Setup A), demonstrate improved performance compared to using fake OOD embeddings alone.
Furthermore, related studies have shown that in adversarial training, using samples near the decision boundary of the distribution improves robustness by encouraging compact representations. This boundary modeling is critical for enhancing the model's robustness, especially against adversarial attacks that exploit vulnerabilities near the decision boundary.
In light of this, our approach shifts focus from generating ``realistic” OOD data to estimating low-likelihood regions of the in-distribution. We generate fake ``near" OOD data that is close to the ID boundary, which is particularly beneficial for adversarial robustness.

For better intuition regarding usefulness of auxiliary near OOD samples here We will provide a simple example that highlights the effectiveness of near-distribution crafted OOD samples in the adversarial setup.  We assume that the feature space is one-dimensional, i.e. $\mathbb{R}$, and the ID class is sampled according to a uniform distribution $U(0, a - \epsilon)$, with $a > 0$, and $\epsilon < a $. We assume that the OOD class is separable with a safety margin of $2\epsilon$ from the ID class to allow a perfectly accurate OOD detector under input perturbations of at most $\epsilon$ at inference. For instance, we let $U(a+\epsilon, b)$ be the feature distribution under the OOD class. The goal is to leverage crafted fake OOD samples to find a robust OOD detector under the $\ell_2$ bounded perturbations of norm $\epsilon$. We assume that the crafted OOD samples data distribution is not perfectly aligned with the anomaly data, e.g. crafted OOD samples comes from $U(a+r, c)$, with $r \geq \epsilon$. It is evident that the optimal robust decision boundary under infinite training samples that separates the ID and crafted OOD samples would be a threshold $k$ satisfying $a \leq k \leq a + r - \epsilon$. The test adversarial error rate to separate ID and OOD classes is $\frac{1}{2}. \mathbb{I}(k \geq a+\epsilon).\left(\frac{k-a-\epsilon}{b-a-\epsilon}  + \frac{\min(k+\epsilon, b)-k}{b - a - \epsilon} \right) + \frac{1}{2}.\mathbb{I}(a < k < a + \epsilon) \frac{\min(k+\epsilon, b) - a - \epsilon}{b - a - \epsilon} $, assuming that the classes are equally probable a prior. It is obvious the adversarial error rate would be zero for $k = a$. But otherwise, if $k \geq a + \epsilon$ the classifier incurs classification error in two cases; in intervals $(a+\epsilon, k)$ (even without any attack), and $(k, \min(k+\epsilon, b))$ in which a perturbation of $-\epsilon$ would cause  classification error. Also if $a < k < a + \epsilon$, classification error only happens at $(a+\epsilon, \min(k+\epsilon, b))$. Now, for the crafted OOD samples to be near-distribution, $r \rightarrow \epsilon$, which forces $k$ to be $a$ in the training, and makes the test adversarial error zero. Otherwise, the adversarial error is proportional to $k$, for $k$ being distant from $b$. Therefore, in the worst case, if $k = a + r - \epsilon$, we get an adversarial error proportional to $r$. As a result, minimizing $r$, which makes the crafted OOD samples near-distribution, would be an effective strategy in making the adversarial error small. Refer to the figure \ref{fig:Theory_Plot} for further intuition and clarity.

To provide a more practical intuition about our crafted fake OOD samples, we present t-SNE visualizations of the embedding space. These visualizations demonstrate that the crafted fake data are positioned near the ID samples and effectively cover their boundary. Please refer to Figures \ref{fig:tsne_plot1} and \ref{fig:tsne_plot2}.

To further demonstrate the superiority of our strategy for crafting fake OOD samples—a simple yet effective technique—we conducted an ablation study by replacing our proposed method with alternative approaches. Please refer to Table \ref{Table14:AROS_fake_better} for the results.

\begin{table*}[!ht]
    \centering
    \begin{subtable}{\linewidth}
        \caption{\textbf{Table 15a: Ablation study on different fake OOD crafting strategies.} Results are measured by AUROC (\%). The perturbation budget $\epsilon$ is set to $\frac{8}{255}$ for low-resolution datasets and $\frac{4}{255}$ for high-resolution datasets. Table cells present results in the `\gr{Clean}/$\text{PGD}^{1000}$' format. We evaluate alternative OOD data synthesis strategies, such as random Gaussian and uniform noise, while keeping other components fixed.}
        \label{Table14:AROS_fake_better}
        \centering
        \resizebox{\linewidth}{!}{%
        \begin{tabular}{l*{9}{c}}
            \toprule
            \textbf{Fake Crafting } & \multicolumn{3}{c}{\textbf{CIFAR-10}} & \multicolumn{3}{c}{\textbf{CIFAR-100}} & \multicolumn{3}{c}{\textbf{ImageNet-1k}} \\
            \cmidrule(lr){2-4} \cmidrule(lr){5-7} \cmidrule(lr){8-10}
             \textbf{Strategy}& \textbf{CIFAR-100} & \textbf{SVHN} & \textbf{LSUN} & \textbf{CIFAR-10} & \textbf{SVHN} & \textbf{LSUN} & \textbf{Texture} & \textbf{iNaturalist} & \textbf{LSUN} \\
            \midrule   
            \textit{AROS} & \gr{88.2/}80.1 & \gr{93.0/}86.4 & \gr{90.6/}82.4 & \gr{74.3/}67.0 & \gr{81.5/}70.6 & \gr{74.3/}68.1 & \gr{78.3/}69.2 & \gr{84.6/}75.3 & \gr{79.4/}69.0 \\
            \midrule   
            Random Gaussian Noise& \gr{85.3/}76.5 & \gr{89.4/}78.1 & \gr{82.5/}77.1&\gr{70.5/}61.3 & \gr{74.4/}62.5& \gr{71.5/}64.6 & \gr{76.1/}67.4 &\gr{81.3/}72.7&\gr{75.8/}67.3 \\
            \midrule   
            Random Uniform Noise & \gr{81.2/}74.3 & \gr{87.1/}79.5 & \gr{84.8/}75.4 & \gr{65.6/}59.6 & \gr{73.8/}63.5 & \gr{65.8/}60.1 & \gr{70.0/}63.8 & \gr{78.8/}67.3 & \gr{71.6/}62.8 \\
            \bottomrule
        \end{tabular}
        }
    \end{subtable}%
\end{table*}

\subsection{orthogonal binary Layer}

\begin{figure}[h]
  \begin{center}
    \includegraphics[width=1\linewidth]{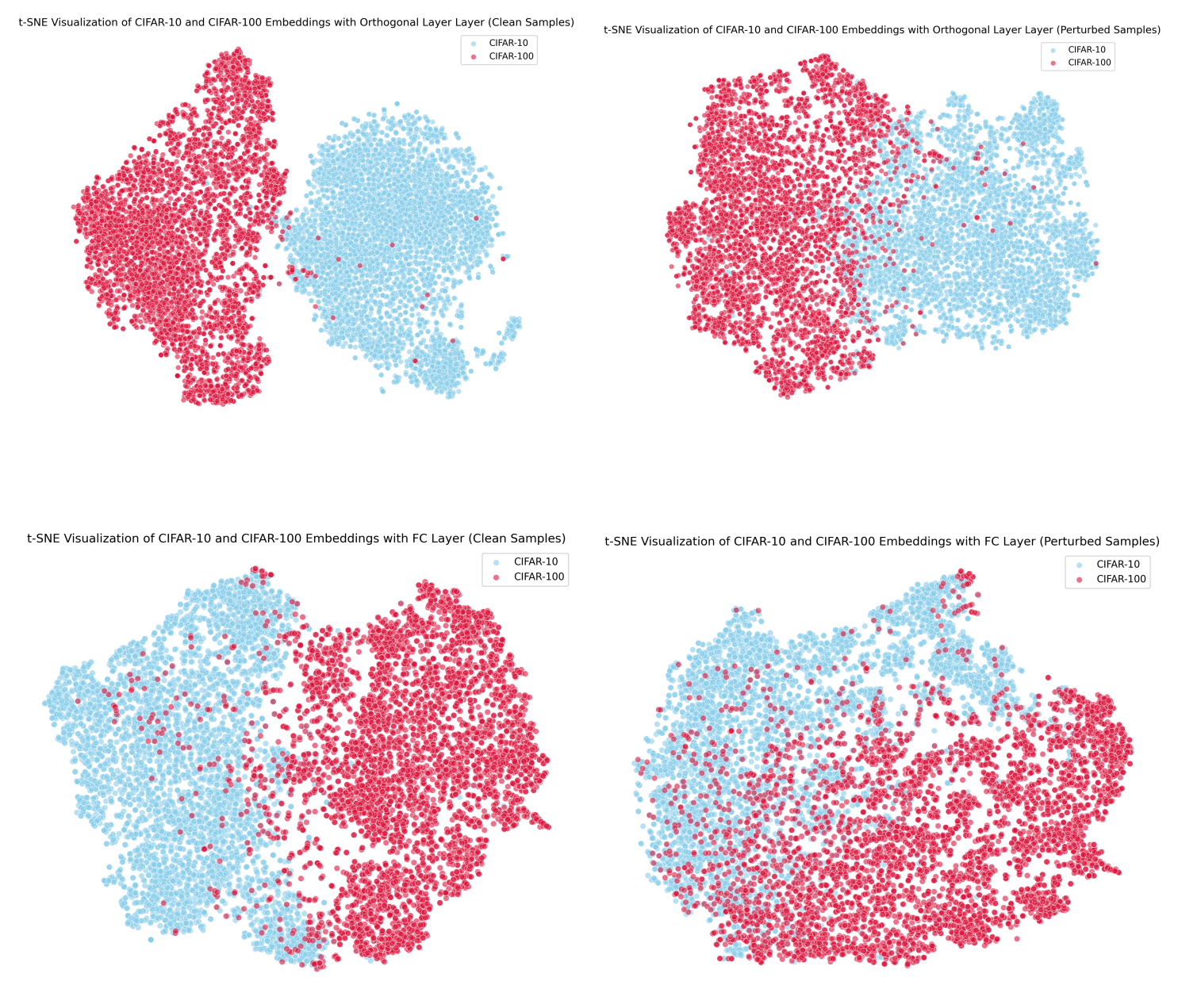}
\caption{t-SNE visualizations of CIFAR-10 (ID) and CIFAR-100 (OOD) embeddings, illustrating the effect of an orthogonal binary layer compared to a regualr fully connected (FC) layer under clean and perturbed sample scenarios. Top row: Embeddings with the orthogonal binary layer for clean (left) and perturbed (right) samples, showing enhanced ID-OOD separation. Bottom row: Embeddings with the FC layer for clean (left) and perturbed (right) samples, demonstrating reduced separation in adversarial settings. These results highlight the orthogonal binary layer's effectiveness in preserving ID-OOD distinction, in both clean and adversarial conditions.}

    \label{fig:Orth_better_tsne}
  \end{center}
\end{figure}

\begin{table*}[!ht]
    \centering
     
    \begin{subtable}{\linewidth}
\caption{\textbf{Table 16a:}Comparison of performance between the orthogonal binary layer (Ortho.) and the regular fully connected (FC) layer across different datasets and perturbation levels ($\epsilon$). The results demonstrate that the orthogonal binary layer consistently outperforms the regular FC layer in both clean ($\epsilon = \frac{0}{255}$) and adversarial scenarios, with varying levels of perturbation ($\epsilon = \frac{2}{255}, \frac{4}{255}, \frac{8}{255}$). }\label{Table16:Ortho_ab}
        \centering
        \resizebox{\linewidth}{!}{%
         \begin{tabular}{l*{11}{c}}
            \toprule
             {\huge $\epsilon$}  & \textbf{Methods} & \multicolumn{3}{c}{\textbf{CIFAR-10}} & \multicolumn{3}{c}{\textbf{CIFAR-100}} & \multicolumn{3}{c}{\textbf{ImageNet-1k}}\\

            \cmidrule(lr){3-5} \cmidrule(lr){6-8} \cmidrule(lr){9-11}
            & & \textbf{CIFAR-100} & \textbf{SVHN} & \textbf{LSUN} & \textbf{CIFAR-10} & \textbf{SVHN} & \textbf{LSUN} & \textbf{Texture} & \textbf{iNaturalist} & \textbf{LSUN}  \\
            \midrule

            \multirow{2}{*}{  $\frac{0}{255}$ (Clean)} & FC & 85.6 & 88.2 & 87.5 & 66.9 & 78.4 & 71.5 & 75.4& 79.8 & 76.1  \\

              & Ortho. &88.2& 93.0& 90.6& 74.3& 81.5 & 74.3& 78.3 & 84.0& 79.4  \\
 
            \midrule  

            \multirow{2}{*}{  $\frac{2}{255}$  } & FC & 82.0 & 84.9 & 84.1 & 63.6 & 75.8 & 68.6 & 73.2 & 76.9 & 72.2   \\

              & Ortho. &85.5 & 90.7 & 88.8 & 72.2 & 79.9 & 72.1 & 75.7 & 82.7 & 76.6 \\
             \midrule

             \multirow{2}{*}{  $\frac{4}{255}$ } & FC & 75.2 & 78.1 & 80.3 & 57.9 & 71.6 & 63.5 & 68.2 & 72.9 & 66.7  \\

              & Ortho.& 83.1& 87.5 & 85.2 & 69.5  & 74.5 &70.4 & 71.6 & 78.3 &68.0
  \\
 
            \midrule  
              \multirow{2}{*}{  $\frac{8}{255}$  } & FC & 67.3 &74.6& 79.5 & 57.1 & 63.3 & 62.4 & 60.7 & 70.2 & 65.1   \\

              & Ortho. & 80.1  & 86.4 &  82.4 & 67.0& 70.6& 68.1&  69.2& 75.3& 69.0 \\
            \midrule

        \end{tabular}
        }
     \end{subtable}%
\end{table*}

To demonstrate the superiority of the orthogonal binary layer used in our pipeline, we conducted an ablation study. In this study, we fixed all other components and replaced the orthogonal binary layer with a regular fully connected (FC) layer, then compared the results across different budget levels. The results of this comparison are presented in Table \ref{Table16:Ortho_ab}. Additionally, we visualized the embedding space corresponding to the regular FC layer and the orthogonal binary layer in Figure \ref{fig:Orth_better_tsne}. These visualizations highlight the enhanced separability provided by the orthogonal binary layer in both clean and adversarial scenarios.

The orthogonal binary layer $B_{\eta}$ is designed to apply a transformation to the NODE output $h_{\phi}(z)$, where the weights $W$ of the layer are constrained to be orthogonal ($W^T W = I$). This constraint encourages maximal separation between the equilibrium points of ID and OOD data by ensuring that the learned representations preserve distinct directions in the feature space. Specifically, the layer operates as:

$$
B_{\eta}(z) = Wz + b, \text{ subject to } W^T W = I,
$$

where $W$ represents the weight matrix, $b$ is the bias term, and the orthogonality constraint is enforced during training using a regularization term.

To clarify its role in $L_{SL}$, we now explicitly annotate $B_\eta$ as the mapping responsible for projecting the NODE output $h_{\phi}(z)$ into a binary classification space (ID vs. OOD). 

The $L_{SL}$ with explicit reference to $W$ and $b$ would be as follows:

$$
L_{SL} = \min_{\phi, w} \frac{1}{\left| X_{\text{train}} \right|} \Bigg(\ell_{\text{CE}}( (W h_{\phi}(X_{\text{train}}) + b), y) + \gamma_1 \| \| h_{\phi}(X_{\text{train}})\|\|_2 
$$  

$$
\quad \quad \quad \quad \quad \quad + \gamma_2 \text{exp} \bigg(-\sum_{i=1}^n [\nabla h_{\phi}(X_{\text{train}})]_{ii} \bigg) 
$$

$$
\quad \quad \quad \quad \quad \quad \quad \quad \quad \quad \quad \quad + \gamma_3 \text{exp} \bigg( \sum_{i=1}^n (-|[\nabla h_{\phi}(X_{\text{train}})]_{ii}| 
$$

$$
\quad \quad \quad \quad \quad \quad \quad \quad \quad \quad \quad \quad + \sum_{j \neq i} |[\nabla h_{\phi}(X_{\text{train}})]_{ij}| ) \bigg) \Bigg)
$$

where the orthogonality constraint $W^T W = I$ is enforced via regularization during optimization.



\section{Details of Evaluation and Experimental Setup}
\label{appendix:evaluation_and_experimental_setup}

OOD detection can be framed as a binary classification task, where training samples are confined to a single set (the ID data), and during testing, input samples from OOD set must be identified. A key challenge is that OOD data is not as clearly defined as ID data; any semantic content absent from the ID distribution is considered OOD. The primary objective in OOD detection is to develop a pipeline capable of assigning meaningful OOD scores to input samples, where a higher score suggests that the model perceives the input as having a greater likelihood of being OOD. An optimal OOD detector would produce score distributions for OOD and ID samples that are fully separated, with no overlap, ensuring clear differentiation between the two.

Formally, the OOD detection decision can be performed as follows:

$$G_\zeta(\mathrm{x})= \begin{cases}\mathrm{ID} & \text { if } S_\mathcal{F}(x) \leq \zeta \\ \mathrm{OOD} & \text { if } S_\mathcal{F}(x)>\zeta\end{cases},$$ 

where $\zeta$ is a threshold parameter.

Thus, the scoring function $S_\mathcal{F}$ is central to the performance of the OOD detector. In the setup of adversarial evaluation, PGD and other attack methods, we adversarially target $S_\mathcal{F}$ and perturb the inputs as described in Section \ref{Preliminaries}. Intuitively, these attacks aim to shift ID samples closer to OOD and vice versa. This approach is fair and consistent with existing OOD detection frameworks, as all such methods define an OOD score function. Furthermore, this adversarial strategy has been previously explored in related work. In the following, we will provide details on AutoAttack and adaptive AutoAttack in our evaluation.

\textbf{AutoAttack.} AutoAttack is an ensemble of six attack methods: APGD with Cross-Entropy loss, APGD with Difference of Logits Ratio (DLR) loss, APGDT, FAB \cite{croce2020minimally}, multi-targeted FAB \cite{croce2020reliable}, and Square Attack \cite{andriushchenko2019square}. However, the DLR loss-based attacks assume that the target model is a classifier trained on more than two classes. In OOD detection, the problem is more akin to binary classification, as we are only distinguishing between OOD and ID classes. Therefore, we excluded these specific DLR-based attacks in our adaptation of AutoAttack and instead used an ensemble of the remaining attacks.

\textbf{Adaptive AA.}  We also employed Adaptive AutoAttack (AA) for evaluation, an attack framework designed to efficiently and reliably approximate the lower bound of a model's robustness. Adaptive AA integrates two main strategies: Adaptive Direction Initialization to generate better starting points for adversarial attacks, and Online Statistics-Based Discarding to prioritize attacking easier-to-perturb images. Unlike standard AutoAttack, Adaptive AA adapts its attack directions based on the model's specific defense properties and dynamically allocates iterations to improve the efficiency of successful attacks. For the perturbation budget, we used a similar budget to that considered for PGD, and for other hyperparameters, we used their default values. For the implementation of adversarial attacks, we utilized the \texttt{Torchattacks} library in PyTorch.

 \subsection{Details of the Reported Results}
 
To report and evaluate previous works, we reproduced their results using the official GitHub repositories. For the clean evaluation setup, if our reproduced results differed from the originally reported values, we reported the higher value. If no results were available, we presented our reproduced results. A similar approach was taken for adversarial evaluation: if their evaluation details and considered datasets matched ours, we used their reported results. However, if the attacks details in their evaluation differed or if they did not consider our datasets or benchmarks, we reported our reproduced results.

 \subsection{Metrics}
 
Our main results are reported using AUROC for comparison across methods, as it is more commonly utilized in the detection literature. However, in Table \ref{Table4.b:various_metrics}, we also compare our method's performance using other metrics, including AUPR and FPR95. Below, we provide an explanation of each of these metrics.

\textbf{AUROC.} The Area Under the Receiver Operating Characteristic Curve (AUROC) is a metric that evaluates classification performance by measuring the trade-off between the True Positive Rate (TPR) and the False Positive Rate (FPR) across various threshold settings. TPR, also known as recall or sensitivity, measures the proportion of actual positives correctly identified, while FPR measures the proportion of actual negatives incorrectly classified as positives. The AUROC, which ranges from 0.5 (random classifier) to 1.0 (perfect classifier), provides an aggregate performance measure. 

\textbf{AUPR.} The Area Under the Precision-Recall Curve (AUPR) is particularly useful for imbalanced datasets where the positive class is rare. The AUPR captures the trade-off between precision (the proportion of predicted positives that are actual positives) and recall. A high AUPR indicates a model with both high precision and recall, providing valuable insights in scenarios where the dataset is skewed. 

\textbf{FPR95.} False Positive Rate at 95\% True Positive Rate (FPR95), which assesses the model's ability to correctly identify ID samples while rejecting OOD samples. Specifically, FPR95 is the false positive rate when the true positive rate is set to 95\%, indicating how often OOD samples are misclassified as ID. Lower values of FPR95 indicate better OOD detection capabilities, as the model more accurately differentiates between ID and OOD samples. Together, these metrics offer a comprehensive understanding of model performance and robustness in classification and OOD detection tasks.

\section{Additional Details about the Datasets}
\label{appendix:dataset_details}
In our experiments, we considered several datasets. Our perturbation budget is determined based on the image size of the ID training set. It is set to  $\frac{4}{255}$ for scenarios where the ID training set consists of high-resolution images (e.g., ImageNet-1K), and  $\frac{8}{255}$ for high-resolution datasets (e.g., CIFAR-10 and CIFAR-100). In all experiments, the ID and OOD datasets contain disjoint semantic classes. If there is an overlap, we exclude those semantics from the OOD test set. In all experiments, the ID and OOD datasets contain disjoint semantic classes. If there is an overlap, we exclude those semantics from the OOD test set. In the following, we provide a brief explanation of the datasets used.

\textbf{CIFAR-10 and CIFAR-100.} are benchmark datasets commonly used for image classification tasks. CIFAR-10 consists of 60,000 color images of size $32\times32$ pixels across 10 classes, with 6,000 images per class. CIFAR-100 is similar but contains 100 classes with 600 images each, providing a more fine-grained classification challenge.

\textbf{ImageNet-1k.} ImageNet-1K contains 1,281,167 training images, 50,000 validation images and 100,000 test images. This dataset was compiled to facilitate research in computer vision by providing a vast range of images to develop and test algorithms, particularly in the areas of object recognition, detection, and classification. 

\textbf{Texture Dataset.} Texture Dataset is designed for studying texture recognition in natural images. It comprises diverse textures captured in the wild, enabling research on classifying and describing textures under varying conditions.

\textbf{Street View House Numbers (SVHN).} SVHN is a real-world image dataset for developing machine learning and object recognition algorithms with minimal data preprocessing. It contains over 600,000 digit images obtained from house numbers in Google Street View images.

\textbf{iNaturalist.} The iNaturalist Species Classification and Detection Dataset aims to address real-world challenges in computer vision by focusing on large-scale, fine-grained classification and detection. This dataset, consisting of 859,000 images from over 5,000 different species of plants and animals, is noted for its high class imbalance and the visual similarity of species within its collection. The images are sourced globally, contributed by a diverse community through the iNaturalist platform, and verified by multiple citizen scientists.

\textbf{Places365.}  Places365 is a scene-centric dataset containing over 10 million images spanning 365 scene categories. It is designed for scene recognition tasks and aids in understanding contextual information in images.

\textbf{Large-scale Scene Understanding (LSUN).}The Large-scale Scene Understanding (LSUN) challenge is designed to set a new standard for large-scale scene classification and comprehension. It features a classification dataset that includes 10 scene categories, such as dining rooms, bedrooms, conference rooms, and outdoor churches, among others. Each category in the training dataset comprises an extensive range of images, from approximately 120,000 to 3,000,000 images. The dataset also provides 300 images per category for validation and 1,000 images per category for testing.

\textbf{iSUN.}The iSUN dataset is a large-scale eye tracking dataset that leverages natural scene images from the SUN database. It was specifically developed to address the limitations of small, in-lab eye tracking datasets by enabling large-scale data collection through a crowdsourced, webcam-based eye tracking system deployed on Amazon Mechanical Turk.

\textbf{MNIST.} The MNIST dataset (Modified National Institute of Standards and Technology dataset) is a large collection of handwritten digits that is widely used for training and testing in the field of machine learning. It contains 70,000 images of handwritten digits from 0 to 9, split into a training set of 60,000 images and a test set of 10,000 images. Each image is a 28x28 pixel grayscale image.

\textbf{Fashion-MNIST (FMNIST).} FMNIST dataset is a collection of article images designed to serve as a more challenging replacement for the traditional MNIST dataset. It consists of 70,000 grayscale images divided into 60,000 training samples and 10,000 test samples, each image having a resolution of 28x28 pixels. The dataset contains 10 different categories of fashion items such as T-shirts/tops, trousers, pullovers, dresses, coats, sandals, shirts, sneakers, bags, and ankle boots. 

\textbf{Imagenette.}  Imagenette is a subset of ImageNet consisting of ten easily classified classes. It was released to encourage research on smaller datasets that require less computational resources, facilitating experimentation and algorithm development.

\textbf{CIFAR-10-C and CIFAR-100-C}. CIFAR-10-C and CIFAR-100-C are corrupted versions of CIFAR-10 and CIFAR-100, respectively. They include various common corruptions and perturbations such as noise, blur, and weather effects, used to evaluate the robustness of image classification models against real-world imperfections.

\textbf{ADNI Neuroimaging Dataset.} The Alzheimer’s Disease Neuroimaging Initiative (ADNI) dataset is a large-scale collection of neuroimaging data aimed at tracking the progression of Alzheimer's disease. In our study, we categorize the dataset into six classes based on cognitive status and disease progression. The classes CN (Cognitively Normal) and SMC (Subjective Memory Concerns) are designated as ID, representing individuals without significant neurodegenerative conditions or with only minor memory concerns. The other four classes—AD (Alzheimer's Disease), MCI (Mild Cognitive Impairment), EMCI (Early Mild Cognitive Impairment), and LMCI (Late Mild Cognitive Impairment)—are treated as OOD, encompassing various stages of cognitive decline related to Alzheimer's. The dataset contains 6,000 records for each class, providing a comprehensive representation of the spectrum of cognitive health and Alzheimer's disease. A summary of each class is as follows:
\begin{itemize}
    \item \textbf{AD}: Patients diagnosed with Alzheimer's disease, showing significant cognitive decline.
    \item \textbf{CN}: Individuals with normal cognitive functioning, serving as a control group.
    \item \textbf{MCI}: A stage of cognitive decline that lies between normal aging and more advanced impairment, often preceding Alzheimer's disease.
    \item \textbf{EMCI}: Patients with early signs of mild cognitive impairment, indicating the initial onset of cognitive issues.
    \item \textbf{LMCI}: Patients with symptoms indicative of more advanced mild cognitive impairment, closer to Alzheimer's in severity.
    \item \textbf{SMC}: Individuals reporting memory concerns but performing normally on cognitive assessments.
\end{itemize}
We used the middle slice of MRI scans from ADNI phases 1, 2, and 3. We should note that the ADNI dataset was the only one we split into ID and OOD using the strategy mentioned earlier. For the other datasets in our OSR experiments, we randomly divided each dataset into ID and OOD at a ratio of 0.6 and 0.4, respectively.

\subsection{Details on Dataset Corruptions}

 \textbf{Corruptions}: The first type of corruption is Gaussian noise, which commonly appears in low-light conditions. Shot noise, also known as Poisson noise, results from the discrete nature of light and contributes to electronic noise. Impulse noise, similar to salt-and-pepper noise but in color, often arises due to bit errors. Defocus blur happens when an image is out of focus. Frosted glass blur is caused by the appearance of "frosted glass" on windows or panels. Motion blur occurs due to rapid camera movement, while zoom blur happens when the camera moves swiftly towards an object. Snow is an obstructive form of precipitation, and frost occurs when lenses or windows accumulate ice crystals. Fog obscures objects and is often simulated using the diamond-square algorithm. Brightness is affected by the intensity of daylight, and contrast levels change based on lighting conditions and the object's color. Elastic transformations distort small image regions by stretching or contracting them. Pixelation results from upscaling low-resolution images. Lastly, JPEG compression, a lossy format, introduces noticeable compression artifacts.

\section{Limitations and Future Research} \label{appendix:limitations}

We considered classification image-based benchmarks, that while difficult, do not cover the full breadth of real-world attacks in situations such as more complex open-set images or video, i.e., time-series data. Future work should test our proposed method in video streaming data (time-series), as inherently we leverage stability properties in dynamical systems that could be very attractive for such settings. Namely, because video frames are often highly correlated, one can even leverage the prior stability points in time in order to make better future predictions. This could also lower the compute time.

Moreover, moving to tasks beyond classification could be very attractive. Concretely, one example is pose estimation. This is a keypoint detection task that can often be cast as a panoptic segmentation task, where each keypoint needs to be identified in the image and grouped appropriately~\cite{Zhou_2023_ICCV,Ye2024SuperAnimalPP}. By using AROS, one could find OOD poses could not only improve data quality, but alert users to wrongly annotated data.  Another example is in brain decoding where video frame or scene classification is critical, and diffusion models are becoming an attractive way to leverage generative models~\cite{chen2023seeing}. Given AROS ability in OOD detection, this could be smartly used to correct wrong predictions. 
In summary, adapting AROS to these other data domains could further extend its applicability.

\section{Theoretical insight and Background}

\begin{figure}[h]
  \begin{center}
    \includegraphics[width=\linewidth]{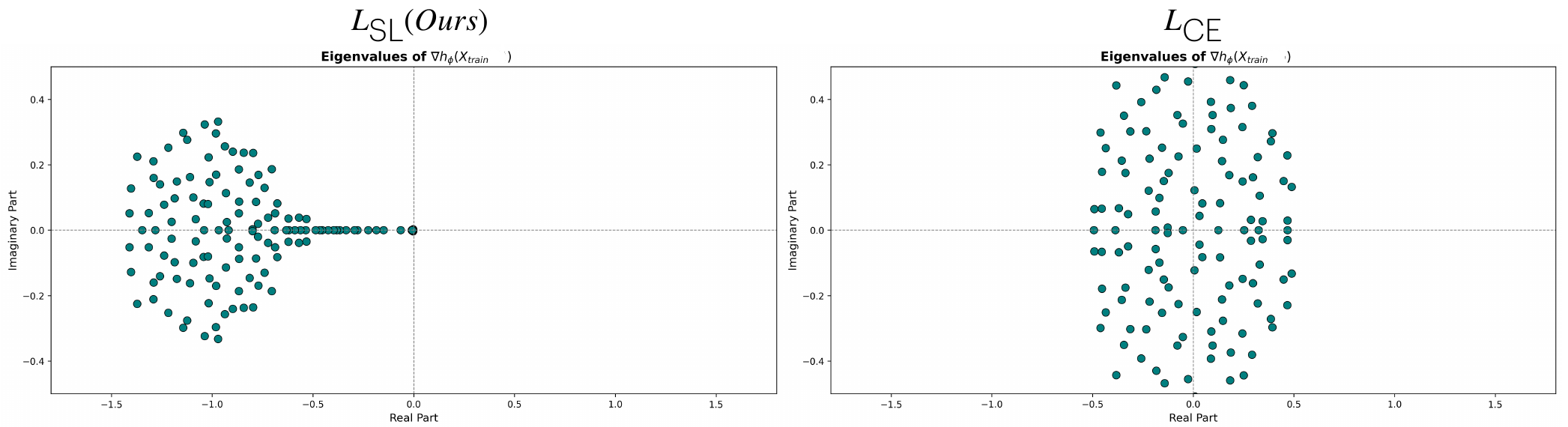}
\caption{Eigenvalue visualization of the Jacobian matrix $\nabla h_\theta(z(0))$ for a NODE trained on CIFAR-10 using using the loss functions $\mathcal{L}_{\text{CE}}$ and $\mathcal{L}_{\text{SL}}$. The results demonstrate that $\mathcal{L}_{\text{SL}}$ encourages eigenvalues with negative real parts, indicating stability.}

    \label{fig:Jacob_Eigen}
  \end{center}
\end{figure}

\begin{figure}[h]
  \begin{center}
    \includegraphics[width=\linewidth]{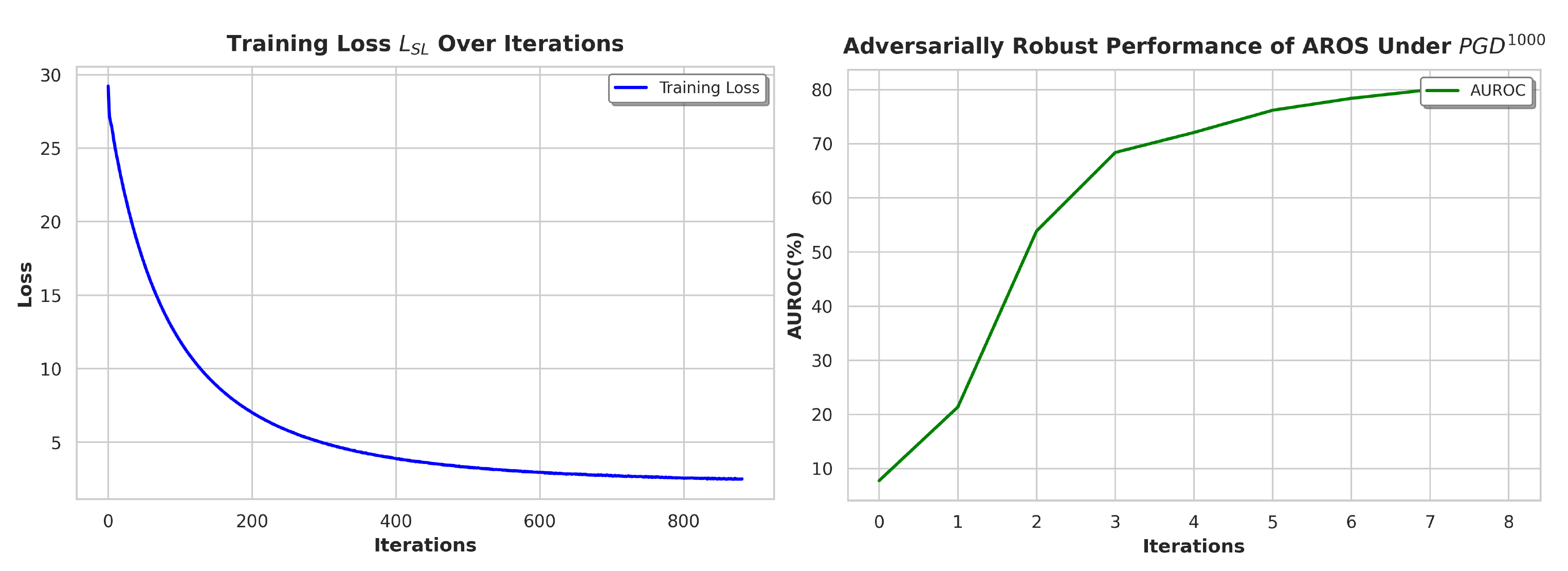}
\caption{  Training loss ($\mathcal{L}_{\text{SL}}$) and adversarial robustness performance of AROS on the CIFAR-10 vs. CIFAR-100 benchmark (CIFAR-10 served as the ID dataset). The left plot shows the convergence of the stability-based loss $\mathcal{L}_{\text{SL}}$ over iterations, demonstrating effective training. The right plot depicts the AUROC performance under PGD$^{1000}$ attacks, highlighting the adversarial robustness achieved by AROS as iterations progress.}

    \label{fig:LOSS_AUC}
  \end{center}
\end{figure}

In this section, we provide additional background on the theorems utilized in the main text. The structure of this background section is inspired by \cite{example2024}. Specifically, we present proofs for the theorems referenced in the main text, including the Hartman-Grobman Theorem, and offer a theoretical justification for our proposed objective function, $\mathcal{L}_{\text{SL}}$. Additionally, we include Figures \ref{fig:Jacob_Eigen} and \ref{fig:LOSS_AUC} as empirical support for the proposed loss function. The first figure illustrates how $\mathcal{L}_{\text{SL}}$ ensures that the real parts of the eigenvalues become negative, indicating stability, while the second figure highlights the stable decrease of the proposed loss function throughout the training process. Inspired by previous works leveraging control theory in deep learning, we set $T = 5$ as the integration time for the neural ODE layer and employ the Runge-Kutta method of order 5 as the solver. This choice ensures a balance between computational efficiency and robustness, allowing the ODE dynamics to stabilize feature representations effectively and mitigate the impact of adversarial perturbations.

  \subsection{Background}

The Hartman-Grobman Theorems is among the most powerful tools in dynamical systems \cite{hartman2002ordinary}. The Hartman-Grobman theorem allows us to depict the local phase portrait near certain equilibria in a nonlinear system using a similar, simpler linear system derived by computing the system's Jacobian matrix at the equilibrium point.

\textbf{Why does linearization at fixed points reveal behavior around the fixed point?}

For an \( n \)-dimensional linear system of differential equations (\(\dot{\mathbf{x}} = A\mathbf{x}\)) with a fixed point at the origin, we can classify behaviors such as saddle points, spirals, cycles, stars, and nodes based on the eigenvalues of the matrix \( A \). These behaviors are well-understood in the linear case. However, for nonlinear systems, analyzing the behavior becomes more challenging. Fortunately, the situation is not entirely intractable. By calculating the Jacobian matrix, or ``total derivative,'' \( J \), of the system and evaluating it at the fixed point, we obtain a linear approximation characterized by the matrix \( J \). The Hartman-Grobman theorem states that, within a neighborhood of the fixed point, if all eigenvalues of \( J \) have nonzero real parts, we can infer qualitative properties of the solutions to the nonlinear system. These include whether trajectories converge to or diverge from the equilibrium point and whether they spiral or behave like a node.

\textbf{\large Definitions}

\textbf{Definition 7.1: Homeomorphism}

A function \( h : X \to Y \) is called a homeomorphism between \( X \) and \( Y \) if it is a continuous bijection (both one-to-one and onto) with a continuous inverse (denoted \( h^{-1} \)). The existence of a homeomorphism implies that \( X \) and \( Y \) have analogous structures, as \( h \) and \( h^{-1} \) preserve the neighborhood relationships of points. Topologists often describe this concept as a process of stretching and bending without tearing.

\textbf{Definition 7.2: Topological Conjugacy}

Consider two maps \( f : X \to X \) and \( g : Y \to Y \). A map \( h : X \to Y \) is called a topological semi-conjugacy if it is continuous, onto, and satisfies \( h \circ f = g \circ h \), where \( \circ \) denotes function composition (sometimes written as \( h(f(\mathbf{x})) = g(h(\mathbf{x})) \) for \( \mathbf{x} \) in \( X \)). Furthermore, \( h \) is a topological conjugacy if it is a homeomorphism between \( X \) and \( Y \) (i.e., \( h \) is also one-to-one and has a continuous inverse). In this case, \( X \) and \( Y \) are said to be homeomorphic.

\textbf{Definition 7.3: Hyperbolic Fixed Point}

A hyperbolic fixed point of a system of differential equations is a point where all eigenvalues of the Jacobian evaluated at that point have nonzero real parts.

\textbf{Definition 7.4: Cauchy Sequence}

For the purposes of this document, we will provide a non-technical definition. A Cauchy sequence of functions is a series \( x_k = x_1, x_2, \ldots \) such that the functions become increasingly similar as \( k \to \infty \).

\textbf{Definition 7.5: Flow}

Let \( \dot{\mathbf{x}} = F(\mathbf{x}) \) be a system of differential equations with initial condition \( \mathbf{x}_0 \). Provided that the solutions exist and are unique (conditions given by the existence and uniqueness theorem; see, for example, (\cite{strogatz2018nonlinear}, pg. 149), the flow \( \phi(t; \mathbf{x}_0) \) of \( F(\mathbf{x}) \) provides the spatial solution over time starting from \( \mathbf{x}_0 \). An important property of flows is that small changes in initial conditions in phase space lead to continuous changes in flows, due to the continuity of the vector field in \( \mathbb{R}^n \).

\textbf{Definition 7.6: Orbit/Trajectory}

The set of all points in the flow \( \phi(t; \mathbf{x}_0) \) for the differential equations \( \dot{\mathbf{x}} = F(\mathbf{x}) \) is called the "orbit" or "trajectory" of \( F(\mathbf{x}) \) with initial condition \( \mathbf{x}_0 \). We denote the orbit as \( \phi(\mathbf{x}_0) \). When considering only \( t \geq 0 \), we refer to the "forward orbit" or "forward trajectory."

\subsection{Theorem and Proof}

\textbf{Theorem 7.1 The Hartman-Grobman Theorem}
\textit{
 Let \( \mathbf{x} \in \mathbb{R}^n \). Consider the nonlinear system \( \dot{\mathbf{x}} = f(\mathbf{x}) \) with flow \( \phi_t \) and the linear system \( \dot{\mathbf{x}} = A\mathbf{x} \), where \( A \) is the Jacobian \( Df(\mathbf{x}^*) \) of \( f \) at a hyperbolic fixed point \( \mathbf{x}^* \). Assume that we have appropriately shifted \( \mathbf{x}^* \) to the origin, i.e., \( \mathbf{x}^* = \mathbf{0} \).}
 
 \textit{ Suppose \( f \) is \( C^1 \) on some \( E \subset \mathbb{R}^n \) with \( \mathbf{0} \in E \). Let \( I_0 \subset \mathbb{R} \), \( U \subset \mathbb{R}^n \), and \( V \subset \mathbb{R}^n \) be neighborhoods containing the origin. Then there exists a homeomorphism  $H : U \to V $ such that, for all initial points \( \mathbf{x}_0 \in U \) and all \( t \in I_0 \),}

$$
H \circ \phi_t(\mathbf{x}_0) = e^{At}H(\mathbf{x}_0).
$$

\textit{Thus, the flow of the nonlinear system is homeomorphic to the flow \( e^{At} \) of the linear system given by the fundamental theorem for linear systems.}

\textbf{Proof}

Essentially, this theorem states that the nonlinear system \( \dot{\mathbf{x}} = f(\mathbf{x}) \) is locally homeomorphic to the linear system \( \dot{\mathbf{x}} = A\mathbf{x} \). To prove this, we begin by expressing \( A \) as the matrix

\[
\begin{pmatrix}
P & 0 \\
0 & Q
\end{pmatrix}
\]

where \( P \) and \( Q \) are sub-matrices of \( A \) such that the real parts of the eigenvalues of \( P \) are negative, and those of \( Q \) are positive. Finding such a matrix \( A \) may require finding a new basis for our linear system using linear algebra techniques. For more details, see section 1.8 on Jordan forms of matrices in Perko \cite{perko2013differential}.

Consider the solution \( \mathbf{x}(t, \mathbf{x}_0) \in \mathbb{R}^n \) given by

\[
\mathbf{x}(t, \mathbf{x}_0) = \phi_t(\mathbf{x}) = 
\begin{pmatrix}
\mathbf{y}(t, \mathbf{y}_0, \mathbf{z}_0) \\
\mathbf{z}(t, \mathbf{y}_0, \mathbf{z}_0)
\end{pmatrix}
\]

with \( \mathbf{x}_0 \in \mathbb{R}^n \) given by
Consider the solution:

\[
\mathbf{x}(t, \mathbf{x}_0) = \phi_t(\mathbf{x}) = 
\begin{pmatrix}
\mathbf{y}(t, \mathbf{y}_0, \mathbf{z}_0) \\
\mathbf{z}(t, \mathbf{y}_0, \mathbf{z}_0)
\end{pmatrix}
\]

where \( \mathbf{x}_0 \in \mathbb{R}^n \) is given by

\[
\mathbf{x}_0 =
\begin{pmatrix}
\mathbf{y}_0 \\
\mathbf{z}_0
\end{pmatrix}
\]

with \( \mathbf{y}_0 \in E^S \) (the stable subspace of \( A \)) and \( \mathbf{z}_0 \in E^U \) (the unstable subspace of \( A \)). The stable and unstable subspaces of \( A \) are defined as the spans of the eigenvectors of \( A \) corresponding to eigenvalues with negative and positive real parts, respectively. Define

\[
\widetilde{\mathbf{Y}}(\mathbf{y}_0, \mathbf{z}_0) = \mathbf{y}(1, \mathbf{y}_0, \mathbf{z}_0) - e^P \mathbf{y}_0,
\]
\[
\widetilde{\mathbf{Z}}(\mathbf{y}_0, \mathbf{z}_0) = \mathbf{z}(1, \mathbf{y}_0, \mathbf{z}_0) - e^Q \mathbf{z}_0.
\]

Here, \( \widetilde{\mathbf{Y}} \) and \( \widetilde{\mathbf{Z}} \) are functions of the trajectory with initial condition \( \mathbf{x}_0 \) evaluated at \( t = 1 \). If \( \mathbf{x}_0 = \mathbf{0} \), then \( \mathbf{y}_0 = \mathbf{z}_0 = \mathbf{0} \), leading to \( \widetilde{\mathbf{Y}}(\mathbf{0}) = \widetilde{\mathbf{Z}}(\mathbf{0}) = \mathbf{0} \) and thus \( D\widetilde{\mathbf{Y}}(\mathbf{0}) = D\widetilde{\mathbf{Z}}(\mathbf{0}) = \mathbf{0} \) since \( \mathbf{x}_0 \) is at the fixed point \( \mathbf{0} \). Since \( f \) is \( C^1 \) on \( E \), it follows that \( \widetilde{\mathbf{Y}} \) and \( \widetilde{\mathbf{Z}} \) are also \( C^1 \) on \( E \). Knowing that \( D\widetilde{\mathbf{Y}} \) and \( D\widetilde{\mathbf{Z}} \) are zero at the origin and that \( \widetilde{\mathbf{Y}} \) and \( \widetilde{\mathbf{Z}} \) are continuously differentiable, we can define a region around the origin such that \( \|\mathbf{y}_0\|^2 + \|\mathbf{z}_0\|^2 \leq s_0^2 \) for some sufficiently small \( s_0 \in \mathbb{R} \), where the norms of \( D\widetilde{\mathbf{Y}} \) and \( D\widetilde{\mathbf{Z}} \) are each less than some real number \( a \):

\[
\|D\widetilde{\mathbf{Y}}(\mathbf{y}_0, \mathbf{z}_0)\| \leq a,
\]
\[
\|D\widetilde{\mathbf{Z}}(\mathbf{y}_0, \mathbf{z}_0)\| \leq a.
\]

We now apply the mean value theorem: Let \( Y \) and \( Z \) be smooth functions such that \( Y = Z = 0 \) when \( \|\mathbf{y}_0\|^2 + \|\mathbf{z}_0\|^2 \geq s_0^2 \), and \( Y = \widetilde{Y} \) and \( Z = \widetilde{Z} \) when \( \|\mathbf{y}_0\|^2 + \|\mathbf{z}_0\|^2 \leq (s_0^2 / 2) \). Then the mean value theorem gives us

\[
|Y| \leq a \sqrt{\|\mathbf{y}_0\|^2 + \|\mathbf{z}_0\|^2} \leq a (\|\mathbf{y}_0\| + \|\mathbf{z}_0\|),
\]
\[
|Z| \leq a \sqrt{\|\mathbf{y}_0\|^2 + \|\mathbf{z}_0\|^2} \leq a (\|\mathbf{y}_0\| + \|\mathbf{z}_0\|).
\]

Let \( B = e^P \) and \( C = e^Q \). With proper normalization (see \cite{hartman2002ordinary}), we have \( b = \|B\| < 1 \) and \( c = \|C^{-1}\| < 1 \). We will now prove the existence of a homeomorphism \( H \) from \( U \) to \( V \) satisfying \( H \circ T = L \circ H \) using the method of successive approximations. Define the transformations \( L \), \( T \), and \( H \) as follows:

\[
L(\mathbf{y}, \mathbf{z}) =
\begin{pmatrix}
B\mathbf{y} \\
C\mathbf{z}
\end{pmatrix}
= e^{A\mathbf{x}}, \tag{7.1}
\]

\[
T(\mathbf{y}, \mathbf{z}) =
\begin{pmatrix}
B\mathbf{y} + Y(\mathbf{y}, \mathbf{z}) \\
C\mathbf{z} + Z(\mathbf{y}, \mathbf{z})
\end{pmatrix}, \tag{7.2}
\]

\[
H(\mathbf{x}) =
\begin{pmatrix}
\Phi(\mathbf{y}, \mathbf{z}) \\
\Psi(\mathbf{y}, \mathbf{z})
\end{pmatrix}. \tag{7.3}
\]

From equations (2.1)-(2.3) and our desired relation \( H \circ T = L \circ H \), we obtain

\[
B\Phi = \Phi(B\mathbf{y} + Y(\mathbf{y}, \mathbf{z}), C\mathbf{z} + Z(\mathbf{y}, \mathbf{z})),
\]
\[
C\Psi = \Psi(B\mathbf{y} + Y(\mathbf{y}, \mathbf{z}), C\mathbf{z} + Z(\mathbf{y}, \mathbf{z})).
\]

We define successive approximations for \( \Psi \) recursively by

\[
\Psi_0 = \mathbf{z}, \tag{7.4}
\]

\[
\Psi_{k+1} = C^{-1}\Psi_k(B\mathbf{y} + Y(\mathbf{y}, \mathbf{z}), C\mathbf{z} + Z(\mathbf{y}, \mathbf{z})), \quad k \in \mathbb{N}_0. \tag{7.5}
\]

This implies that we can increasingly approximate the function \( \Phi \) by following the recursion relations defined in equations (7.4)-(7.5). By induction, it follows that all \( \Psi_k \) are continuous because the flow \( \phi_t \) is continuous, which means \( \Psi_0 \) is continuous. Since \( C^{-1} \) is continuous, \( \Psi_1 \) is also continuous, and by induction, \( \Psi_k \) is continuous for all \( k \in \mathbb{N}_0 \). Additionally, it follows that \( \Psi_k(\mathbf{y}, \mathbf{z}) = \mathbf{z} \) whenever \( |\mathbf{y}| + |\mathbf{z}| \geq 2s_0 \)  \cite{perko2013differential}.

It can be shown by induction \cite{perko2013differential} that

\[
|\Psi_j(\mathbf{y}, \mathbf{z}) - \Psi_{j-1}(\mathbf{y}, \mathbf{z})| \leq Mr^j(|\mathbf{y}| + |\mathbf{z}|)^\sigma
\]

where \( j = 1, 2, \ldots \), \( r = c[2\max(a, b, c)]^\sigma \), \( c < 1 \), and \( \sigma \in (0, 1) \) such that \( r < 1 \). This leads to the conclusion that \( \Psi_k(\mathbf{y}, \mathbf{z}) \) forms a Cauchy sequence of continuous functions. These functions converge uniformly as \( k \to \infty \), and we denote the limiting function by \( \Psi(\mathbf{y}, \mathbf{z}) \). As with the \( \Psi_k \), it holds that \( \Psi(\mathbf{y}, \mathbf{z}) = \mathbf{z} \) for \( |\mathbf{y}| + |\mathbf{z}| \geq 2s_0 \).

A similar argument applies for \( B\Phi = \Phi(B\mathbf{y} + Y(\mathbf{y}, \mathbf{z}), C\mathbf{z} + Z(\mathbf{y}, \mathbf{z})) \), which can be rewritten as \( B^{-1}\Phi(\mathbf{y}, \mathbf{z}) = \Phi(B^{-1}\mathbf{y} + Y_1(\mathbf{y}, \mathbf{z}), C^{-1}\mathbf{z} + Z_1(\mathbf{y}, \mathbf{z})) \), where \( T^{-1} \) defines \( Y_1 \) and \( Z_1 \) as follows:

\[
T^{-1}(\mathbf{y}, \mathbf{z}) =
\begin{pmatrix}
B^{-1}\mathbf{y} + Y_1(\mathbf{y}, \mathbf{z}) \\
C^{-1}\mathbf{z} + Z_1(\mathbf{y}, \mathbf{z})
\end{pmatrix}.
\]

We can then solve for \( \Phi \) in the same manner as we solved for \( \Psi \) earlier, starting with \( \Phi_0 = \mathbf{y} \). After performing the necessary calculations to find \( \Psi \) and \( \Phi \), we obtain the homeomorphism \( H : \mathbb{R}^n \to \mathbb{R}^n \) given by

\[
H =
\begin{pmatrix}
\Phi \\
\Psi
\end{pmatrix}.
\tag{7.6}
\]

\subsection{Analysis and Justification of $\mathcal{L}_{\text{SL}}$}

 Consider a nonautonomous initial value ODE problem:
$ \frac{d\mathbf{z}(t)}{dt} = \mathbf{h}(\mathbf{z}(t), t), t \geq t_0;\quad   \mathbf{z}(t_0) = \mathbf{z}_0.$

Let $\mathbf{s}(\mathbf{z}_0, t_0, t)$ denote the solution of the ODE corresponding to the initial input $\mathbf{z}_0$ at time $t_0$.

\textbf{Definition 7.7 }\textit{(Equilibrium \cite{vidyasagar2002nonlinear})} A vector \( \mathbf{x}^* \) is called an equilibrium of a system if \( \mathbf{h}(\mathbf{x}^*, t) = 0, \, \forall t \geq 0 \).

\textbf{Definition 7.8 }\textit{(Stability \cite{bhatia2002stability})} A constant vector \( \mathbf{x}^* \in \mathbb{R}^d \) is a stable equilibrium point for a system if, for every \( \epsilon > 0 \) and every \( t_0 \in \mathbb{R}^+ \), there exists \( \delta(\epsilon, t_0) \) such that for each \( \mathbf{z}_0 \in B_\delta(\mathbf{x}^*) \), it holds that \( \|\mathbf{s}(\mathbf{z}_0, t_0, t) - \mathbf{x}^*\| < \epsilon, \, \forall t \geq t_0 \). Where $B_\delta(\mathbf{x}^*) = \{\mathbf{x} \in \mathbb{R}^d : \|\mathbf{x} - \mathbf{x}^*\| < \delta\}.
$

\textbf{Definition 7.9} \textit{(Attractivity \cite{bhatia2002stability})}  A constant vector \( \mathbf{x}^* \in \mathbb{R}^d \) is an attractive equilibrium point for (1) if for every \( t_0 \in \mathbb{R}^+ \), there exists \( \delta(t_0) > 0 \) such that for every \( \mathbf{z}_0 \in B_\delta(\mathbf{x}^*) \),

\[
\lim_{t \to +\infty} \|\mathbf{s}(\mathbf{z}_0, t_0, t) - \mathbf{x}^*\| = 0.
\]

\textbf{Definition 7.10} \textit{(Asymptotic stability \cite{bhatia2002stability})} A constant vector \( \mathbf{x}^* \in \mathbb{R}^d \) is said to be asymptotically stable if it is both stable and attractive. Note that if an equilibrium is exponentially stable, it is also asymptotically stable with exponential convergence.

Our goal is to make the contaminated instance \( \hat{\mathbf{x}} \) converge to the clean instance \( \mathbf{x} \). To achieve this evolution, we impose constraints on the ODE to output \( \mathbf{z}(T) = \mathbf{x} \) when the input is \( \mathbf{z}(0) = \hat{\mathbf{x}} \in B_\delta(\mathbf{x}) \). To ensure that

\[
\lim_{t \to +\infty} \|\mathbf{s}(\hat{\mathbf{x}}, t) - \mathbf{x}\| = 0,
\]

where \( \hat{\mathbf{x}} \in B_\delta(\mathbf{x}) \), we make all \( \mathbf{x} \in \mathcal{X} \) asymptotically stable equilibrium points.

\textbf{Theorem 7.2} \textit{Suppose the perturbed instance \( \hat{\mathbf{x}} \) is produced by adding a perturbation smaller than \( \delta \) to the clean instance. If all the clean instances \( \mathbf{x} \in \mathcal{X} \) are asymptotically stable equilibrium points of ODE (1), then there exists \( \delta > 0 \) such that for each contaminated instance \( \hat{\mathbf{x}} \in \{\hat{\mathbf{x}} : \hat{\mathbf{x}} \in \hat{\mathcal{X}}, \hat{\mathbf{x}} \not\in \mathcal{X}\} \), there exists \( \mathbf{x} \in \mathcal{X} \) satisfying }

\[
\lim_{t \to +\infty} \|\mathbf{s}(\hat{\mathbf{x}}, t) - \mathbf{x}\| = 0.
\]

\textbf{Proof:}

According to the definition of asymptotic stability, a constant vector of a system is asymptotically stable if it is both stable and attractive. Based on the definition of stability of (1), for every \( \epsilon > 0 \) and every \( t_0 \in \mathbb{R}^+ \), there exists \( \delta_1 = \delta(\epsilon, 0) \) such that

\[
\forall \hat{\mathbf{x}} \in B_{\delta_1}(\mathbf{x}) \implies \|\mathbf{s}(\hat{\mathbf{x}}, t) - \mathbf{x}\| < \epsilon, \, \forall t \geq t_0.
\]

Based on the attractivity definition, there exists \( \delta_2 = \delta(0) > 0 \) such that

\[
\hat{\mathbf{x}} \in B_{\delta_2}(\mathbf{x}), \, \lim_{t \to +\infty} \|\mathbf{s}(\hat{\mathbf{x}}, t) - \mathbf{x}\| = 0.
\]

We set \( \delta = \min\{\delta_1, \delta_2\} \). Since the perturbed instance \( \hat{\mathbf{x}} \) is produced by adding a perturbation smaller than \( \delta \) to the clean instance, then for each contaminated instance \( \hat{\mathbf{x}} \in \{\hat{\mathbf{x}} : \hat{\mathbf{x}} \in \hat{\mathcal{X}}, \hat{\mathbf{x}} \not\in \mathcal{X}\} \), there exists a clean instance \( \mathbf{x} \in \mathcal{X} \) such that \( \hat{\mathbf{x}} \in B_\delta(\mathbf{x}) \). Because the clean instance \( \mathbf{x} \) is an asymptotically stable equilibrium point of (1), we have

\[
\lim_{t \to +\infty} \|\mathbf{s}(\hat{\mathbf{x}}, t) - \mathbf{x}\| = 0.
\]
 
This theorem guarantees that if we make the clean instance \( \mathbf{x} \) an asymptotically stable equilibrium point, the ODE can reduce the perturbation and cause the perturbed instance to approach the clean instance. This can help improve the robustness of the DNN and aid it in defending against adversarial attacks.

\textbf{Theoretical Justification for Lyapunov-Stable Embedding Representations in Neural ODEs:}

We denote by \( \lambda \) the pushforward measure, which is a probability distribution derived from the original distribution of the input data under the continuous feature extractor mapping \( h_\phi \). The conditional probability distribution for the embeddings of each class in the training set, \( l \in \{1, \ldots, L\} \), has compact support \( E_l \subset \mathbb{R}^n \), as \( E_l \) is closed and \( h_\phi(X) \) is bounded in \( \mathbb{R}^n \).

\noindent\textbf{Premise} The input data are sampled from a probability distribution defined over a compact metric space. The feature extractor \( f_\theta \) is injective and continuous. Furthermore, the supports of each class in the embedding space are pairwise disjoint,

\textbf{Lemma 1.} \textit{Given \( k \) distinct points \( \mathbf{z}_i \in \mathbb{R}^n \) and matrices \( \mathbf{A}_i \in \mathbb{R}^{n \times n} \), for \( i = 1, \ldots, k \), there exists a function \( g \in \mathcal{C}^1(\mathbb{R}^n, \mathbb{R}^n) \) such that \( g(\mathbf{z}_i) = 0 \) and \( \nabla g(\mathbf{z}_i) = \mathbf{A}_i \).}

\textbf{Proof:} The set of finite points \( \{z_1, \ldots, z_k\} \) is closed, and this lemma is an immediate consequence of the Whitney extension theorem \cite{mcshane1934extension}.

We restrict \( h_\phi \) to be in \( C^1(\mathbb{R}^n, \mathbb{R}^n) \) to satisfy the condition in Theorem 1. $C^1$ represents the function with first-order derivative. From \cite{hornik1991approximation}, we also know that standard multilayer feedforward networks with as few as a single hidden layer and arbitrary bounded and non-constant activation functions are universal approximators for \( C^1(\mathbb{R}^n, \mathbb{R}^n) \) functions with respect to some performance criteria, provided only that sufficiently many hidden units are available.

Suppose for each class \( l = 1, \ldots, L \), the embedding feature set \( E_l = \{z_1^{(l)}, \ldots, z_k^{(l)}\} \) is finite. For each \( i = 1, \ldots, k \), let \( A_i \in \mathbb{R}^{n \times n} \) be a strictly diagonally dominant matrix with every main diagonal entry negative, such that the eigenvalues of \( A_i \) all have negative real parts. From Theorem 3, each \( A_i \) is non-singular and every eigenvalue of \( A_i \) has negative real part. Therefore, from Theorem 2 and Lemma 1, there exists a function \( h_\phi \) such that all \( z_i^{(l)} \) are Lyapunov-stable equilibrium points with corresponding first derivative \( \nabla h_\phi(z_i^{(l)}) = A_i \). This shows that if there exist only finite representation points for each class, we can find a function \( h_\phi \) such that all inputs to the neural ODE layer are Lyapunov-stable equilibrium points for \( h_\phi \) and
\begin{enumerate}
    \item[(I)] \( \mathbb{E}_{\lambda} \|h_\phi(X_{\text{train}})\|_2 = 0 \),
    \item[(II)] \( \mathbb{E}_{\lambda} \big[\nabla h_\phi(X_{\text{train}})\big]_{ii} < 0 \),
    \item[(III)] \( \mathbb{E}_{\lambda} \left[\big[\nabla h_\phi(X_{\text{train}})\big]_{ii} - \sum_{j \neq i} \big[\nabla h_\phi(X_{\text{train}})\big]_{ij}\right] > 0. \)
\end{enumerate}

We will show that under mild conditions, for all \( \epsilon > 0 \), we can find a continuous function \( h_\phi \) with finitely many stable equilibrium points such that conditions (II) and (III) above hold and condition (I) is replaced by \( \mathbb{E}_{\lambda} \|h_\phi(X_{\text{train}})\|_2 < \epsilon \). This motivates the optimization constraints in (I, II, III). 

\textbf{Theorem 7.3.} \textit{Suppose Premise hold. If \( \lambda \) is not a continuous measure on \( E_l \) for each \( l = 1, \ldots, L \), then the following holds:}
\textit{\begin{enumerate}
    \item The function space satisfying the constraints in (I, II, III) is non-empty for all \( \epsilon > 0 \).
    \item If additionally the restriction of \( \lambda \) to any open set \( O \subset E_l \) is a continuous measure, then we can find such a function such that each support \( E_l \) almost surely satisfies the conditions in (I, II, III).
\end{enumerate}
}

\textbf{Proof:} Consider \( g(z) = [g^{(1)}(z^{(1)}), \ldots, g^{(n)}(z^{(n)})] \) with each \( g^{(i)}(z^{(i)}) \in C^1(\mathbb{R}, \mathbb{R}) \). Since \( g^{(i)}(z^{(i)}) \) depends only on \( z^{(i)} \), \( \nabla g_\theta(z) \) is a diagonal matrix with all off-diagonal elements being \( 0 \). The constraint (III) is thus immediately satisfied, and it suffices to show that there exists such an \( f \) satisfying the constraints (II) and (III).

Select a point \( z_l = (z_l^{(1)}, \ldots, z_l^{(n)}) \) from the interior of each \( E_l \), for \( l = 1, \ldots, L \). Let \( g^{(i)}(z^{(i)}) = -\nu(z^{(i)} - z_l^{(i)}) \) on each \( E_l \), where \( \nu > 0 \). Then \( g(z) \) satisfies (II) for all \( \nu > 0 \), and \( z_l \) is a Lyapunov-stable equilibrium point for each \( l \) since \( \nabla h_\phi(z) \) is a diagonal matrix with negative diagonal values. Since each \( E_l \subset \mathbb{R}^n \) is compact, we have that \( \forall \epsilon > 0, \exists \nu > 0 \) sufficiently small such that \( |f^{(i)}(z^{(i)})| < \epsilon \) for all \( z \in \bigcup_l E_l \). The constraint (I) is therefore satisfied for \( f(z) \) with a sufficiently small \( \nu \). Since \( \bigcup_l E_l \) is closed, the Whitney extension theorem \cite{mcshane1934extension} can be applied to extend \( f(z) \) to a function in \( C^1(\mathbb{R}^n, \mathbb{R}^n) \).

\end{document}